\pdfoutput=1
\documentclass[11pt,onecolumn]{article}
\usepackage{files/cvpr}
\usepackage{times}
\usepackage{epsfig}
\usepackage{amsmath,  amssymb, graphicx, xfrac, amsfonts, amsbsy, algorithm, color, cite, url, multirow}
\usepackage[caption=false]{subfig}
\usepackage[noend]{algpseudocode}

\usepackage[pagebackref=true,breaklinks=true,colorlinks,bookmarks=false]{hyperref}

\cvprfinalcopy % *** Uncomment this line for the final submission

\def\Dbf{\mathbf{D}}
\def\phibf{\pmb{\phi}}

\def\Phibf{\mathbf{\Phi}}
\def\xbf{\mathbf{x}}

\def\tbf{\pmb{\theta}}

\def \g { \mathbf{D} }
\def \L { \mathbf{L} }
\def\yphin{\pmb{ \tilde{\mathrm{y}} }^{ \pmb{\phi} }_n }
\def\Ybf{ \pmb{ \mathrm{y} } }

\def\gYbf{ \g \Ybf }
\def \gYbfe{ \g \widehat{ \Ybf } }
\def \gkYbfe{ \g_k \widehat{ \Ybf } }

\def\Yphibf{ \Ybf^{ \pmb{\phi} } }
\def\gYphibf{ \g \Yphibf }
\def\w{ \mathrm{w} }
\def\wbf{ \pmb{\w } }
\def\wbfe{ \widehat{\pmb{\mathrm{w}} } }
\def\X {\mathrm{x} }
\def\Xbf{ \pmb{ \X } }
\def \gXbf { \g \Xbf }
\def \gkXbf { \g_k \Xbf }
\def \gxXbf { \g_x \Xbf }
\def \gyXbf { \g_y \Xbf }

\def\Ophi{ \Omega^{\pmb{\phi}} }
\def\Mphin{ M^{\phibf}_n }

\def\dO{ \partial \Omega }
\def\O'{ \Omega'}

\def\P{ \mathbf{ P } }
\def\PO{ \P_{\Omega} }
\def\I{ \mathbf{ I } }
\def\IO{ \I_{\Omega} }

\def\Xest{ \widehat{ \Xbf } }

\def\R{\mathbb{R}}

\def\Ncal{\mathcal{N}}
\def\Dcal{\mathcal{D}}
\def\Rcal{\mathcal{R}}
\def\Xcal{\mathcal{X}}
\def\Tcal{\mathcal{\varTheta}}

\def\argmin{\mathop{\mathrm{arg\,min}}}
\def\argmax{\mathop{\mathrm{arg\,max}}}

\makeatletter
\def\BState{\State\hskip-\ALG@thistlm}
\makeatother

\ifcvprfinal\fi
\begin{document}

%%%%%%%%%%%%%%%%%%%%%%%%%%%%%%%%%%%%%%%%%%%%%
%% Title
%%%%%%%%%%%%%%%%%%%%%%%%%%%%%%%%%%%%%%%%%%%%%

\title{ \LARGE {\normalfont Computational Mapping of the Ground Reflectivity with Laser Scanners}}

%%%%%%%%%%%%%%%%%%%%%%%%%%%%%%%%%%%%%%%%%%%%%
%% Authors
%%%%%%%%%%%%%%%%%%%%%%%%%%%%%%%%%%%%%%%%%%%%%

\author{ Juan Castorena \thanks{ J. Castorena (email: jcastore@ford.com) is with Ford Motor Company, Dearborn, MI, USA  } }

\maketitle
%\thispagestyle{empty}

%%%%%%%%%%%%%%%%%%%%%%%%%%%%%%%%%%%%%%%%%%%%%
%% Abstract
%%%%%%%%%%%%%%%%%%%%%%%%%%%%%%%%%%%%%%%%%%%%%

\begin{abstract}
	{\normalfont
In this investigation we focus on the problem of mapping the ground reflectivity with multiple laser scanners mounted on mobile robots/vehicles. The problem originates because regions of the ground become populated with a varying number of reflectivity measurements whose value depends on the observer and its corresponding perspective. Here, we propose a novel automatic, data-driven computational mapping framework specifically aimed at preserving edge sharpness in the map reconstruction process and that considers the sources of measurement variation. Our new formulation generates map-perspective gradients and applies sub-set selection fusion and de-noising operators to these through iterative algorithms that minimize an $\ell_1$ sparse regularized least squares formulation. Reconstruction of the ground reflectivity is then carried out based on Poisson's formulation posed as an $\ell_2$ term promoting consistency with the fused gradient of map-perspectives and a term that ensures equality constraints with reference measurement map data. We demonstrate our new framework outperforms the capabilities of existing ones with experiments realized on Ford's fleet of autonomous vehicles. For example, we show we can achieve map enhancement (i.e., contrast enhancement), artifact removal, de-noising and map-stitching without requiring an additional reflectivity adjustment to calibrate sensors to the specific mounting and robot/vehicle motion. 
}
\end{abstract}

%%%%%%%%%%%%%%%%%%%%%%%%%%%%%%%%%%%%%%%%%%%%%
%% Sections
%%%%%%%%%%%%%%%%%%%%%%%%%%%%%%%%%%%%%%%%%%%%%

\section{Introduction}
\label{Sec:Introduction} 

% Mapping motivation
Robotic mapping is one of the key fundamental tasks within the context of autonomous mobile robots and in particular in autonomous vehicles. %Part of the current ongoing research within these areas focuses on developing more advanced techniques and algorithms that can provide reliable and precise 3-D maps of the environment. 
The interest on this area is based on results that show that map augmentation can facilitate mobile robot/vehicle navigation %(e.g.,localization, path planning) 
and perception %(e.g, seeing under weather adverse conditions, obstacle detection/tracking) 
\cite{Google2016, Popular2016}.
\begin{figure}[t]
	\centering
	{\includegraphics[width = 9 cm]{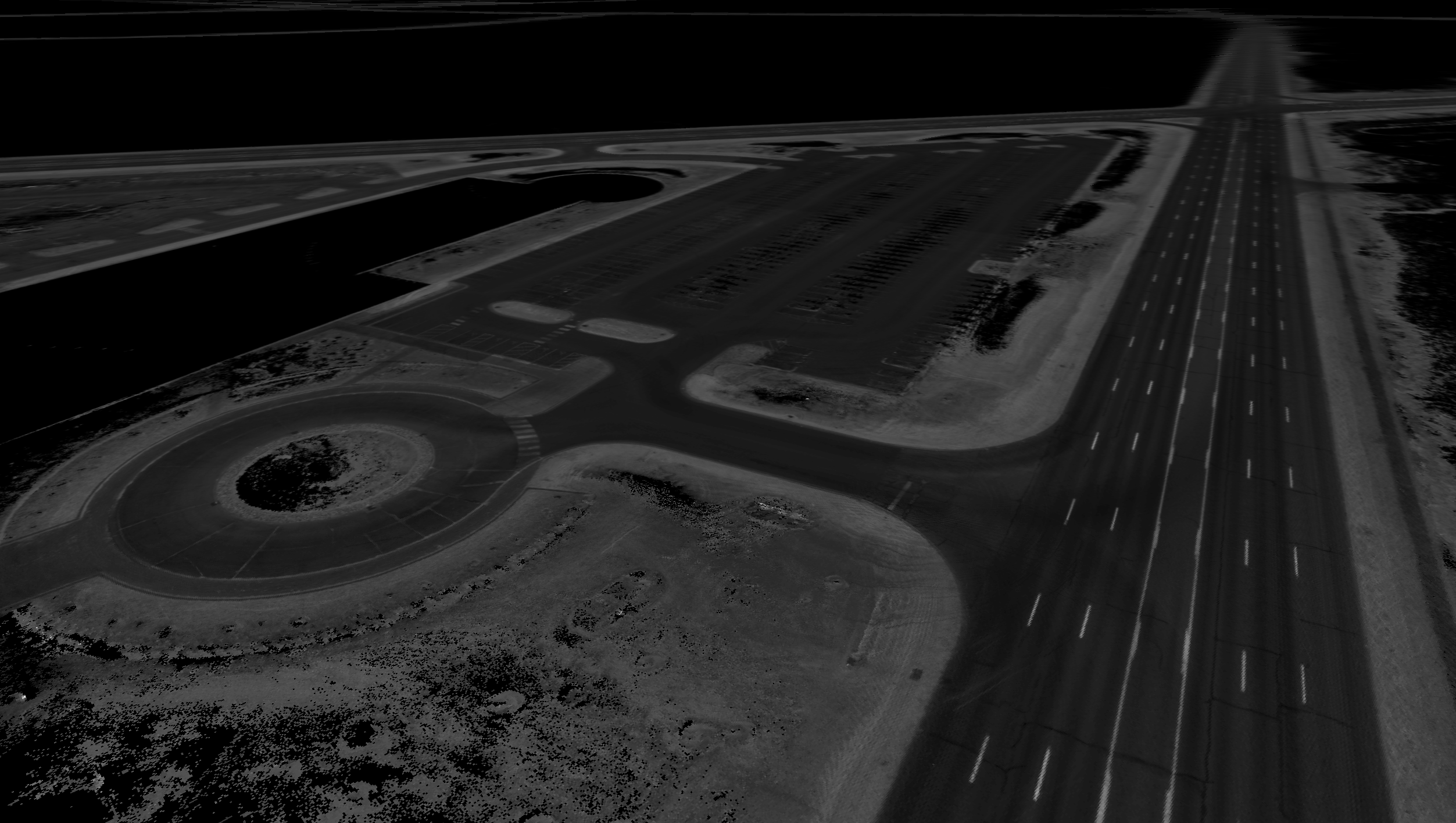}}
	
	\caption{Map of the ground reflectivity (after full-SLAM). 
		%(e.g., iSAM \cite{Kaess.Ranganathan.Dellaert2008, Kaess.etal2012}). 
		%Reflectivity value at a cell is given as the expectation over all calibrated measurements associated with that cell.
		} %( an example of the distribution of the intensity returns in a cell is shown in Figure~\ref{fig:cellReturns} ). }
	\label{fig:fullUncalibratedMap}
\end{figure}
Mapping of the environment is typically carried out by a mobile robot/vehicle equipped with position sensors (e.g., GPS, IMU, odometry) and range finders (e.g., lasers, radars, sonar) and/or cameras that perceive the environment while in motion. For example, \cite{Napier.Newman2012} generates maps by capturing local orthographic projections of the ground with multiple vision cameras, \cite{Se.Lowe.Little2001, Se.Lowe.Little2002, Se.Lowe.Little2005, Karlsson.etal2005} encodes a map of landmarks captured with a camera and matches them with those captured on the fly for localization. Unfortunately, capturing the outdoor environment with cameras is still susceptible to ambient illumination (e.g., unpredictable lighting) problems. LIDAR on the other hand rely on active illumination, does not significantly suffer from ambient illumination while also provides direct 3-D measurements of the environment. For these reasons, it is considered among one of the most reliable sensing modalities for 3-D mapping \cite{Levinson.Thrun2007, Levinson10}. 

Current LIDAR configurations are capable of measuring both the range and reflectivity for each pulse emitted. Here, reflectivity implies a factory calibration against a target of known reflectance has been performed. Specific benefits that accompany this additional modality include that of increased perception capabilities, for example to distinguish road features \cite{Guan14, Olsen.Kuester.Chang.Hutchinson.2010, Guan.etall2014, Pu.Rutzinger.Vosselman.Elberink.2011, Chin.Olsen.2014} in urban environments. %Research efforts have used it to distinguish road features \cite{Guan14, Olsen.Kuester.Chang.Hutchinson.2010, Guan.etall2014, Pu.Rutzinger.Vosselman.Elberink.2011} establishing as the top choice to collect and classify road profiles \cite{Pu.Rutzinger.Vosselman.Elberink.2011, Chin.Olsen.2014}. 
One of the problems associated with mapping using this additional modality is that its measurements depend on a reflectivity response. This itself is dependent upon the laser-detector pair (i.e., observer), angle of incidence and range (i.e., perspective) \cite{Kashani15}. Thus, assuming a reflectivity invariant measurement model is quite often inadequate \cite{Harvey.Lichti02, Burgard.Triebel.Andreasson2005} especially if one considers that the ground is viewed from multiple locations, perspectives and observers. 
%multiple mobile robots/vehicles equipped with multiple LIDARs survey the environment from different positions during different times. 
%In this investigation we deal with this problem with a special focus on mapping the ground reflectivity. Most of the algorithms developed to this date have been designed for the purpose of properly projecting the range-reflectivity pairs of measurements into a global reference frame, for example by solving an online or a full SLAM problem \cite{Lu.Milios1997, Thrun.Leonard2008, Durrant-Whyte.Bailey2006, Bailey.Durrant-Whyte2006}. Unfortunately, the development of advanced methods to optimize usage of the vast quantity of projected reflectivity measurements is very limited. 
%Here, we propose a novel automatic, data-driven computational mapping of the ground reflectivity framework for this purpose. 

\begin{figure}[t]
	\centering
	\subfloat[Asphalt: $\mu \approx 25, \sigma \approx 12$, $D_{KL}(P || Q) = 0.38 $ bits]{ {\includegraphics[width = 7.5 cm]{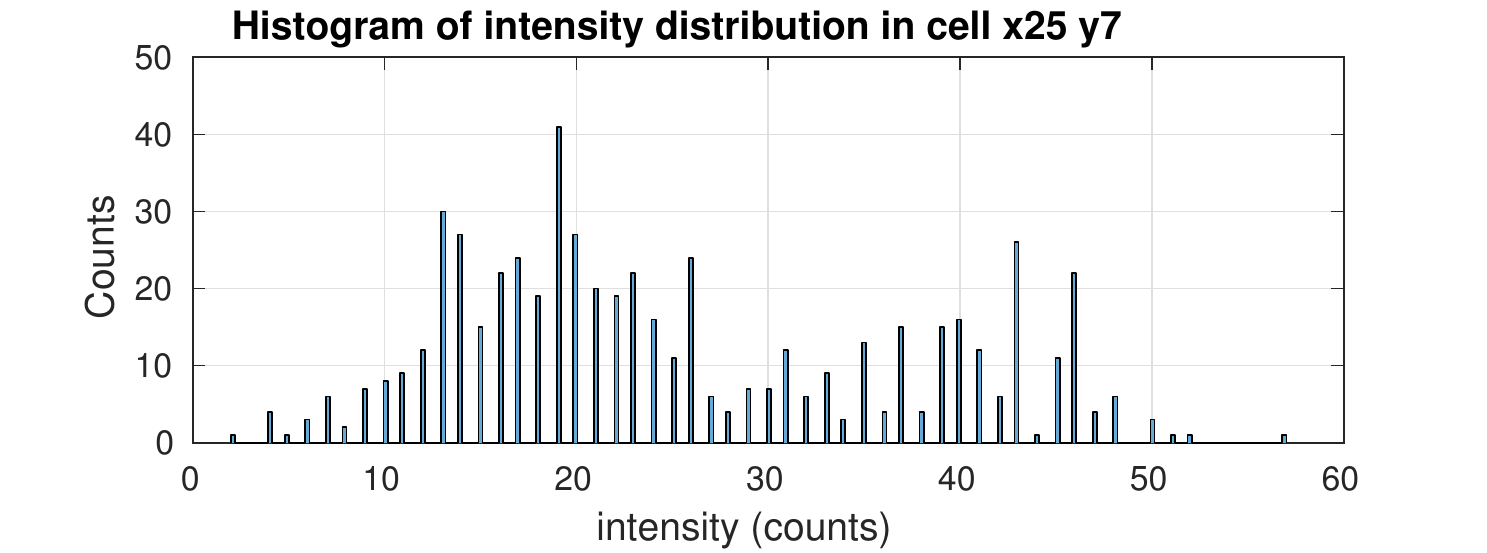} } }
	\subfloat[W. lane: $\mu \approx 30, \sigma \approx 10$, $D_{KL}(P || Q) = 0.31 $ bits]{ {\includegraphics[width = 7.5 cm]{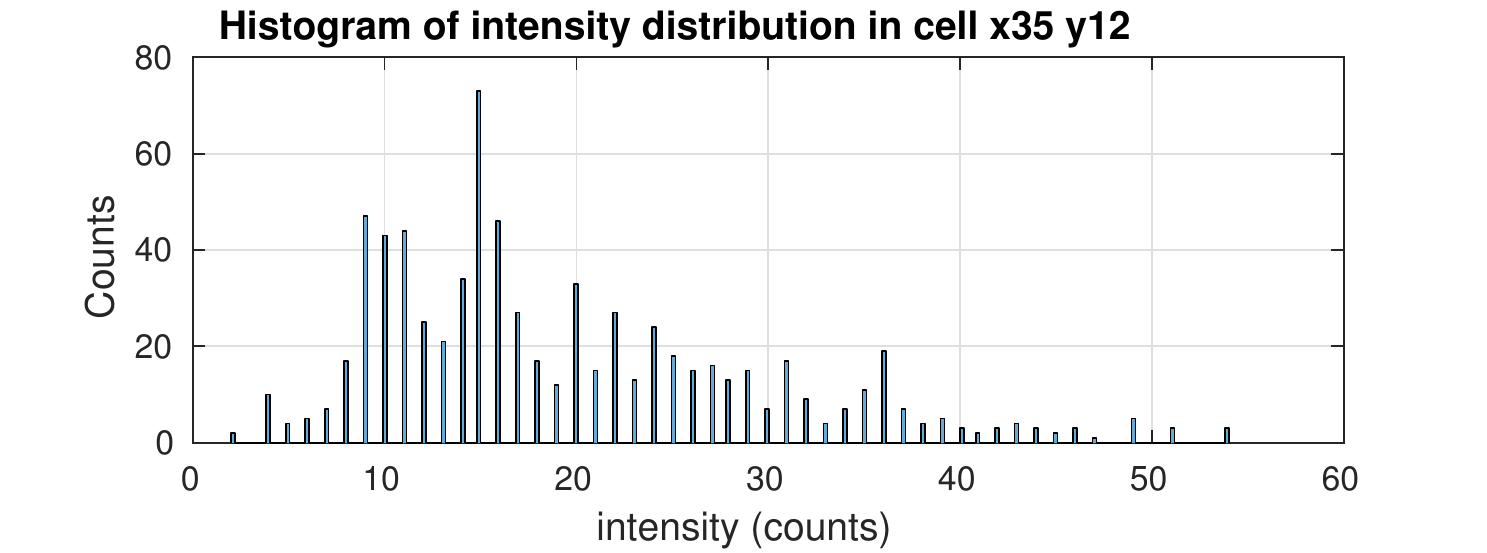}} }	
	\caption{Histogram distributions of reflectivity measurements from multiple HDL-32E's in $10 \times 10$ cm. cells. (a) Cell in plain asphalt. (b) Cell in a white lane marking.}
	\label{fig:histogramIntensity}
\end{figure}

Typically, a map of the ground reflectivity is approximated as a uniformly quantized space plane of cell reflectivities. %This plane is populated with globally projected range-reflectivity pair measurements of the ground using the robot/vehicle pose and an online or full-SLAM post alignment \cite{Lu.Milios1997, Thrun.Leonard2008, Durrant-Whyte.Bailey2006, Bailey.Durrant-Whyte2006}.
%Under this scenario, each cell in the map of area size $\partial x \times \partial x $ (e.g., \cite{Levinson10} uses $\partial x  = 15 $ cm) is populated with a varying number of measurements.
Each of these cells represent an area of size $\partial x \times \partial x $ (e.g., \cite{Levinson10} uses $\partial x  = 15 $ cm). 
When a robot/vehicle is surveying a region, each reflectivity measurement of the ground is associated with a cell in the map depending on the projection of its corresponding range into the global reference frame (assuming full SLAM \cite{Durrant-Whyte.Bailey2006, Bailey.Durrant-Whyte2006} has already taken place). 
These projected measurements are non-uniformly distributed across the ground depending on the sensor configuration, the scanning pattern and the navigation path followed by the robot/vehicle while surveying.
%
% Option 2: Thus, each reflectivity measurement can be associated with a cell position given its globally projected range measurements. %Most of the algorithms developed to this date have been designed for the purpose of properly projecting the range-reflectivity pairs of measurements into a global reference frame, for example by solving an online or a full SLAM problem \cite{Lu.Milios1997, Thrun.Leonard2008, Durrant-Whyte.Bailey2006, Bailey.Durrant-Whyte2006}. %Unfortunately, the development of advanced methods to optimize usage of the vast quantity of projected reflectivity measurements is very limited.  
%
Na\"{\i}ve methods for estimating the cell reflectivity average all the raw reflectivity measurements associated to the cell, assuming these follow a Gaussian distribution model. However, as shown in Figure \ref{fig:histogramIntensity} and as measured by the Kullback-Leibler distance $ D_{KL}(P || Q)$ this model is far from being true in practice. %Examples of the Kullback-Leibler distance between the distribution $P$ and a Gaussian distribution $Q$ is denoted by $ D_{KL}(P || Q)$. %Maps generated with this method present many characteristic artifacts, examples of these are shown in Figures \ref{fig:artifactRemoval},a,d,g.

% Target based approaches
Current mapping methods use an additional after-factory reflectivity calibration stage to decouple the observations from the variations in reflectivity ~\cite{Kashani15}. Most of these approaches aim at estimating the response of each laser and its perspectives through a curve fit between the measurement signatures of a reference target and its corresponding known reflectance. For example,~\cite{Ahokas.etal2006, Kaasalainen.etal2009, Pfeifer.etal2007, Pfeifer.etal2008} exploit standard calibration targets and uses least squares to estimate the response of the laser scanner. Unfortunately, the options of target reflectances is very limited and measurements of these represent only a tiny fraction of the available data one may encounter in real world scenarios. %Moreover, the cost of these targets is unfeasible for the general consumer. %Follow up work in \cite{Coren.Sterzai2006, Hofle.Pfeifer2007} extended this approach by using road features manually predefined instead of standard calibration targets. 

% Target-less based approaches
Within the context of autonomous vehicles, \cite{Levinson10, Levinson.Thrun2010 } propose an automatic, target-less calibration method that uses a probabilistic framework to learn the reflectivity response model of each laser from its measurements. This method uses the expectation maximization iterative algorithm to generate the maximum likelihood model $p(y|x;b)$, that indicates the probability that for a given reflectivity $x$, laser $b$ observes reflectivity $y$. However, to establish correspondences, the algorithm iterates between improving both the likelihood of the map $\Ybf$ and the laser responses. In practice, the whole process is time consuming and computationally intensive since it requires a data collection and its corresponding map generation step. Moreover,
%Moreover, this calibration process will work optimally only on the calibration data since it learned under the specific settings at the time of collection (e.g., sensor configuration, robot/vehicle navigation path). %Calibrations used with different sensor configurations and surveying paths are sup-optimal. Thus, 
even after application of state of the art calibrations results in maps with artifacts, noise and of low contrast from global smoothing. 

% Our new approach
In this investigation we propose a novel computational mapping framework specifically aimed at preserving edge sharpness in the map reconstruction process. %Measurements of the ground reflectivity can come from multiple laser scanners surveying from different positions during different times. 
In particular, our approach first generates all the map-perspectives where each of these represent the map from a perspective of a laser observer. The problem with operating on these map-perspectives is that although these have overlapping observations of the same region, its perspective influences the characteristic reflectivity response. To decouple these from the measurement parameters we apply the gradient operator to each of them.  Under this representation, the components of the gradients of map-perspectives become statistically stationary \cite{Unser.Tafti2011, Bostan.Kamilov.Nilchian.Unser2013} which allows one to operate on each invariantly. The algorithmic operators here proposed are those of perspective selection and de-noising which solve a corresponding $\ell_1$ sparse regularized least squares cost function. Finally, the resulting map-perspective gradients are fused and used in a Poisson's formulation for map reconstruction of the ground reflectivity. The optimization carried out is posed as an $\ell_2$ term promoting consistency with the gradient of fused map-perspective gradients and a term that ensures equality constraints with reference measured data. %As a note, we would like to add that the idea of using gradients on LIDAR data has been used previosuly in~\cite{Castorena.Kamilov.Boufounos2016}.

%In the next section we formulate and develop our novel computational mapping framework along with the proposed algorithms. Section~\ref{Sec:Experiments} presents experimental validating results with data collected by surveying urban ground environments with Ford's fleet of autonomous vehicles. Finally, Section~\ref{sec:conclusion} concludes and discusses our findings.

% Our new approach: contributions
\subsection{Contributions}
\label{sec:contribution}

In this investigation we propose a novel computational mapping framework specifically aimed at preserving edge sharpness in the map reconstruction process. %Measurements of the ground reflectivity can come from multiple laser scanners surveying from different positions during different times. 
In particular, our approach first generates all the map-perspectives where each of these represent the map from a perspective of an individual laser observer. The problem with operating on these map-perspectives is that although these have overlapping views of the same region, its perspective influences the characteristic reflectivity response. To decouple these from the measurement parameters we apply the gradient operator (providing the edge information) to each of them.  Under this representation, the components of the gradients of map-perspectives become statistically stationary \cite{Unser.Tafti2011, Bostan.Kamilov.Nilchian.Unser2013} which allows one to operate on each invariantly. Summarizing, the main contributions of this investigation are:
\begin{itemize}
	% Formulation contribution
	\item We develop a new edge guided fusion and reflectivity map reconstruction formulation for laser scanned data. The strategy used in the formulation consists as a first step on the fusion of reflectivity edges generated by the individual lasers through any combination of the following: (a) selection of all edges or sparse selection of only the best edges and (b) denoising edge information. Both sparse selection and denoising are solved through a corresponding inverse problem that minimizes an $\ell_1$ sparse regularized least squares cost function. The second map reconstruction step consists on solving an inverse problem with a cost involving an $\ell_2$ term that promotes consistency with the map of fused edges and a term that ensures equality constraints with reference measurement map data.
	%Reconstruction is carried out in a Poisson's formulation throught the optimization of an $\ell_2$ term promoting consistency with the fused edges and a term that ensures equality constraints with reference measured data.

	% Algorithmic approach contribution: Optimization approach to estimate the ground reflectivity map.
	\item Propose to use Nesterov's accelerated gradient descent \cite{Nesterov1983} to develop a faster Poisson based map reconstruction optimization which convergences with rate of $O(1/k^2)$ over conventional $O(1/k)$ of gradient descent.
	
	% Experimentation findings contribution
	\item We provide the experimentation that validates our new formulation and that demonstrates substantial improvements. In particular we demonstrate that our edge guided fusion of individual gradient-maps and reconstruction framework (a) avoids the formation of mapping artifacts generated from vehicle motion and laser scanning patterns, (b) improves upon the quality of reconstructed maps, (c) solves the reflectivity measurement dependencies upon the laser beam, angle of incidence and range which previously required an offline post-factory reflectivity calibration procedure and (d) is effective also in the context of localization within a map and road mark segmentation.
\end{itemize}

\subsection{Outline}
\label{sec:outline}
In section \ref{Sec:ProposedApproach} we formulate and develop our novel computational edge guided mapping framework along with the proposed algorithms. Section~\ref{Sec:Experiments} presents experimental validating results of our map reconstruction formulation with data collected by surveying urban ground environments with Ford's fleet of autonomous vehicles. In addition, we also included experimental work on the application of our formulation to localization and road mark segmentation tasks. Finally, Section \ref{sec:Conclusion} concludes our findings.
 		% Introduction
%\input{SecBackground} 		% Background
\section{Proposed Approach}
\label{Sec:ProposedApproach}

%%%%%%%%%%%%%%%%%%%%%%%%%%%%%%%%%%%%%%%%%%%%%
%% Subsection 1
%%%%%%%%%%%%%%%%%%%%%%%%%%%%%%%%%%%%%%%%%%%%%

\subsection{Problem Formulation}
\label{sec:Formulation}

The formulation we propose considers a map of the ground reflectivity compactly represented by the reflectivities $ \Xbf \in \R^{N_x \times N_y} \cup \{ \infty\}^{N_x \times N_y} $ where $N_x$ and $N_y$ are the number of horizontal and vertical grid cells respectively. Here, the finite values in $\Xbf$ represent the occupancy of measurements within the gridded map comprised of $N=N_xN_y$ cells. Note that in the remainder of this paper we often use $n\in\{1,\ldots,N\}$ to index the cells of $\Xbf$ and other similarly sized matrices, essentially vectorizing them.

\begin{figure*}[t]
	\centering
	\subfloat[Map-perspective of laser 1]{\includegraphics[width=0.24\linewidth]{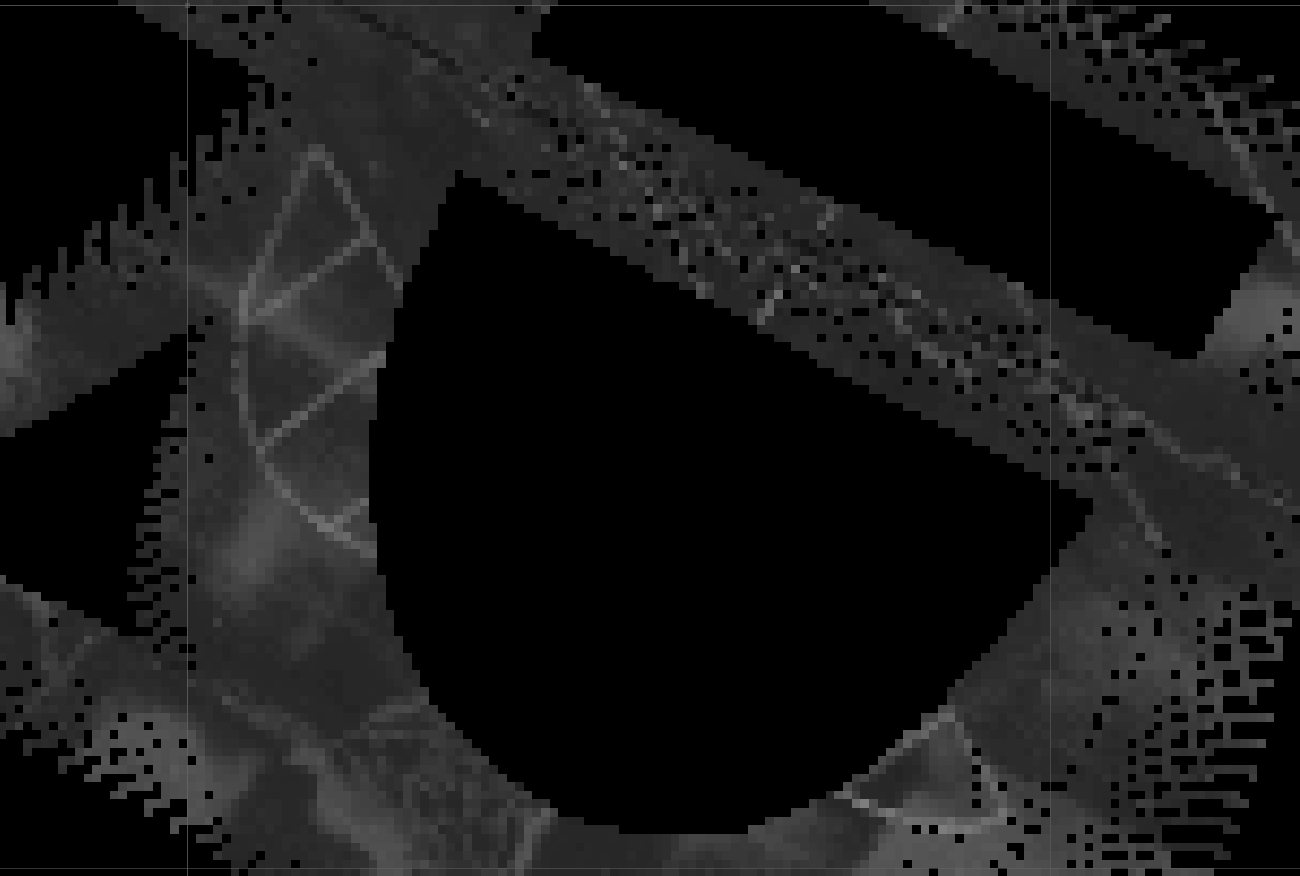}}
	\subfloat[Map-perspective of laser 16]{\includegraphics[width=0.24\linewidth]{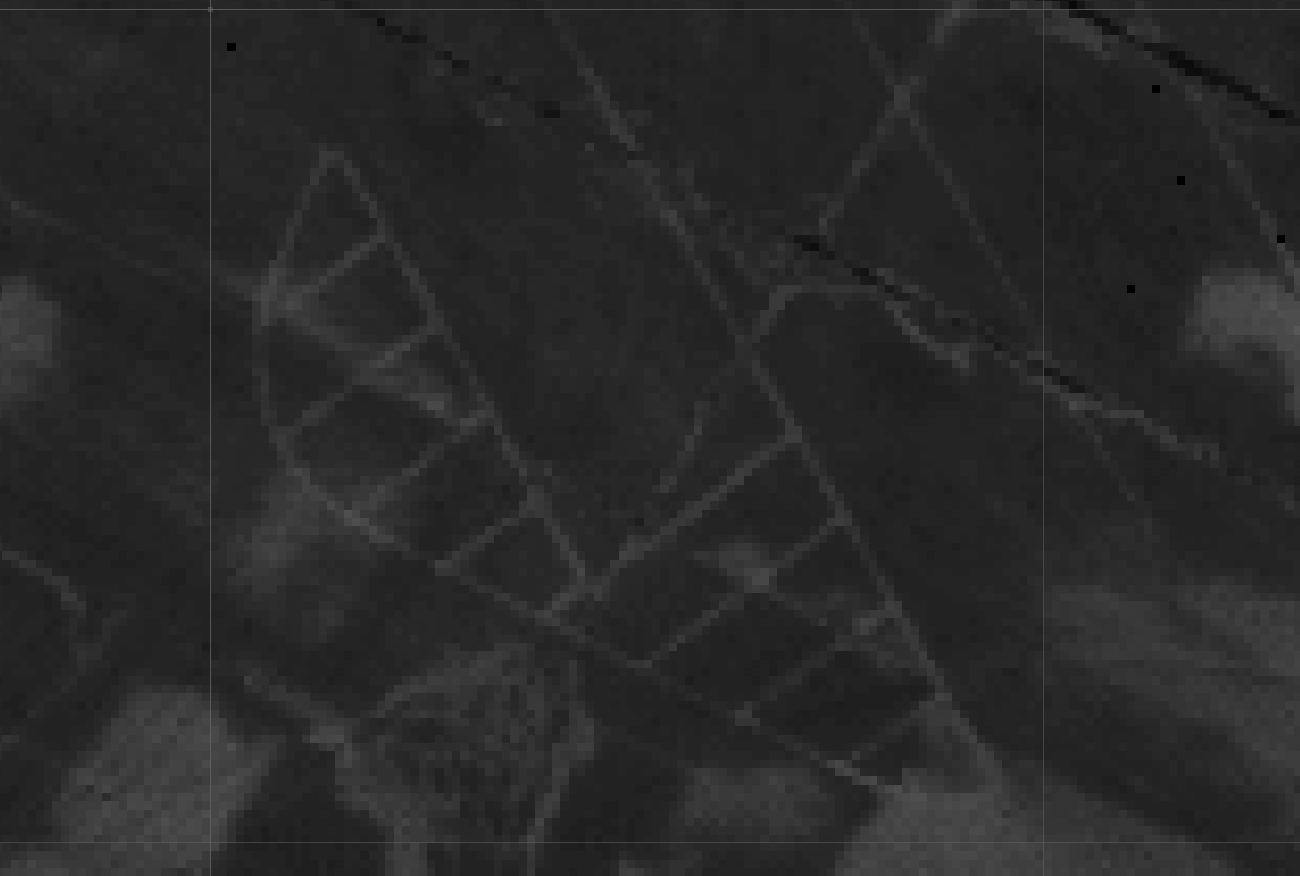}}
	\subfloat[Map-perspective of laser 19]{\includegraphics[width=0.24\linewidth]{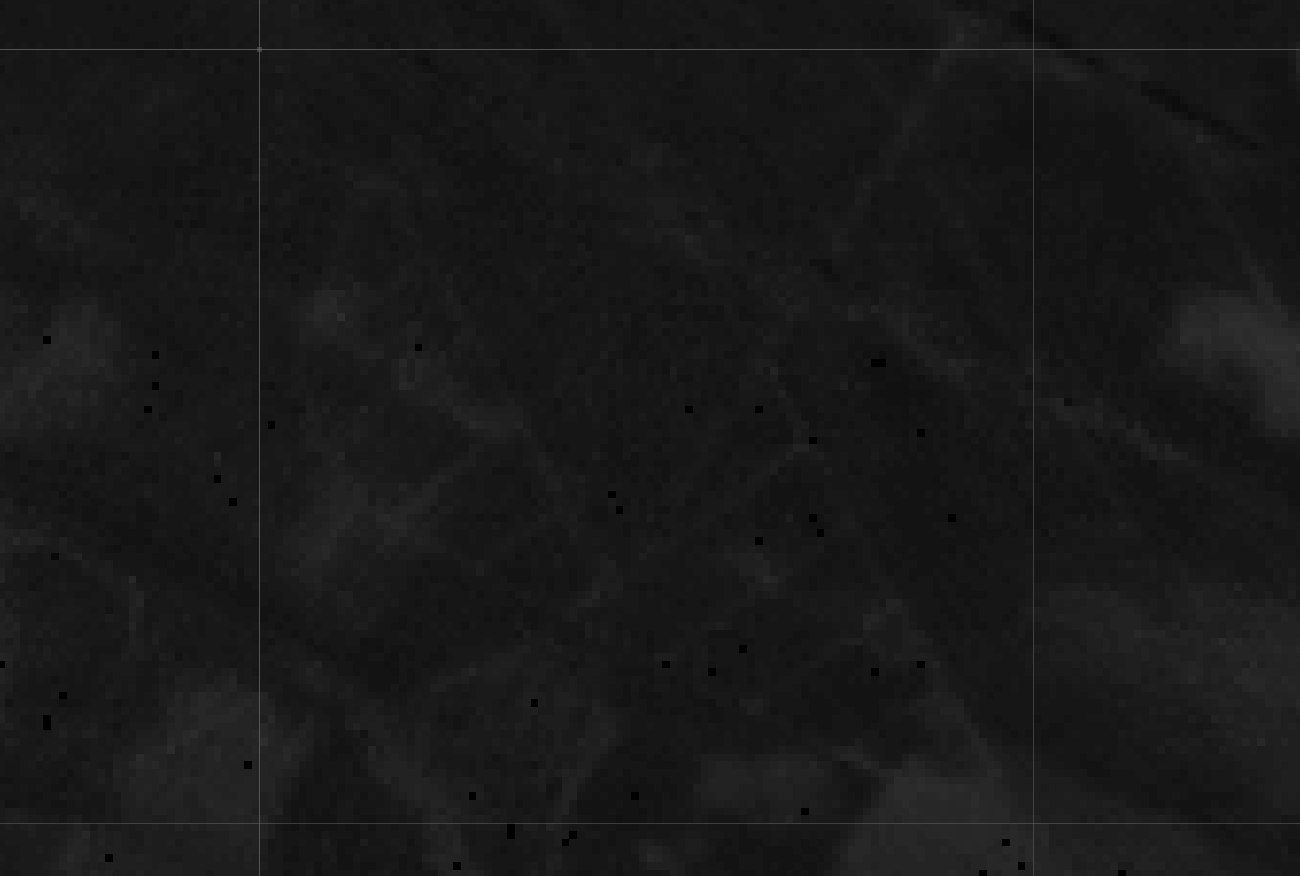}}
	\subfloat[Map-perspective of laser 21]{\includegraphics[width=0.24\linewidth]{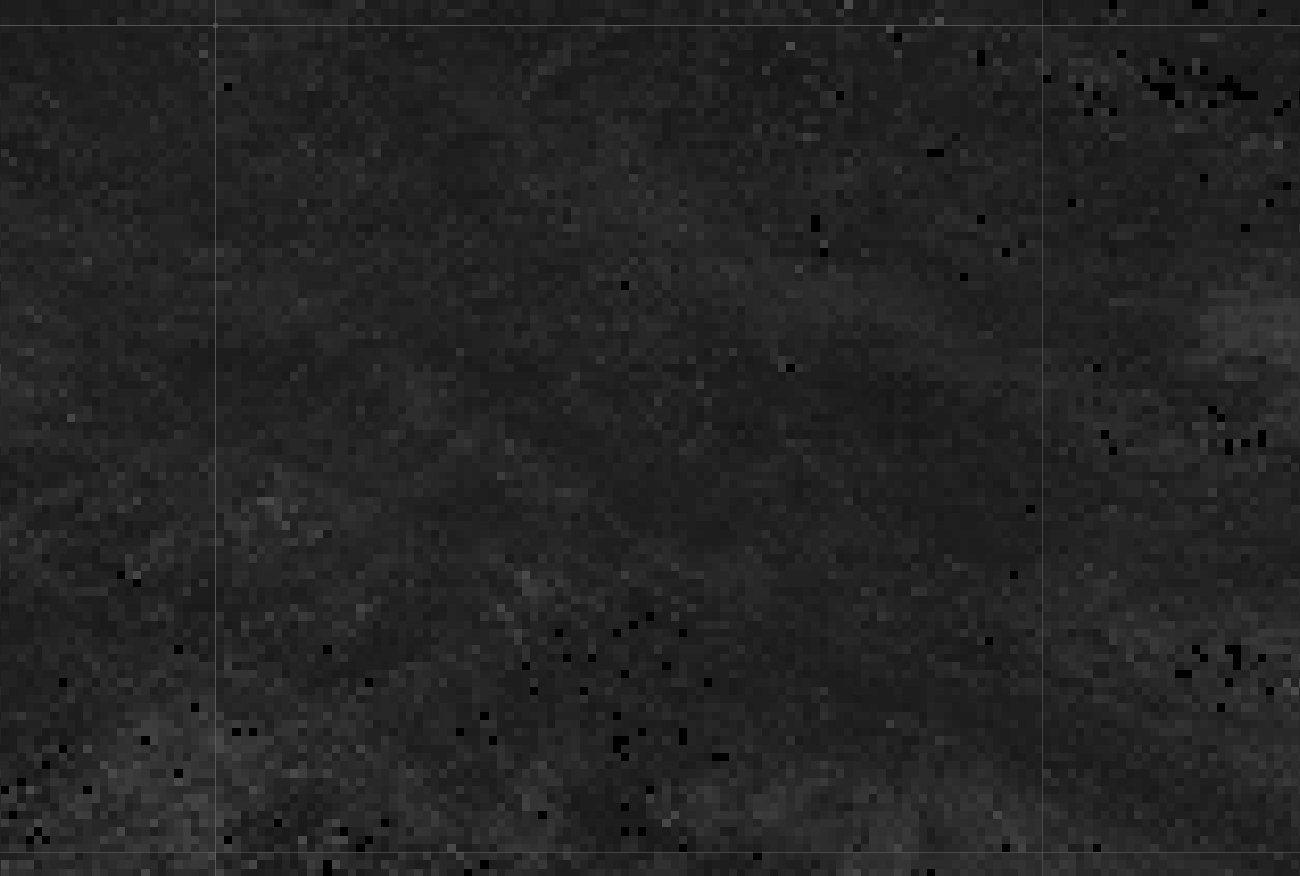}}
	\caption{Ortographic view of patches of the ground reflectivity from the perspective of a given laser (indexed by order from bottom to top) in a Velodyne HDL-32E. Patches obtained while mapping with a vehicle in motion.  
		(a) View from the perspective of laser 1, (b) View of beam 16: reduced contrast and full view of patch domain, (c) View of beam 19: showing bad contrast characteristic of high ranges and angles of incidence, (d) View of laser 21 presents the worst contrast.}
	\label{fig:intensityViewedBeam}
\end{figure*}

The laser reflectivity measurement model we use is the white Gaussian additive noise (AWGN) model
\begin{equation} \label{measurementModel}
y^{\phibf} =  \underbrace{f^b(x, \phibf)}_{x^{\phibf}} + \epsilon^{\phibf}	
\end{equation}
where $ f^b: \R \rightarrow \R$ is the non-linear response of laser with index $b$ at $\phibf$ to a surface of reflectivity $x \in \R$, $ x^{\phibf} \in \R$ is the surface reflectivity from the perspective of $\phibf$ and $\epsilon^{\phibf} \sim \Ncal(0, \sigma_{\phibf})$ is an i.i.d. random variable independent of $f^b$. The set $\phibf$ indicates the laser observer and its perspective parameters; for example, $\phibf = (b, \theta, r)$ indicates the laser with global index $b$ at an angle of incidence $\theta $ and range $r$. %Note here that $b$ can represent a global index of all the laser observers in the multiple laser scanners used while mapping. 
The set and number of globally projected (using full SLAM \cite{Durrant-Whyte.Bailey2006, Bailey.Durrant-Whyte2006} for example) reflectivity measurements of the ground at cell $n$ taken from the perspective of $\phibf$ are denoted with $\yphin \in \R^{M_n^{\phibf}}$ and $M_n^{\phibf}$, respectively.
%When a robot/vehicle is surveying a region, each reflectivity measurement of the ground is associated with a cell in the map depending on the projection of its corresponding range into the global reference frame (assuming full SLAM \cite{Durrant-Whyte.Bailey2006, Bailey.Durrant-Whyte2006} has already taken place). 
%These projected measurements are non-uniformly distributed across the ground depending on the sensor configuration, the scanning pattern and the navigation path followed by the robot/vehicle while surveying. Here, we denote the set of reflectivity measurements taken from the perspective of $\phibf$ at cell $n$ with $\yphin \in \R^{M_n^{\phibf}}$, where $M_n^{\phibf}$ is the number of recorded reflections. 
At a fixed $\phibf$, these measurements follow the AWGN model in \eqref{measurementModel} with the proper extension of terms, for example, the noise term follows $ \pmb{\epsilon}_{n}^{\phibf} \sim \Ncal(\mathbf{0}_{ M_n^{\phibf} \times M_n^{\phibf} }, \sigma_{\phibf} \cdot \I_{M_n^{\phibf} \times M_n^{\phibf}} )$. From these measurements, we generate a set of map-perspectives $\Yphibf \in \R^{N_x \times N_y} \cup \{ \infty\}^{N_x \times N_y}$ which represent the map of ground reflectivity from the perspective of $\phibf$. These are defined element-wise as
\begin{equation} \label{perspectives}
[\Yphibf]_n = 
\begin{cases} 
\frac{1}{\Mphin} \sum\limits_{m = 1}^{\Mphin} [\yphin]_m, & \mbox{for} \quad n \in \Omega^{\phibf} \\
\quad\quad\quad \infty & \mbox{for } \quad n \notin \Omega^{\phibf}
\end{cases}
\end{equation}
for all $\phibf \in \Phibf$ with $\Phibf = \{ \phibf_1, \phibf_2,...\phibf_B \}$ including as elements all of the possible perspective parameters from which measurements were collected and $ B = |\Phibf| < \infty$ represents the total number of different perspectives from all observers. Examples of map-perspectives are shown in Figure \ref{fig:intensityViewedBeam} where the non-black regions represent the cell map occupancy $\Ophi$ of measurements from $\phibf$. The complete cell map occupancy or domain $\O' \subset \R^2$ is given by the union
%
%\begin{equation}
$	\O' = \bigcup_{\phibf \in \Phibf} \Ophi  $
%\end{equation} 
%
of the map-perspective individual occupancies $\Omega^{\phibf}$.
Note that by generating these map-perspectives a normalization to unbias for the non-uniform number of measurements from an observer and its perspectives in cells is carried out.
Also, note that the uncertainty of the elements $[\Yphibf]_n$ is now distributed as $ \sim \Ncal(0, \sigma_{\phibf} /M_n^{\phibf} )$, essentially reducing noise at each cell by a factor of $ M_n^{\phibf} $. %In \eqref{perspectives}, the set $\Omega^{\phibf}$ represents the cell map-occupancy of measurements from $ \phibf$. %To define the complete cell occupancy or domain of the map, we let the map domain $S$ be a closed subset of $\R^2$ and let $\O'$ be the subset of $S$ where there are laser reflectivity measurements available. Thus, the map domain $\O'$ is given by the union 

\begin{figure*}[t]
	\centering
	{\includegraphics[width = 13 cm]{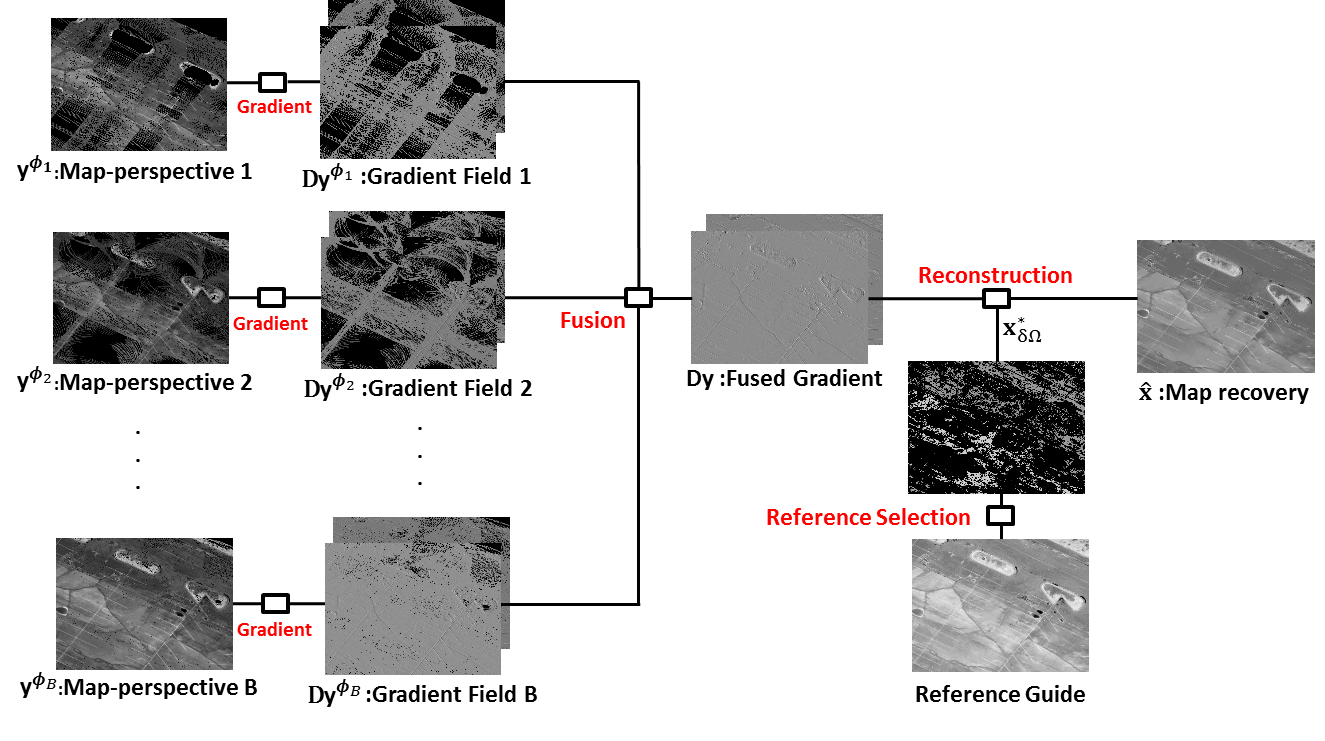}}
	\caption{Schematic of our new method. The input images in the left column represent the map-perspectives obtained by using \eqref{perspectives}. The gradient field for each of these map-perspective images is then computed in the second column, followed by application of \eqref{fusionFunction} and \eqref{reconstruction} within the fusion and reconstruction blocks, respectively.}
	\label{fig:schematic}
\end{figure*}

The problem of operating on these map-perspectives is that although these 
have overlapping observations of the same regions (i.e., overlapping occupancies), its perspective influences the characteristic reflectivity response. To decouple these from the perspective parameters we apply the gradient to each of them. Here, we denote the discrete gradient by $ \g : \mathbb{R}^N \rightarrow \mathbb{R}^{N \times 2}$ composed of the matrices $ \g_x : \mathbb{R}^N \rightarrow \mathbb{R}^{N}$ and $ \g_y : \mathbb{R}^N \rightarrow \mathbb{R}^{N}$ that represent the first order finite forward difference along the horizontal and vertical axes, respectively. Under this representation, the gradients of map-perspectives components $[\gYphibf]_n$ become statistically stationary which allows one to operate on each invariantly. The algorithmic operators here proposed are those of selection, de-noising and fusion which are compactly represented via the function
\begin{align} \label{fusionFunction}
\widehat{\gYbf} = f_{\mbox{fusion}}( \Ybf^{\phibf_1}, \Ybf^{\phibf_2},..., \Ybf^{\phibf_B} ).
\end{align}
where $\widehat{\gYbf} \in \mathbb{R}^{N \times 2}$ denotes an estimate of the gradient of fused map-perspectives of $ \Ybf$.
The goal of reconstruction on the other hand is to recover a map estimate $\Xest \in \R^{N_x \times N_y}$ of the ground reflectivity consistent with the fused gradients of map-perspectives and a reference measured data $\Xbf^*$. Here, we compactly represent this reconstruction process by
\begin{align} \label{reconstruction}
\Xest = f_{\mbox{reconstruction}}( \gYbf, \Xbf^* ).
\end{align}
Overall, our method is summarized in Figure~\ref{fig:schematic}. 

In addition to this general formulation, we include here a few definitions. First, the sampling matrix $\I_{\Omega} \in \R^{|\Omega| \times N} $ is the submatrix of the identity obtained by extracting the rows indexed by $\Omega $. This matrix is defined element-wise by
\begin{equation} \label{sampler}
[ \I_{\Omega} ]_{i,j} =  
\begin{cases} 
1 \quad i \in \{ j \cap \Omega \} \\ 
0 \quad \mbox{ otherwise }
\end{cases}.
\end{equation} 
Thus, a sampled vector $\Xbf_{\Omega} \in \R^{ | \Omega|}$ is the result of computing $ \Xbf_{\Omega} = \IO \Xbf $. Second, the projection operator $ \P_{\Omega}: \R^{N} \rightarrow \R^N$ projects inputs into the domain $\Omega$. %This operator is equal to the identity except at the diagonal entries which are not in the domain $\Omega$ in which case these are set to zero. In other words, the projection operator is applied element-wise as
Application of this operator is performed element-wise as
\begin{equation} \label{projection}
[\P_{\Omega} \Xbf]_n =  
\begin{cases} 
[\Xbf]_n \quad n \in \Omega \\ 
0 \quad \mbox{ otherwise }
\end{cases}.
\end{equation} 
Finally, for ease of notation, every element-wise operation that involves an $\infty$ value will be determined by ignoring it from its computation. For example, 
\begin{equation} \label{equivalence}
\sum\limits_{\phibf \in \Phibf} [\Yphibf]_n = \sum\limits_{\phibf \in \Phibf} [\Yphibf]_n \cdot \textbf{1} \{ [\Yphibf]_n < \infty \}.
\end{equation}
%

%%%%%%%%%%%%%%%%%%%%%%%%%%%%%%%%%%%%%%%%%%%%%
%% Subsection 2
%%%%%%%%%%%%%%%%%%%%%%%%%%%%%%%%%%%%%%%%%%%%%

\subsection{ Sparse selection, Denoising and Fusion} 
\label{Sec:SensorFusion}

\begin{algorithm}[t]
	\caption{ Fusion of the gradient field with \eqref{fusion} } \label{gradientFusionAlgorithm}
	\begin{algorithmic}[1]
		\BState \textbf{input:} Map-perspective gradient $ \gYphibf $ for $ \phibf \in \Phibf$, step $\gamma > 0$, and thresholding constant $\tau>0$.
		\State \textbf{set: }  $t \leftarrow 1$,  $q_0 \leftarrow 1$ %, $\gYbfe \leftarrow \sum\limits_{\phi = 1}^B | \nabla \Yphibf | $ 
		\BState \textbf{repeat} %$T = T_{init}$ \mbox{through}  $T = T_{final}$
		\State \quad $\pmb{s}^t \leftarrow \eta_{\tau}(\pmb{s}^{t-1} - \gamma \nabla \Dcal_1( \pmb{s}^{t-1}) ) $
		\State \quad $ q_t \leftarrow \frac{1}{2} \left ( 1 + \sqrt{1+4 q^2_{t-1}} \right )$ 
		\State \quad $\wbf^t \leftarrow \pmb{s}^t + ( (q_{t-1} -1)/q^t) ( \pmb{s}^t - \pmb{s}^{t-1}) $.
		\State \quad $ t \leftarrow t + 1$
		\BState \textbf{until: } stopping criterion
		\State \textbf{return: } Fused gradient $ \gYbfe = \sum\limits_{\phibf \in \Phibf} \w^t_{\phibf} \cdot g_d(\gYphibf) $
	\end{algorithmic}
\end{algorithm}

We propose here a fusion model of the map-perspectives in the gradient domain. This model consists in the weighted average of the map-perspective gradients $ \gYphibf$ obtained from each of the available $\Yphibf$ (with $\infty$ values ignored as the example in \eqref{equivalence}). In other words, fusion is carried out in the gradient domain by 
\begin{equation} \label{fusion}
\widehat{\gYbf} = \sum_{\phibf \in \Phibf} \w_{\phibf} \cdot g_d(\gYphibf).
\end{equation}
where $ g_{d}: \R^{N \times 2} \rightarrow \R^{N \times 2} $ can be any function including identity.
The vector of weights $\wbf \in \R^{ |\Phibf|}$ is chosen here to be a sparse vector that selects map-perspectives $\Yphibf$ via its non-zero entries. In other words, we will find the best sparse representation of an overall averaged isotropic gradient of map-perspectives. The reason for this is that some map-perspectives should be penalized or even removed from the reconstruction process as illustrated in Figure \ref{fig:intensityViewedBeam}. To achieve this, we find the weights that minimize the isotropic sparse promoting least squares optimization problem
\begin{equation} \label{weightOptimization}
\wbfe = \arg \min\limits_{ \wbf \in \mathcal{W} } 
\left \{  
\Dcal_1(\wbf) + \lambda \Rcal(\wbf)
\right \},
\end{equation}
where minimization is carried out over a fidelity term $\Dcal_1: \R^{|\Phibf|} \rightarrow \R$ and a regularization term $\Rcal: \R^{|\Phibf| } \rightarrow \R$ controlled by the parameter
$\lambda > 0$ which determines the sparseness strength. In this problem the fidelity term is defined by
\begin{equation} \label{l2Fidelity}
\Dcal_1(\wbf) =  \frac{1}{2} 
\left \| \sum\limits_{\phibf \in \Phibf} | \gYphibf | - 
\sum\limits_{\phibf \in \Phibf} \w_{\phibf}| \gYphibf | \right \|_{\ell_2}^2 
\end{equation}
which promotes consistency between the weighted average and the average of map-perspective gradients. The second regularization term
\begin{equation} \label{l1Regularizer}
\Rcal(\wbf) = \| \wbf \|_{\ell_1}
\end{equation}
promotes sparse solutions. Adjustment of the parameter $\lambda$ can be useful to control the number of map-perspectives $\Yphibf$ that will go into the fusion process and map reconstruction. 

The problem in \eqref{weightOptimization} involving the non-convex functional in \eqref{l1Regularizer} can be solved iteratively using a proximal gradient algorithm. This involves the alternate application of a gradient-descent step on the fidelity term in \eqref{l2Fidelity} followed by the application of the non-linear soft-thresholding operator $\eta_{\tau}(\cdot): \R^N \rightarrow \R^N $ defined element-wise by
\begin{equation} \label{soft}
[\eta_{\tau}(\Xbf) ]_n = \mbox{sgn} ( [\Xbf]_n ) (|[\Xbf]_n | - \tau )_+
\end{equation}
to promote sparsity of the solution at each iteration. The accelerated implementation of this algorithm is known as fast iterative shrinkage thresholding algorithm (FISTA) ~\cite{Beck.Teboulle2009a} which converges with a rate of $O(1/k^2)$. A summary of the selection algorithm is given in Algorithm \ref{gradientFusionAlgorithm}.

%Fusion of the map-perspectives gradients could be further improved by using a de-noising algorithm on each gradient component. Under this scenario we would modify \eqref{fusion} and apply a denoising function to the gradient field term as
%%
%\begin{equation} \label{denoisedFusion}
%	\widehat{\gYbf} = \sum_{\phibf \in \Phibf} \w_{\phibf} \cdot g_{d}( \gYphibf ),
%\end{equation}
%%
%where $ g_{d}: \R^{N \times 2} \rightarrow \R^{N \times 2} $ is again the non-linear soft-thresholding denoising operator \cite{Donoho95} applied independently to each of the horizontal and vertical gradient components. Application of this denoising function is summarized in Algorithm \ref{gradientFieldDenoising} in the Appendix. In the imaging community, this denoising idea is referred to as total variation \cite{Rudin.etal1992}.
In \eqref{fusion}, we include the function $ g_{d}$ and propose the option to use it for denoising using the non-linear soft-thresholding operator \cite{Donoho95} applied independently to each of the horizontal and vertical gradient components. The basis of this idea can be found in \cite{Rudin.etal1992} and our proposed implementation is summarized in Algorithm \ref{gradientFieldDenoising} in the Appendix.  

%%%%%%%%%%%%%%%%%%%%%%%%%%%%%%%%%%%%%%%%%%%%%
%% Subsection 2
%%%%%%%%%%%%%%%%%%%%%%%%%%%%%%%%%%%%%%%%%%%%%

\subsection{Map reconstruction}
\label{Sec:Reconstruction}

The next step consists of the reconstruction of the map of ground reflectivity based on the output of the fusion process. This step is required since we are interested in reconstructing a map of reflectivity and not of gradients. Specifically, we use $ \widehat{\gYbf}$ from \eqref{fusionFunction} to compactly denote an estimate of the fused gradient of map-perspectives. Reconstruction is then carried out by minimizing a cost function on $\Xbf$ based on Poisson's formulation \cite{Perez.etal03}. Here, this problem is posed as an $\ell_2$ term promoting consistency with the fused gradient of map-perspectives and a term that ensures equality constraints with reference measurement data on a subset $\dO \subset \O'$ of cells of $\Xbf$. %This definition of $\dO$ is used for consistency with~\cite{Perez.etal03}, although this set does not necessarily represent a boundary. 
Based on this descriptions the map of ground reflectivity can be recovered by solving %the following optimization
\begin{equation} \label{LaplaceNeumman}
\Xest = 
\argmin_{ \Xbf \in \mathcal{X} } 
\left\{  	  \frac{1}{2} \sum_{k = \{x,y\} } \left\| \gkYbfe - \gkXbf  \right\|^2_{\ell_2} \right\},
\text{  s.t.   }
\Xbf_{\dO} = \Xbf_{\dO}^*  
\end{equation}
The minimizer of \eqref{LaplaceNeumman} is the unique solution to the equivalent Poisson equation with Dirichlet boundary conditions problem posed in matrix form as:
\begin{equation} \label{PoissonEquation}
\IO \L	\Xbf = \IO[ \g_x^2 \Ybf + \g_y^2 \Ybf ], \quad 
\mbox{s.t.} \quad
\Xbf_{\dO} = \Xbf_{\dO}^*  
\end{equation}
where $ \L: \R^{N} \rightarrow \R^{N} $ is a discrete Laplacian operator defined in \eqref{Laplacian} in the appendix. To simplify this problem, we recast \eqref{PoissonEquation} in a form such that optimization is carried out only on the entries in the domain of $ \Omega = \dO^c$ representing the cell locations where reconstruction is to be performed, since we already know $ \xbf_{\dO} $. To accomplish this, we use the projection and sampling operators defined in \eqref{projection} and \eqref{sampler}, respectively, and rewrite \eqref{PoissonEquation} as the minimization problem:
\begin{equation} \label{PoissonRearranged}
\Xest_{\Omega} = 
\argmin_{ \Xbf_{\Omega} \in \Xcal_{\Omega} } 
\left\{  \Dcal_2( \Xbf_{\Omega} ) \right\},
\quad \text{  and   }  \quad 
\Xest_{\dO} = \Xbf_{\dO}^*  
\end{equation}
where $ \Dcal_2 : \R^{ | \Omega |} \rightarrow \R $ represents the fidelity term that satisfies \eqref{PoissonEquation} over the domain $\Omega$. In other words,
\begin{equation} \label{LaplaceNeummanFidelity}
\Dcal_2(\Xbf) =
\frac{1}{2}
\left \| \pmb{\Phi} \Xbf - \mathbf{b}  
\right \|^2_{\ell_2}
\end{equation} 
with $\pmb{\Phi}$ representing the $\Omega$ projected and sampled Laplacian of the left hand side of the first equality in \eqref{PoissonEquation} and $\mathbf{b}$ represents its right side with incorporation of the Laplacian projection of the second equality contraint. Mathematically, these two terms are defined via  
\begin{equation}
\pmb{\Phi} =  \IO \L \PO	\IO^T \quad \mbox{and} \quad 
\mathbf{b} = \I_{\Omega} [\g_x^2 \Ybf + \g_y^2 \Ybf -\L \P_{\dO} \Xbf^*]
\end{equation}
where the superscript $^T$ denotes a matrix transpose. Since the cost functional in \eqref{LaplaceNeummanFidelity} is convex, we propose to use the accelerated gradient projection method of Nesterov~\cite{Nesterov1983} which achieves a rate of convergence of $O(1/k^2)$. This iterative method is summarized in Algorithm \ref{priorMapReconstruction}.

\begin{algorithm}[t]
	\caption{ Map reconstruction of ground reflectivity } \label{priorMapReconstruction}
	\begin{algorithmic}[1]
		\BState \textbf{input:} Fused gradient $\gYbf$, reference $ \xbf_{\dO}^*$ and $\gamma > 0$.
		\BState \textbf{set: }  $ t \leftarrow 1$		
		\BState \textbf{repeat}
		\State \quad $\pmb{s}^t \leftarrow \pmb{s}^{t-1} - \gamma \nabla \Dcal_2( \pmb{s}^{t-1})  $
		\State \quad $ q_t \leftarrow \frac{1}{2} \left ( 1 + \sqrt{1+4 q^2_{t-1}} \right )$ 
		\State \quad $\Xbf_{\Omega}^t \leftarrow \pmb{s}^t + ( (q_{t-1} -1)/q^t) ( \pmb{s}^t - \pmb{s}^{t-1}) $.
		\State \quad $ t \leftarrow t + 1$
		\BState \textbf{until: } stopping criterion
		\BState \textbf{set:  } $ \Xest_{\dO} \leftarrow \xbf_{\dO}^* $, $ \Xest_{\Omega} \leftarrow \Xbf_{\Omega}^t $
		\State \textbf{return: } Map of ground reflectivity $ \Xest $
	\end{algorithmic}
\end{algorithm}

To define $ \Xbf_{\dO}^* $ and the set $\dO$ there are several alternatives that could be used. Here, the method we propose simply consists on choosing a map-perspective $ \Ybf^{\phibf_r}$ with $ \phibf_r \in \Phibf$ characterized with a strong gradient compared to others within some region (e.g., $|[\Dbf(\Ybf^{\phibf_r})]_n| > |[\Dbf(\Ybf^{\phibf})]_n|;$ for some $n \in \Omega^{\phibf_r} \cap \Omega^{\phibf}; \forall \phibf \neq \phibf_r$ if it exists). From the map-perspective $ \Ybf^{\phibf_r}$, we extract the subset of cells indexed by $\dO$ and use the reflectivity values in the corresponding cells to construct our reference as $ \Xbf_{\dO}^* = \Ybf^{\phibf_r}_{\dO}$. %In other words, once $\dO$ is defined we set $ \Xbf_{\dO}^* = \Ybf^{\phibf_1}_{\dO}$ and set $ \Xbf_{\Omega}^* $ to any value, since the later cells are not used in the optimization \eqref{LaplaceNeummanFidelity}. 
The set $\dO$ is determined from $ \Ybf^{\phibf_r}$ as the cells with the most-likely reflectivity value, in other words
$
\dO  = \{n | [\Ybf^{\phibf_r}]_n = %\argmax\limits_{y_n} \prod\limits_{n = 1}^{N} p_{y_n}( [\Yphibf ]_n )
\argmax\limits_{y} p_{\Ybf^{\phibf_r}}( \Ybf^{\phibf_r} )
\}	. 
$

 		% Proposed Approach

\section{Experiments}
\label{Sec:Experiments}

% Sensor suit
To verify our approach, we use mapping datasets from \cite{Pandey.McBride.Eustice2011} and a few additional we collected with Ford's Fusion testing fleet of autonomous vehicles. These are outfitted with four Velodyne HDL-32E 3D-LIDAR scanners and an Applanix POS-LV 420 (IMU). The four HDL-32E's were mounted on the roof above the B pillars of the vehicle; two which rotate in an axis perpendicular to the ground (i.e., non-canted) and two that are canted. %Time-registered Velodyne datasets are stored which contain points of Cartesian coordinates from which a LIDAR pulse reflection has been recorded, a reflectivity measurement for each of these points and the laser beam information. 
Extrinsic calibration of the LIDAR's is performed using GICP \cite{Segal.Haehnel.Thrun2009}. %This algorithm iteratively optimizes for the 6 DOF parameters that align the overlapping range scans.
When mapping, all LIDAR measurements are projected into a global reference frame using the vehicle pose from the GPS, IMU and odometry with a post-correction by full-SLAM \cite{Kaess.Ranganathan.Dellaert2008, Kaess.etal2012}.
%hich maximizes the likelihood of the map and vehicle positions given the LIDAR and position measurements.
%
%Data used in our experiments 
The datasets we used were all collected at the Ford campus in Dearborn, Michigan. To test our approach we use only data from the two non-canted LIDARs. Under this configuration, the angle of incidence and range per laser recorded from the ground remains approximately constant over the course of the 360\textdegree scan trip. Thus, each laser index implicitly accounts for these two which then results in a total of $B = 64$ map-perspectives. %Note that when running our algorithm we fix the map to any orthographic view of the 2-D ground-plane and do all the processing with this view fixed. 
%
% Data used for comparing against our own implementation of an algorithm.
To compare against state of the art, we test against an implementation of the deterministic reflectivity calibration method in \cite{Levinson10,Levinson.Thrun2010}. The data we used in this case is collected from all four LIDARs. The two main differences in our implementation are: (1) we make an approximation to generate the look-up table in one data pass and (2) we include an angle of incidence dependency for the canted LIDARs. We would like to mention that the calibration and map generation dataset is the same thus ensuring the best possible map results. %Thus, the quality of the maps resulting from this method will be as good as they can be.  

%Given that maps of the ground can be very extensive, 
The implementation of our method is applied independently on disjoint patches of the map of size $400 \times 400$ cells with $10 \times 10$ cm cells. %However, our method is flexible enough to use other sizes as long as the gradient field is spatially informative. 
First, the search spaces are defined by $\Xcal = \{ \Xbf \in \R^{N}: 0 \leq \Xbf \leq 255 \}$, $\Xcal_{\Omega} = \{ \Xbf_{\Omega} \in \R^{|\Phibf|}: 0 \leq \Xbf_{\Omega} \leq 255 \}$, $\Xcal_{\g} = \{ \g_k \Xbf \in \R^{N}: -255 \leq \g_k \Xbf \leq 255 \}$. Second, we find that Algorithm \ref{gradientFusionAlgorithm} converges in $K \leq 100$ iterations with $\lambda = 1.2e-3 $, $ \gamma = 1e^{-3}$ and $ \tau = 2.3e^{-3} $. %Under these settings, approximately $5/64$ map-perspectives are selected to sparsely represent the overall fused gradient field. 
In the denoising Algorithm \ref{gradientFieldDenoising} we use the same parameter settings as in Algorithm \ref{gradientFusionAlgorithm}. For the reconstruction in Algorithm \ref{priorMapReconstruction}, we found that the relative energy step in two successive iterations $\| x_{\Omega}^t - x_{\Omega}^{t-1} \|_{\ell_2} / \|  x_{\Omega}^{t-1} \|_{\ell_2} < 1e^{-3}$ or $ K \leq 256 $ iterations is sufficient to yield good reconstructions. To give an intuition of the run-times, it took 45 secs to reconstruct a patch of size $40 \times 40$ meters. %In the following, we include examples that show how our method outperforms the quality of existing ones.

\subsection{Contrast enhancement}
\label{sec:enhancement}

\begin{figure}[t]
	\centering
	\subfloat[Raw na\"{\i}ve fusion]{\includegraphics[width=0.5\linewidth]{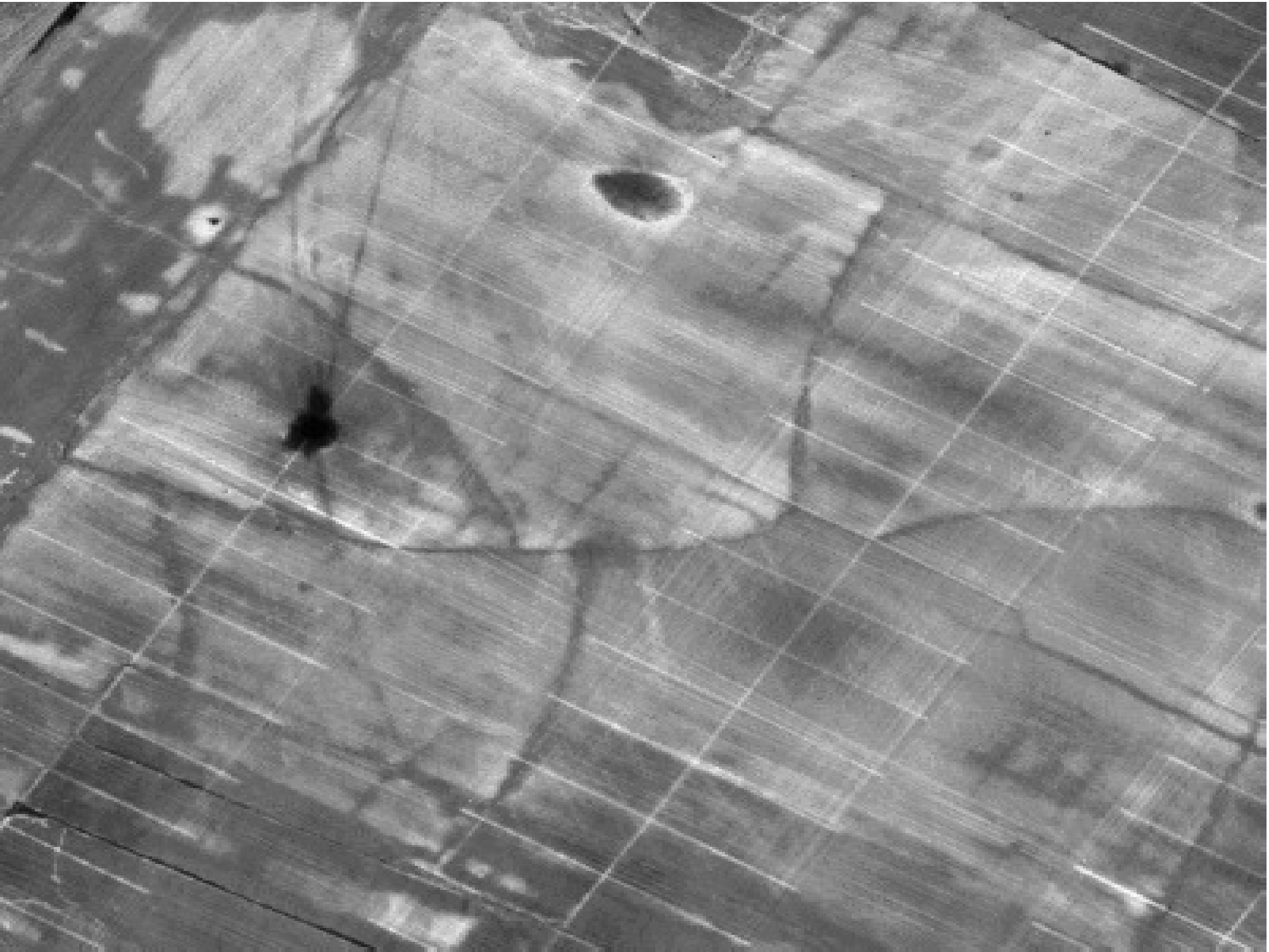}}
	\subfloat[Our method]{\includegraphics[width=0.5\linewidth]{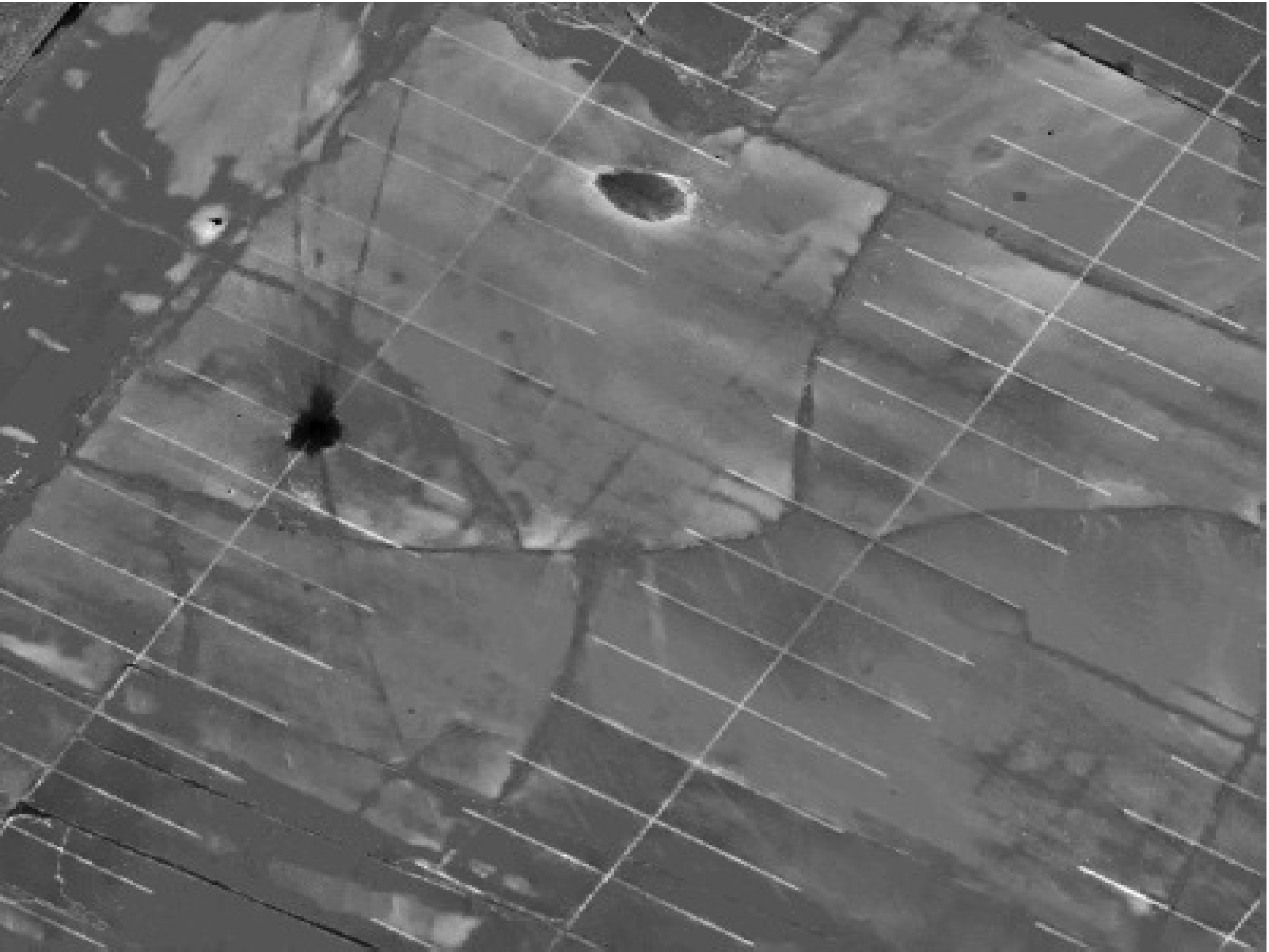}}
		
	\subfloat[Reflectivity calibration based]{\includegraphics[width=0.5\linewidth]{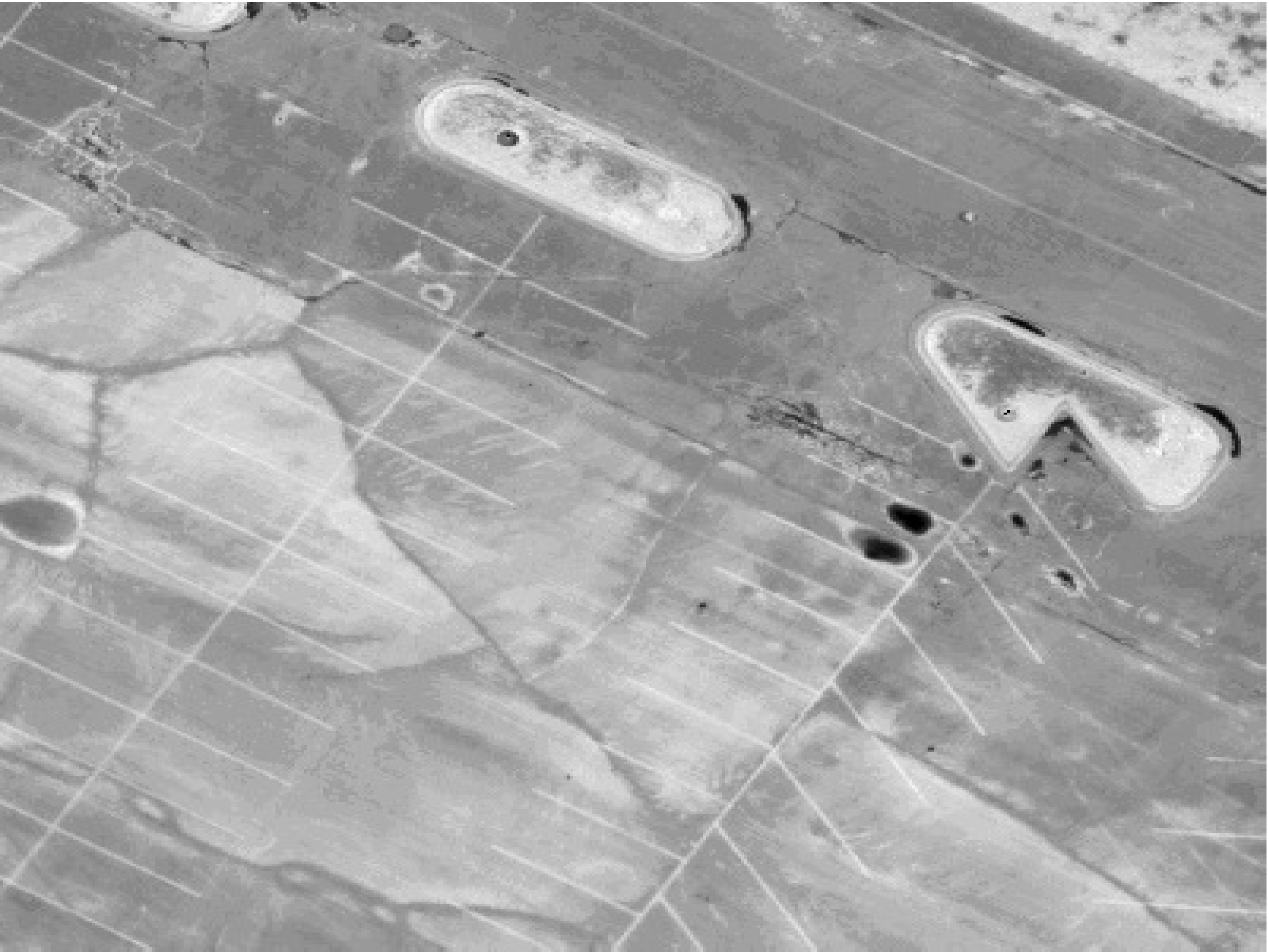}}
	\subfloat[Our method]{\includegraphics[width=0.5\linewidth]{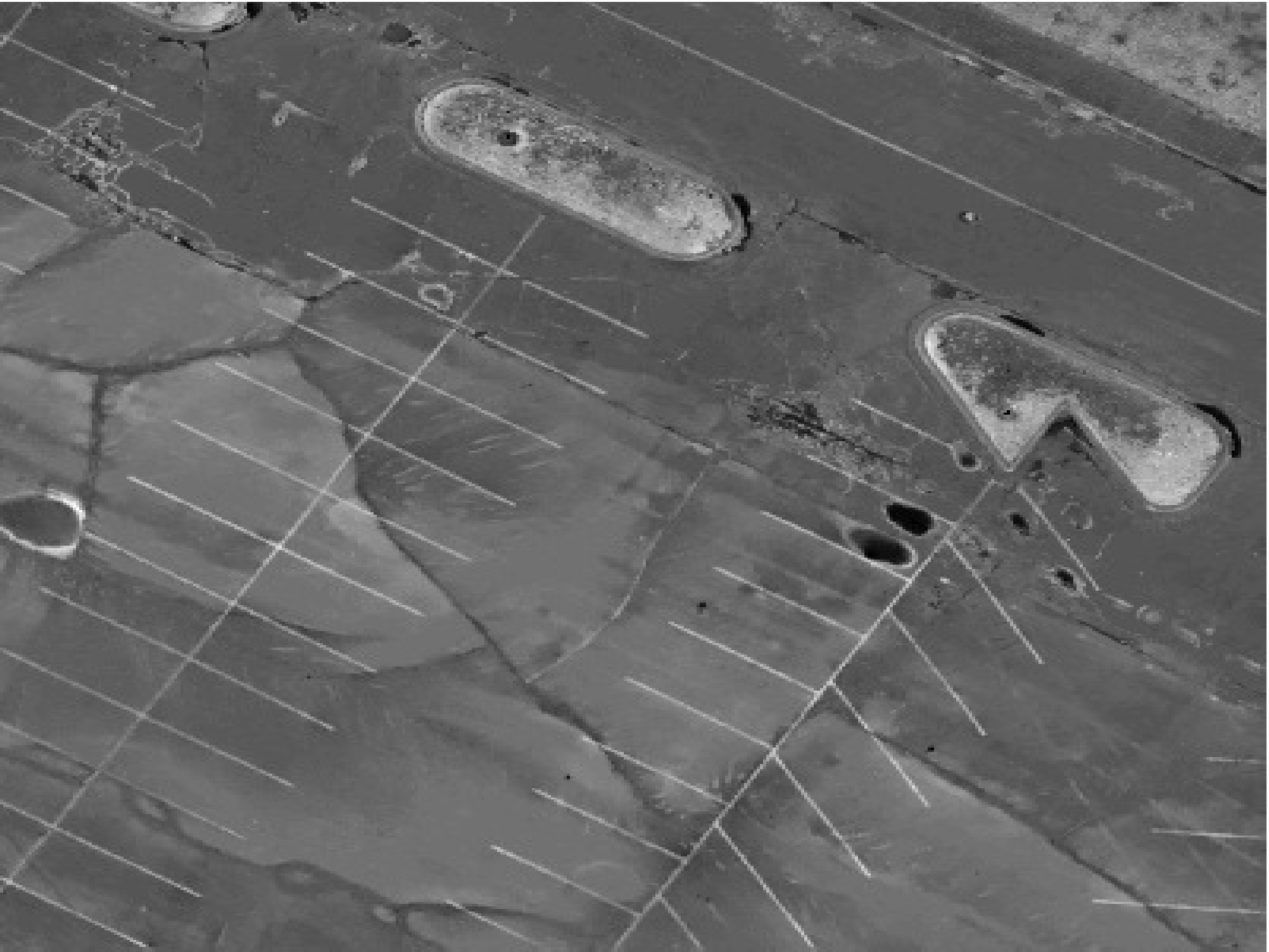}}
	
%	\subfloat[Patch of map generated from calibrated measurements]{\includegraphics[width=0.506\linewidth]{figures/Patch12Calib}}
%	\subfloat[Our method]{\includegraphics[width=0.5\linewidth]{figures/Patch12SoftTopNoDenoising}}
	%
	\caption{Contrast enhancement in map patches with ice on the ground. Note: gray scale value represents reflectivity. \textbf{Zoom in to better appreciate image quality.}}
	\label{fig:contrastEnhancement}
\end{figure}

In this subsection we show that our method results in maps of enhanced contrast. Figure~\ref{fig:contrastEnhancement} includes a comparison of map patches between raw na\"{\i}ve fusion (i.e., average of the raw reflectivity measurements) in  Figure~\ref{fig:contrastEnhancement}.a, our implementation of state of the art calibration \cite{Levinson10} in Figure~\ref{fig:contrastEnhancement}.c and the result  of our approach applied to the two corresponding cases in Figure~\ref{fig:contrastEnhancement}.b and d. Note that our approach yields in general significantly better contrast (e.g., smoother regions with sharper edges). Even some of the edges from lane markings which were close to being indistinguishable appear more visible with our method.

\subsection{Artifact removal}
\label{sec:articactRemoval}

We have found that under certain driving conditions the non-uniform laser reflectivity response results in map patches corrupted with artifacts.  
%In certain circumstances that occur frequently we have found that the conditions under which the surveying vehicle navigates impacts the map generation by means of the appearance of artifacts in the map. 
These have been observed also in other works, for example even after applying the reflectantance calibration in \cite{Levinson.Thrun2010, Levinson.Thrun2013}. In this section, we include examples in which we show three types of artifacts. %To illustrate the appearance of these artifacts, we include maps obtained by averaging the raw laser reflectivity measurements (i.e., raw na\"{\i}ve fusion).

\begin{figure*}[t]
	\centering
	\subfloat[Raw na\"{\i}ve fusion]{\includegraphics[width=0.337\linewidth]{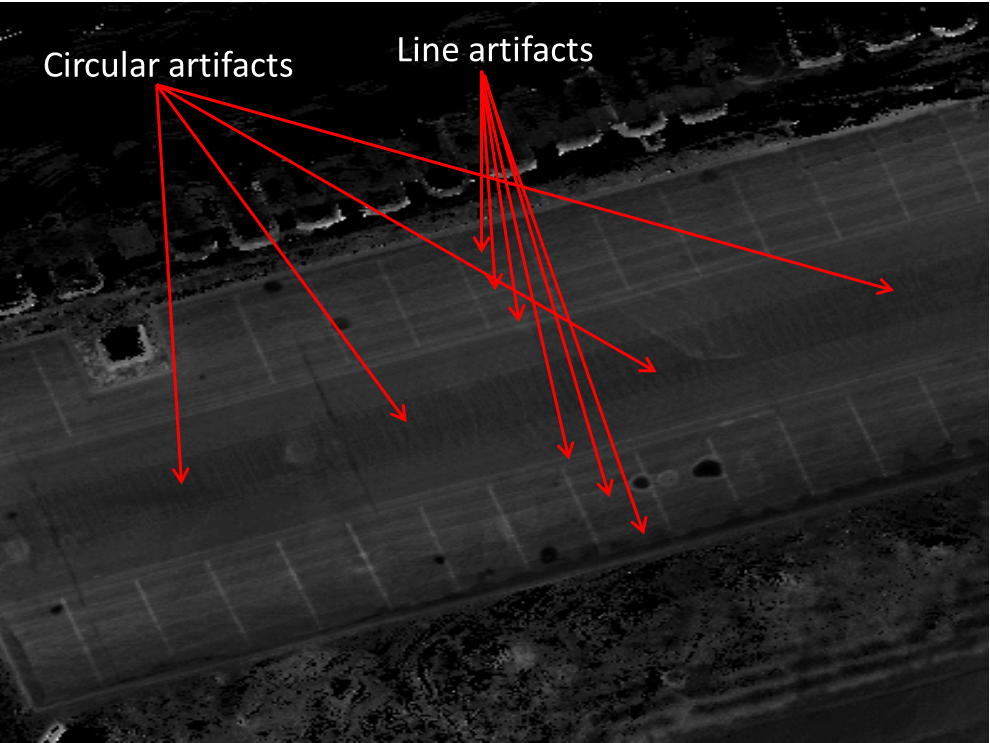} } 
	\centering
	\subfloat[Implementation of Levinson \& Thrun~\cite{Levinson.Thrun2010}]{\hspace{-0.2em}\includegraphics[width=0.34\linewidth]{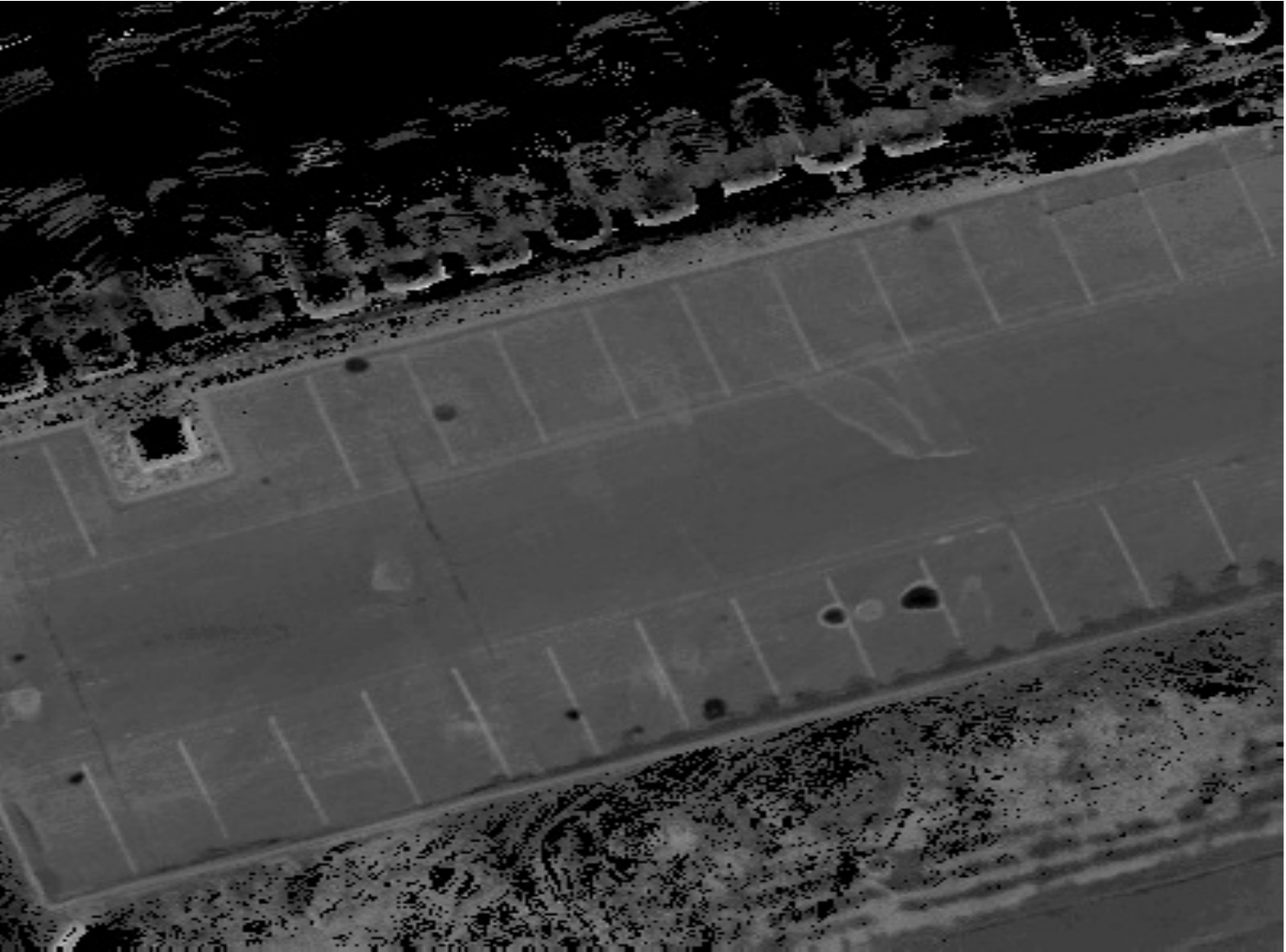}}
	\centering
	\subfloat[Our method]{\includegraphics[width=0.34\linewidth]{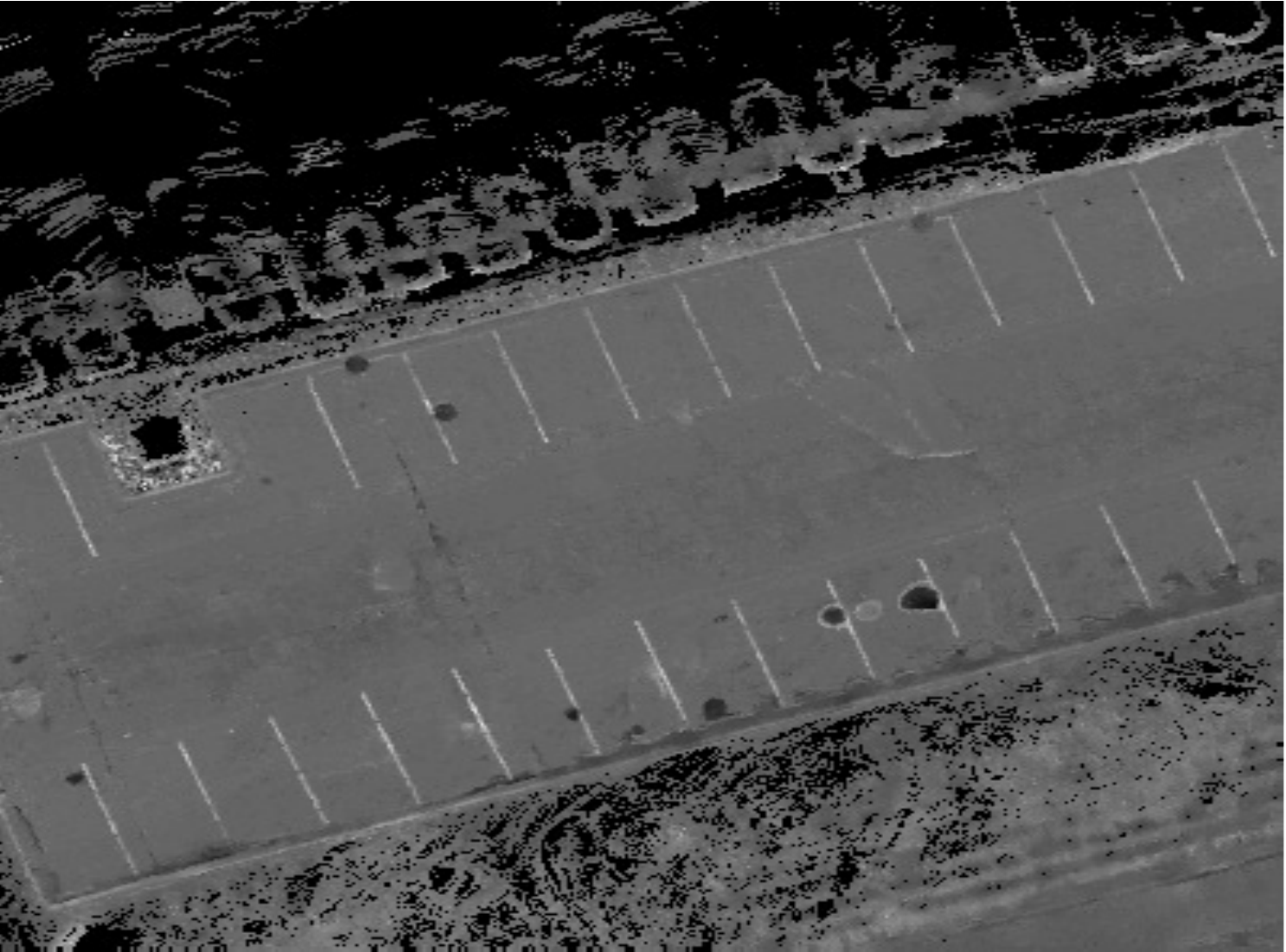}}
	
	\centering
	\subfloat[Raw na\"{\i}ve fusion]{\includegraphics[width=0.338\linewidth]{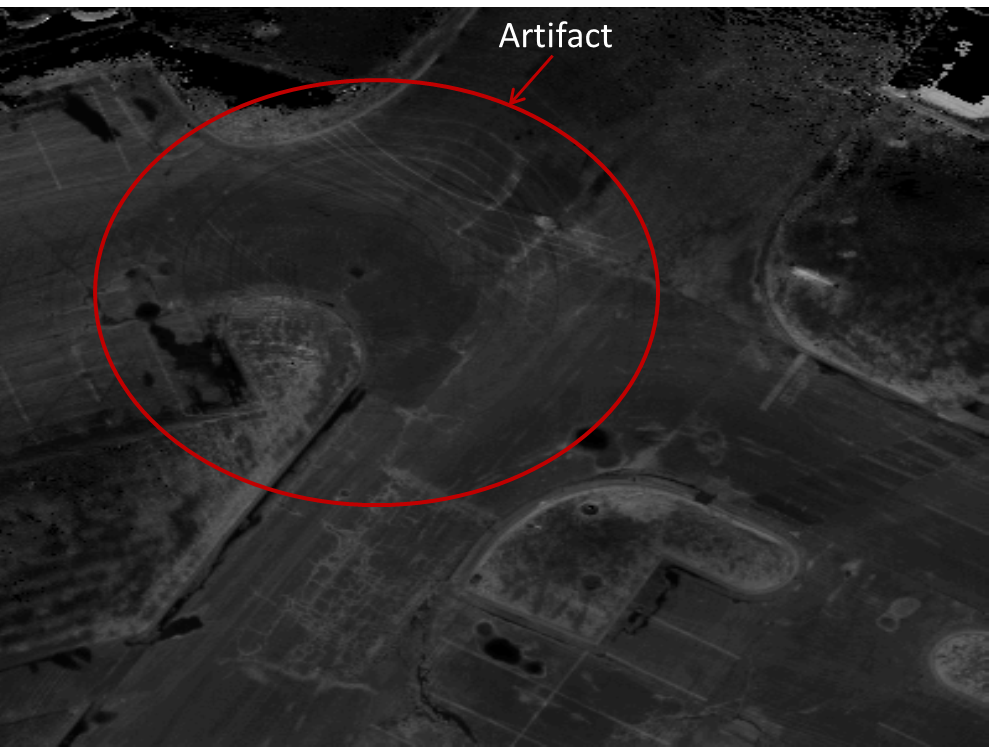}}
	\centering
	\subfloat[Implementation of Levinson \& Thrun~\cite{Levinson.Thrun2010}]
	{\includegraphics[width=0.337\linewidth]{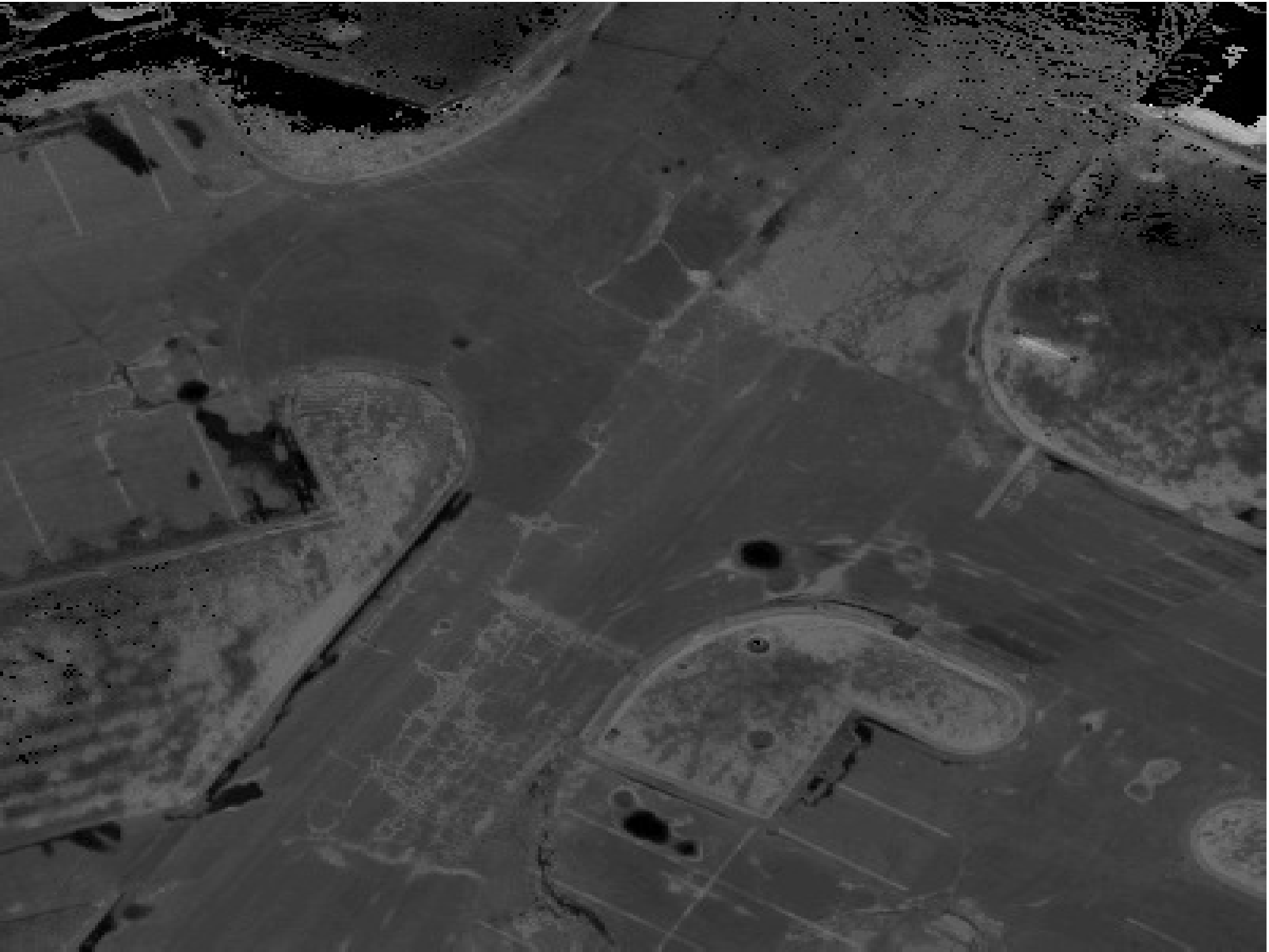}}
	\centering
	\subfloat[Our method]{\includegraphics[width=0.337\linewidth]{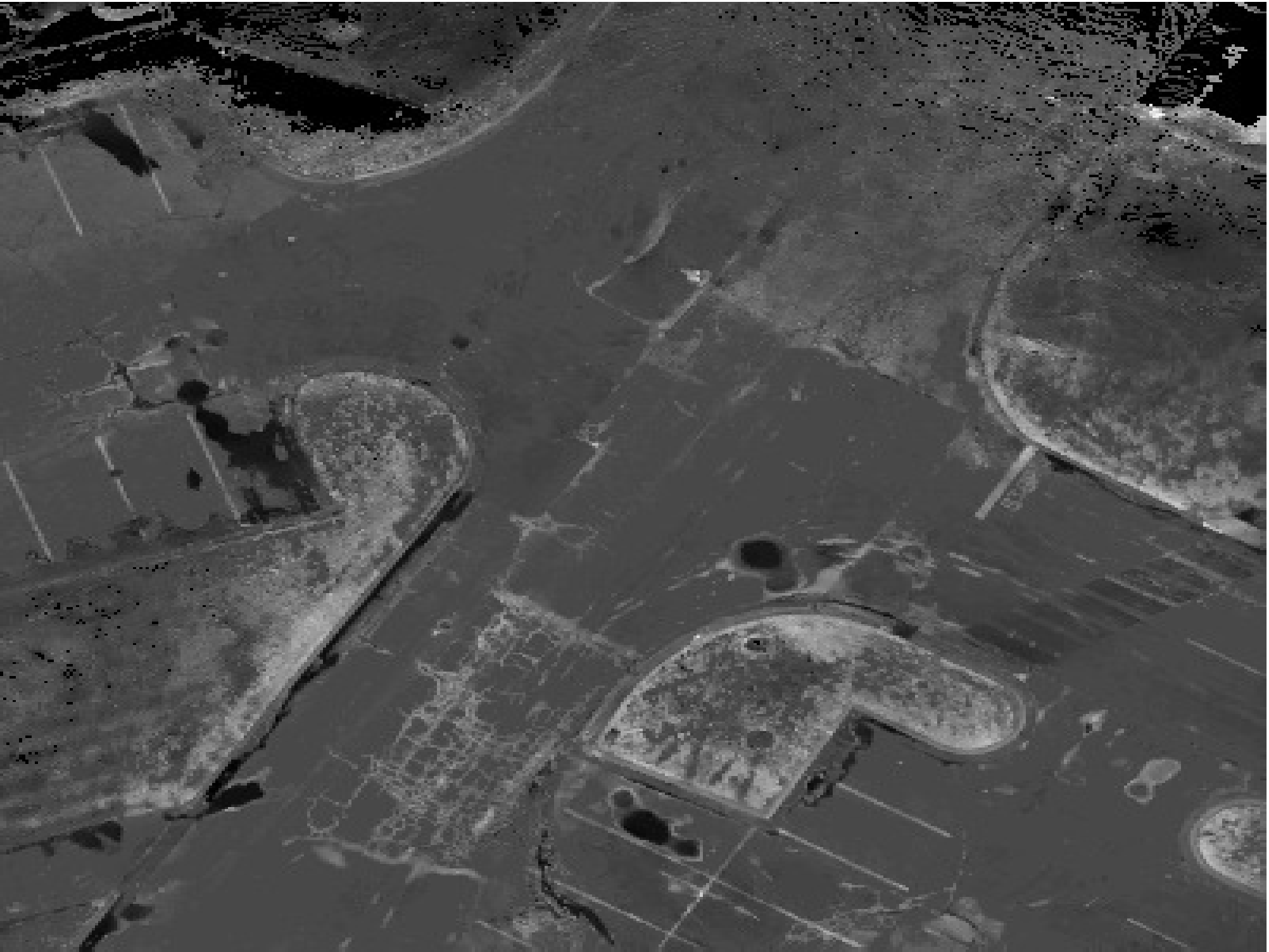}}
	
	\centering
	\subfloat[Raw na\"{\i}ve fusion]{\includegraphics[width=0.337\linewidth]{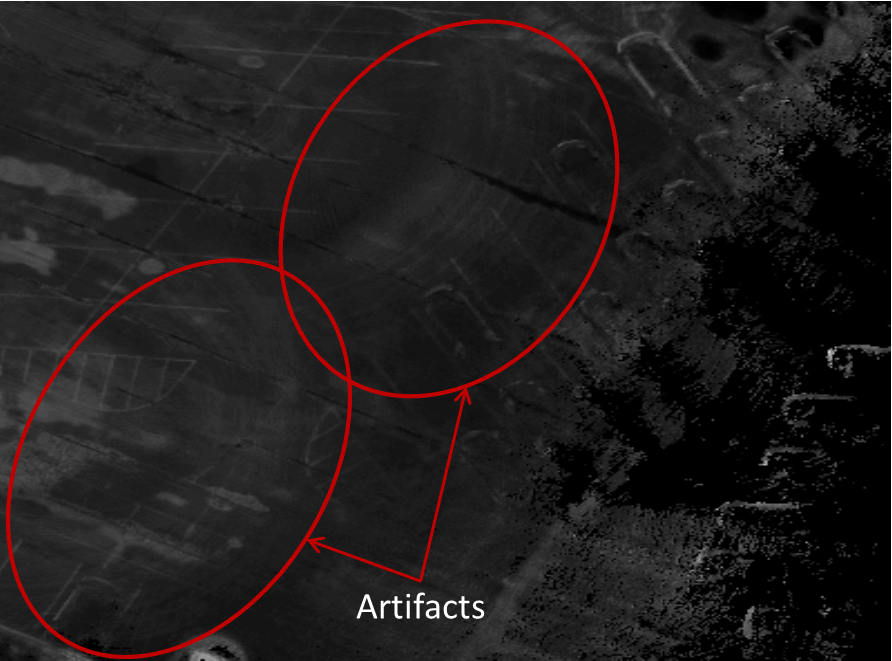}}
	\centering
	\subfloat[Implementation of Levinson \& Thrun~\cite{Levinson.Thrun2010}]
	{\includegraphics[width=0.337\linewidth]{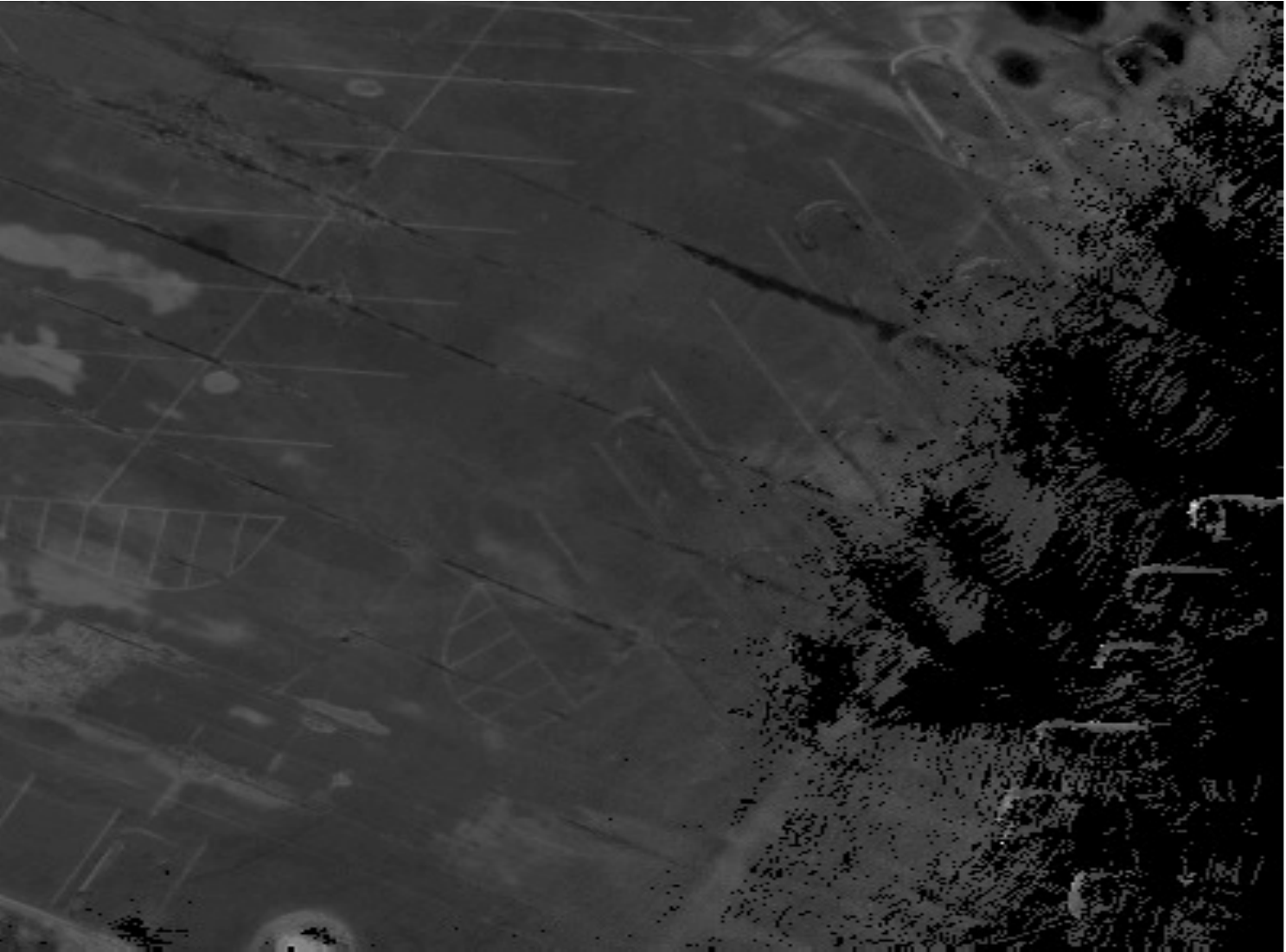}}
	\centering
	\subfloat[Our method]{\includegraphics[width=0.337\linewidth]{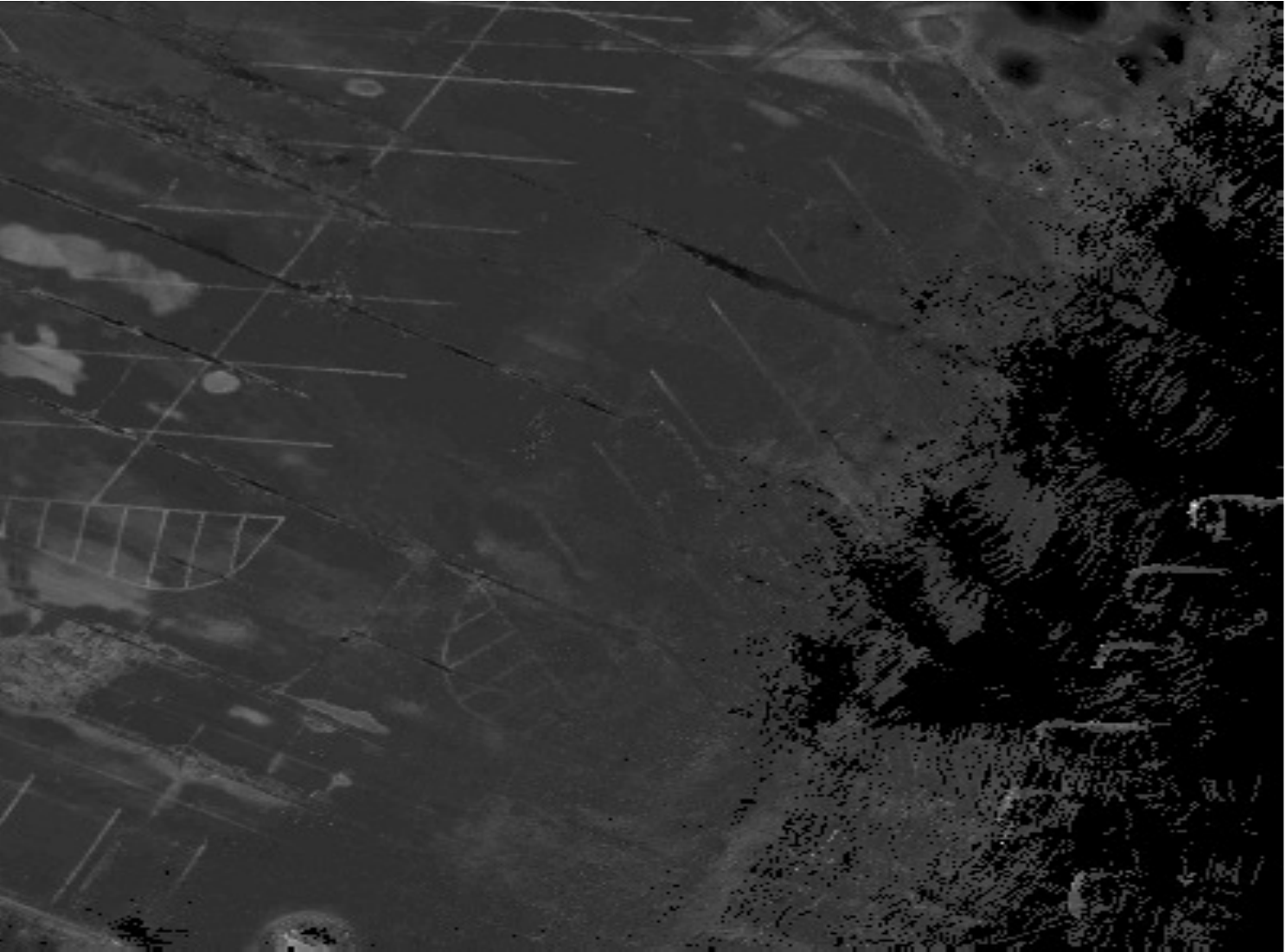}}
	\caption{Patch comparison of artifact correction. (a,d,g) Raw na\"{\i}ve fusion, note: red lines indicate location of artifacts. (b,e,h) Results of our implementation of the reflectivity calibration in \cite{Levinson.Thrun2010}, (c,f,i) Our method. \textbf{Note: Zoom in to better appreciate image quality.}}
	\label{fig:artifactRemoval}
\end{figure*}

Figure \ref{fig:artifactRemoval}.a points in red the circular artifacts that are generated close to the middle of the road. This artifact pattern coincides with the 360\textdegree ~laser scanning pattern while the vehicle is undergoing motion at longitudinal velocity $>6$ m/s. Figure \ref{fig:artifactRemoval}.b shows the generated result when reconstructed with the calibration method of \cite{Levinson.Thrun2010}. This method eliminated most of the artifact, although, at the cost of global smoothing. Figure \ref{fig:artifactRemoval}.c shows the result obtained with our method. Note that a great deal of the artifact is removed while also enhancing (with respect to \ref{fig:artifactRemoval}.a and \ref{fig:artifactRemoval}.b) edge sharpness. 

Figure \ref{fig:artifactRemoval}.d shows an artifact example generated when the vehicle stopped for more than $> 5$ seconds at an intersection. The artifact follows the pattern of the lasers while undergoing a 360\textdegree ~scan trip and is caused by cells accumulating measurements from lasers with dominant reflectivity responses. Figure \ref{fig:artifactRemoval}.e shows the result of applying our implementation of \cite{Levinson.Thrun2010}. Note, that the global smoothing removed a great deal of the artifact, although not in its entirety. Finally, Figure \ref{fig:artifactRemoval}.f illustrates the map with sharp edges and removed artifact obtained with our reconstruction approach.  

The last case shown in Figure \ref{fig:artifactRemoval}.g presents an artifact generated when the vehicle underwent a u-turn. Figure \ref{fig:artifactRemoval}.h shows the result obtained with our practical implementation of \cite{Levinson.Thrun2010}. The artifact generated in the u-turn is eliminated, but again at the cost of global smoothing. In contrast, our method illustrated in Figure \ref{fig:artifactRemoval}.i presents smooth areas while maintaining sharp edges and eliminating the u-turn artifact.   

\subsection{Sparse selection and Denoising}
\label{sec:denoisingSection}

\begin{figure}
	\centering
	\subfloat[Raw na\"{\i}ve fusion]{\includegraphics[width=0.35\linewidth]{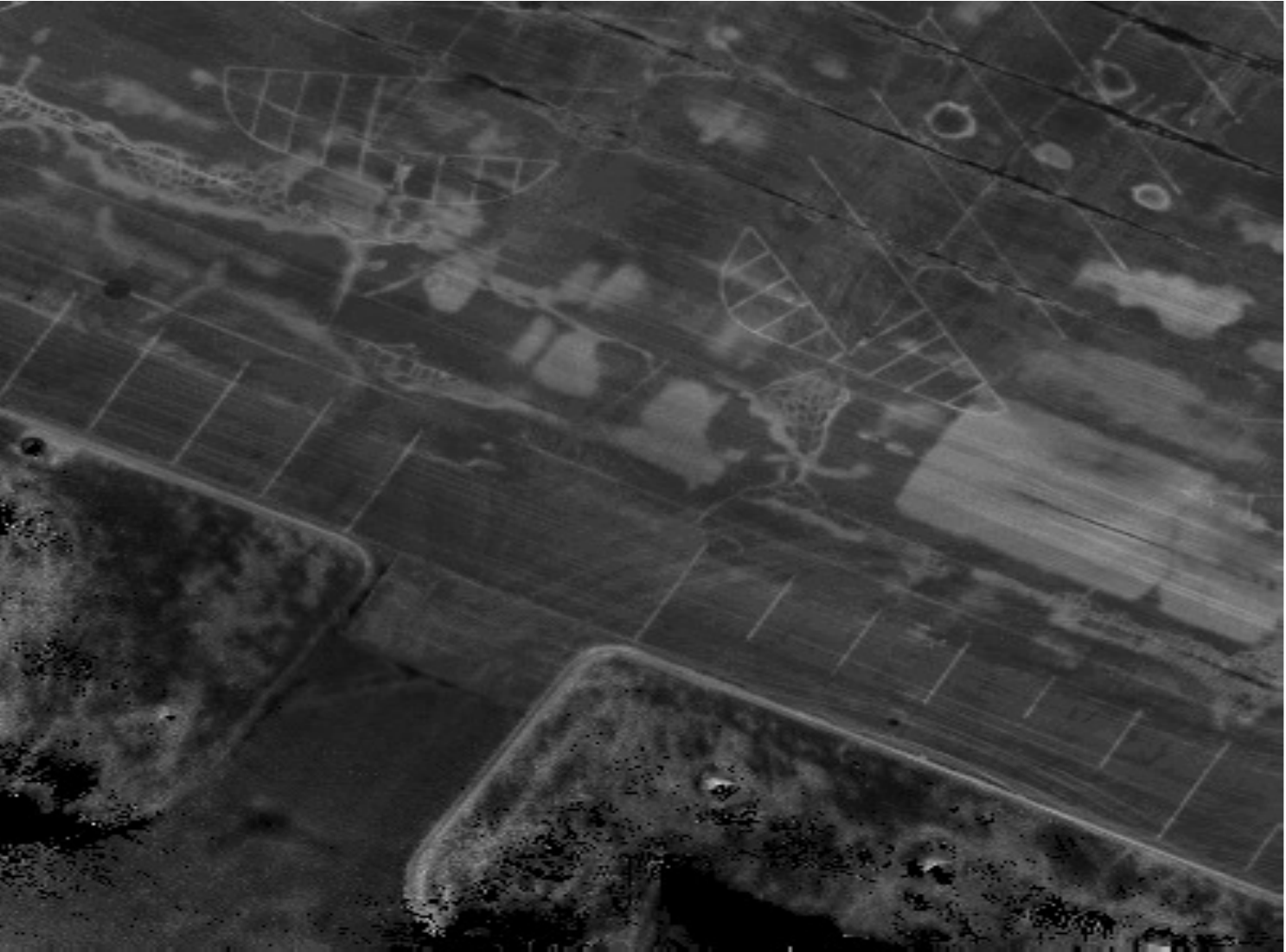}}
	\centering	
	\subfloat[Calibrated fusion]{\includegraphics[width=0.356\linewidth]{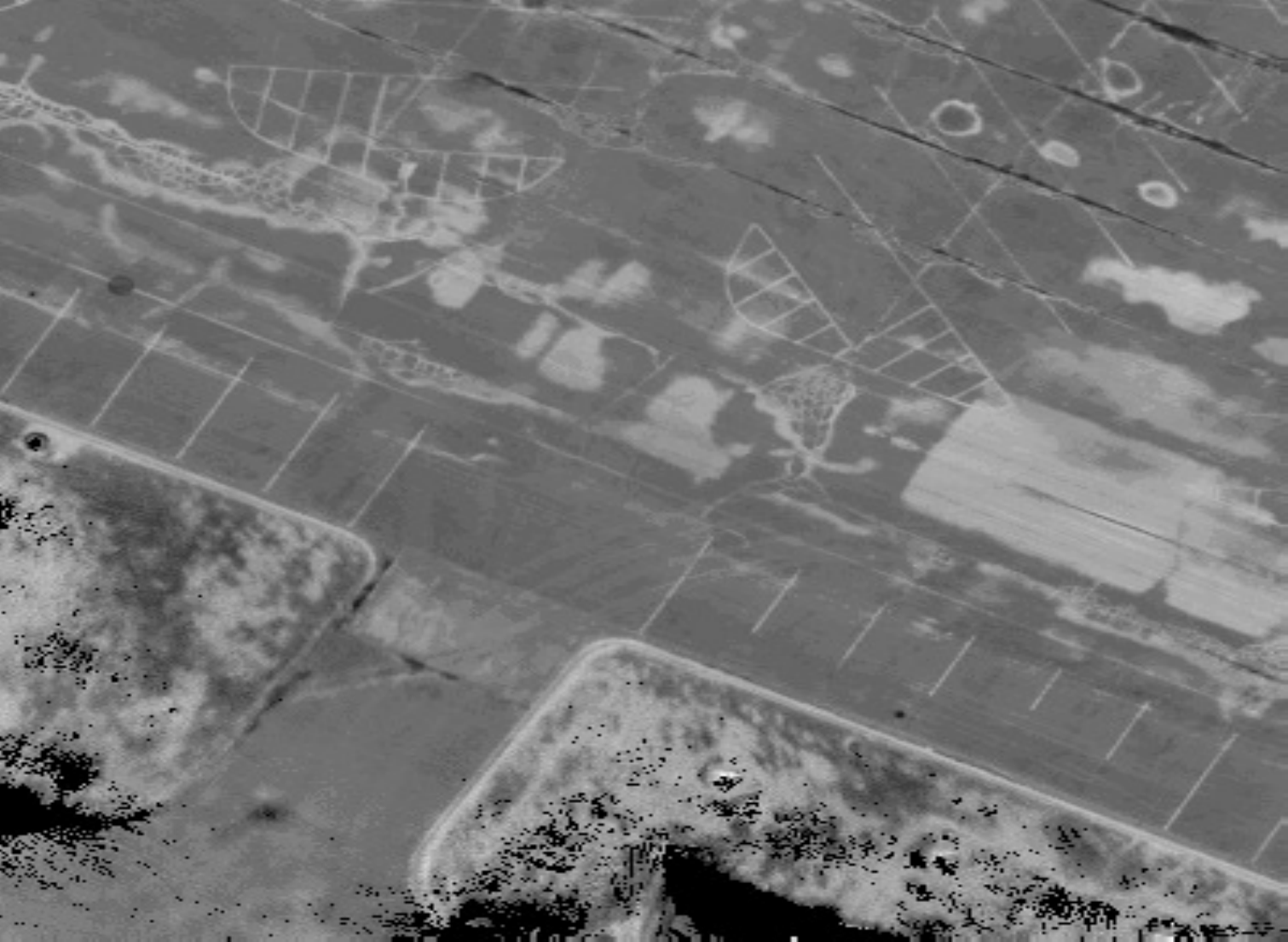}}
	
	\centering
	\subfloat[Mean fusion no denoising ]{\includegraphics[width=0.35\linewidth]{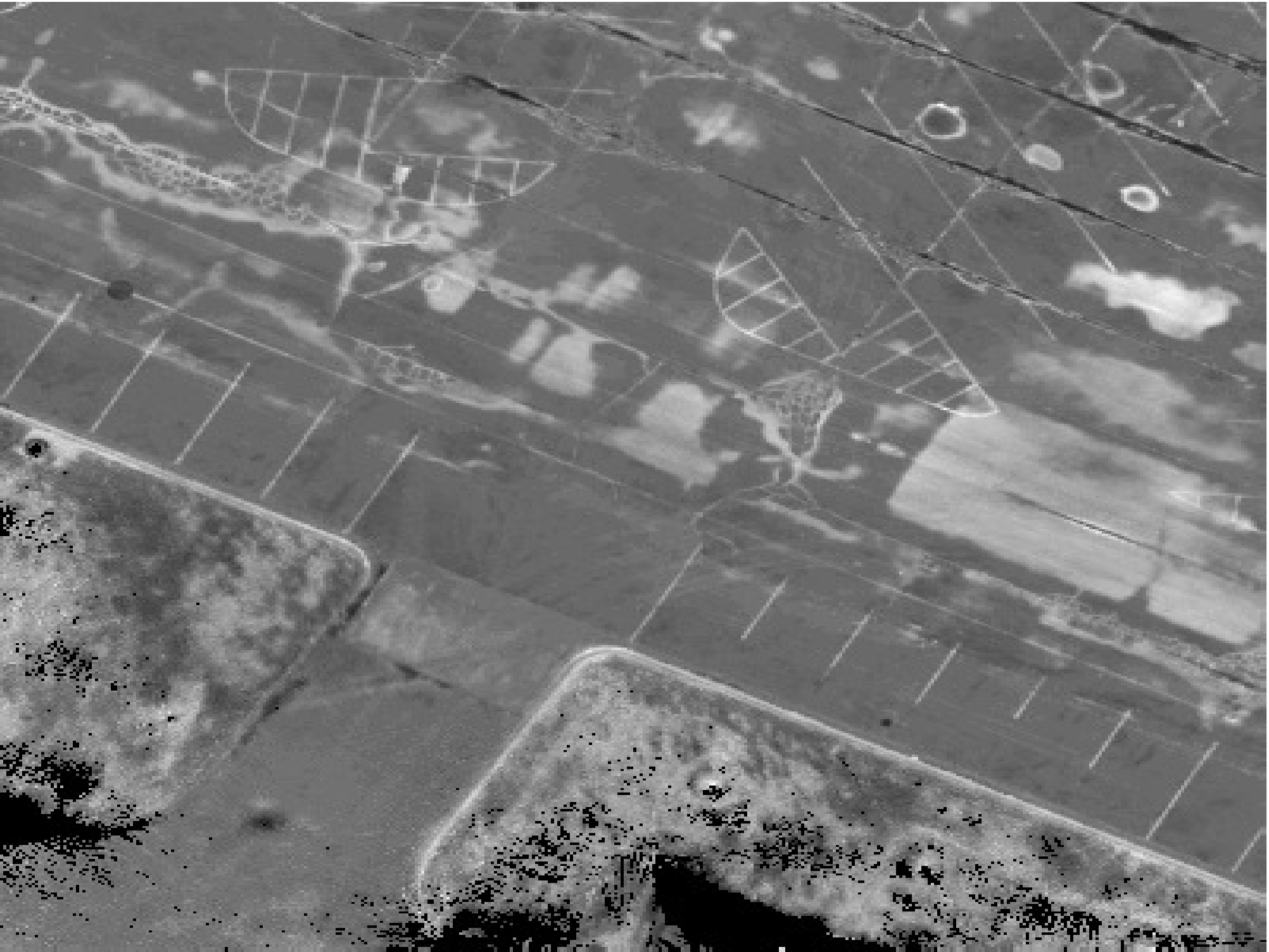}}
	\centering
	\subfloat[Mean fusion with denoising]{\includegraphics[width=0.35\linewidth]{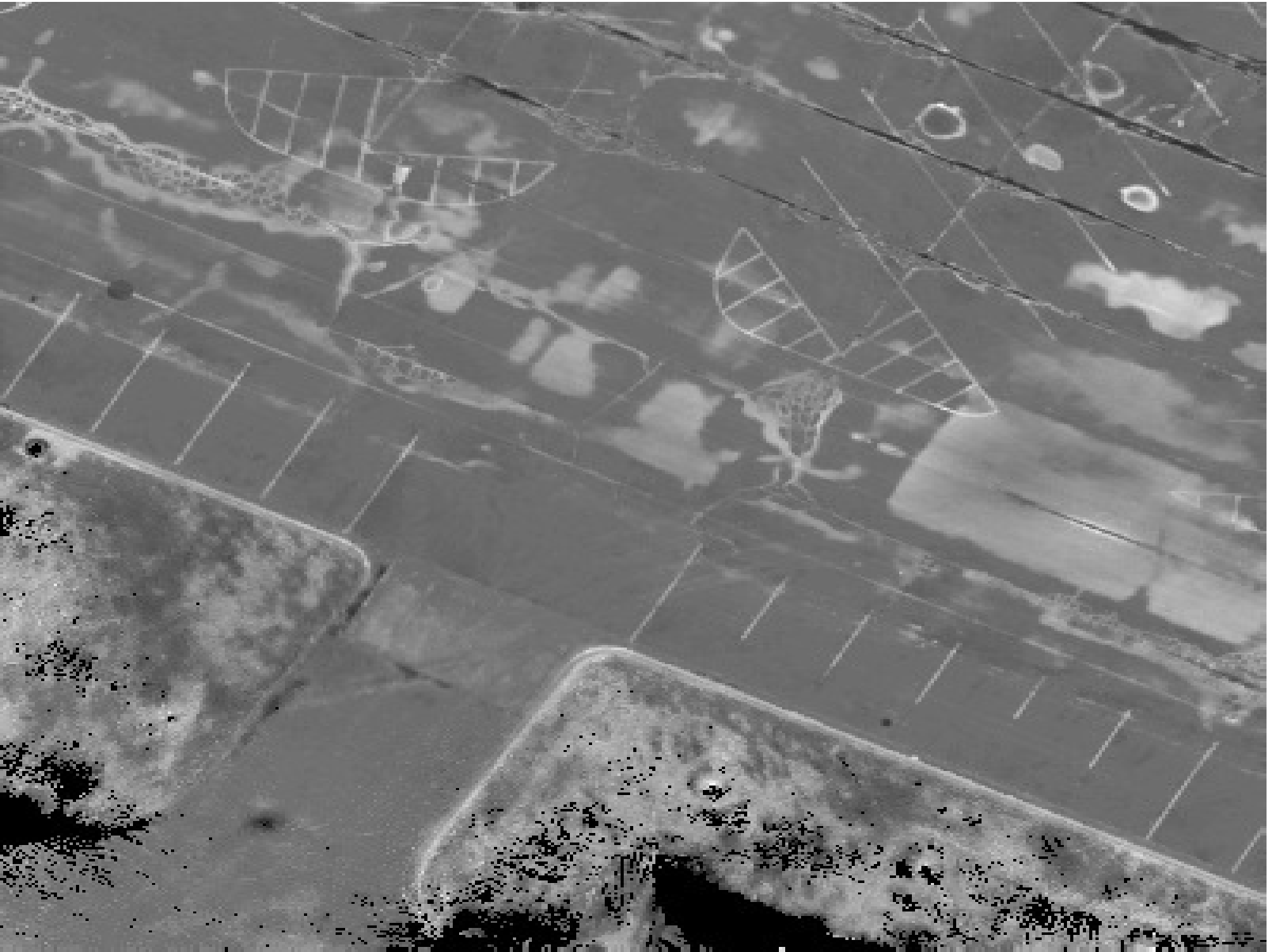}}
		
	\centering
	\subfloat[Top 5 fusion no denoising]{\includegraphics[width=0.35\linewidth]{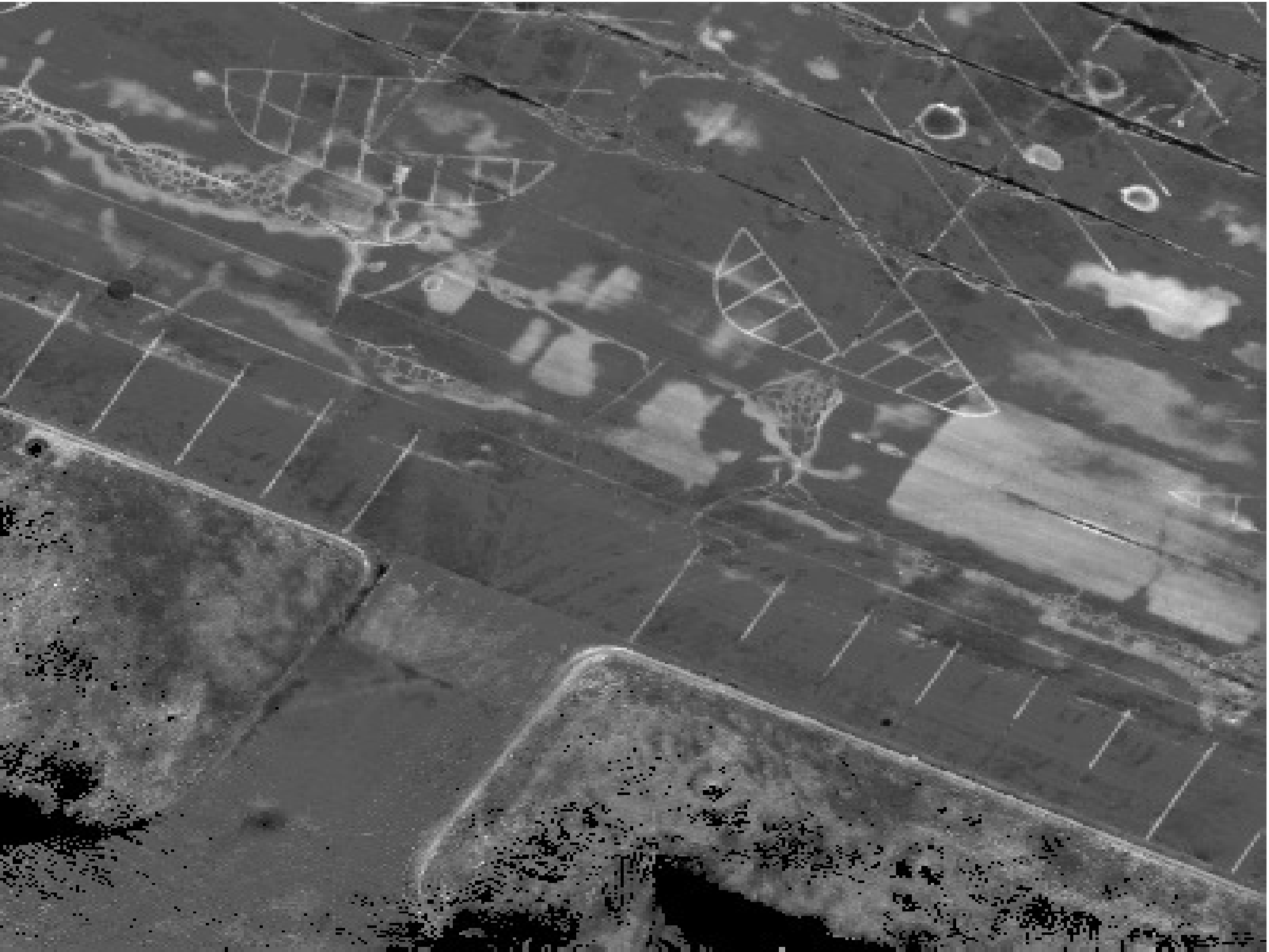}}
	\centering
	\subfloat[Top 5 fusion with denoising]{\includegraphics[width=0.35\linewidth]{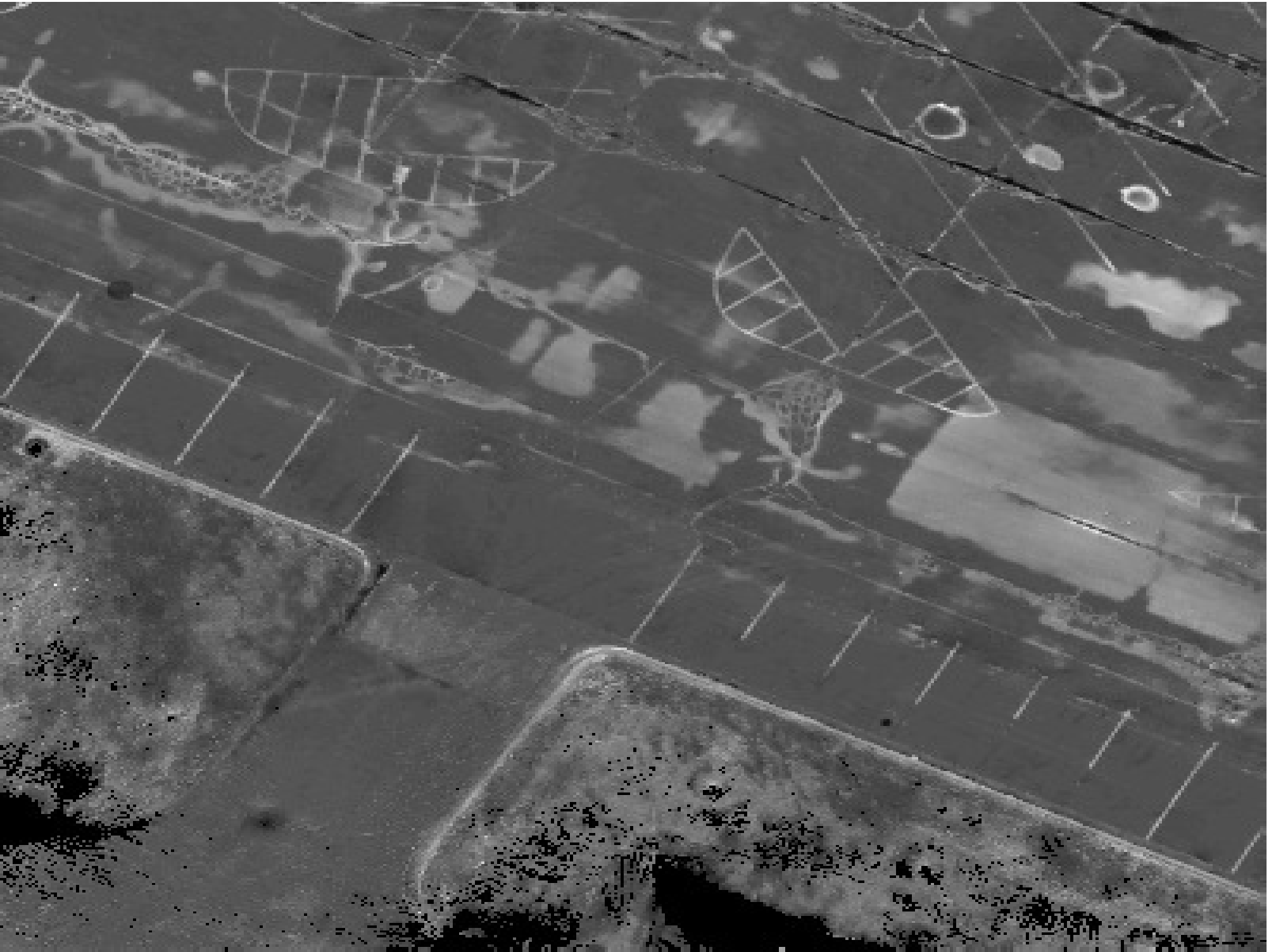}}
	\caption{Map patches of ground reflectivity. (a) Raw na\"{\i}ve fusion, (b) Our practical implementation of~\cite{Levinson.Thrun2010}, (c) Our reconstruction using all map-perspectives, (d) same as (c) with denoising, (e) Our reconstruction with only five map-perspectives, (f) Same as (e) but with denoising. \textbf{Zoom in to better appreciate image quality.}}
	\label{fig:denoisingExample}
\end{figure}

In this subsection we include results that show the plausible operators that can be applied to render maps with improved quality characteristics at the cost of a computational complexity trade-off. For example, the possiblity of sparsely selecting the map-perspectives using \eqref{weightOptimization} and/or include the denoising step through $g_d$ in \eqref{fusion}. %We would like to also add that different performances can be obtained by adjusting the parameters of Algorithms \ref{gradientFusionAlgorithm} and \ref{gradientFieldDenoising}. 
Here, we include a representative example that illustrates the differences between the quality of maps in four cases: (1) Uniform fusion weights (i.e., $\w_{\phibf}=1$) and no denoising, (2) Uniform fusion weights with denoising, (3) Selection of 5 map-perspectives through control of $ \lambda = 1.2e-3$ in \eqref{weightOptimization} with no denoising and (4) Same as (3) but with de-noising.
Figure \ref{fig:denoisingExample} shows a comparison between the aforementioned cases and the raw na\"{\i}ve fusion in Figure \ref{fig:denoisingExample}.a and our implementation of the calibration of \cite{Levinson10} in Figure \ref{fig:denoisingExample}.b. The obtained results with the adjustments (1),(2), (3) and (4) are included in Figures \ref{fig:denoisingExample}.c-f, respectively.
Note that even the result in Figure \ref{fig:denoisingExample}.c obtained when averaging all map-perspectives with no denoising presents smoother regions with sharper edges in comparison to Figure \ref{fig:denoisingExample}.a-b. Figure \ref{fig:denoisingExample}.d adds more smoothing while preserving edge sharpness to \ref{fig:denoisingExample}.c with the denoising step. Finally, Figures \ref{fig:denoisingExample}.e and \ref{fig:denoisingExample}.f illustrate that the sparse selection in the two cases: no denoising and with denoising, respectively results in maps with improved edge sharpness. %and contrast. contrast.

\subsection{Localization}
\label{sec:localization}

\begin{figure}
	\centering
	\subfloat[Local reflectivity map]{\includegraphics[width=0.3\linewidth]{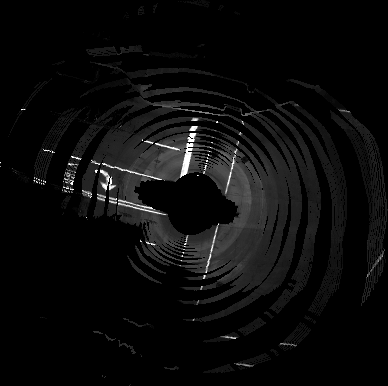}}
	\subfloat[Local gradient magnitude map]{\includegraphics[width=0.3\linewidth]{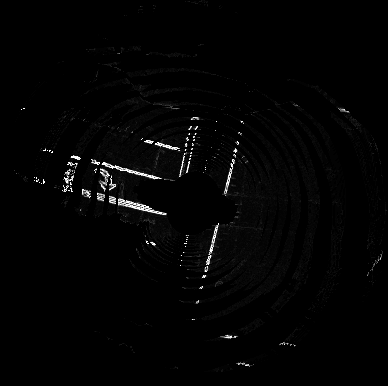}}
	\caption{Ortographic view of a local map of the ground of an area of size $40 \times 40$ m obtained from 8 full-LIDAR revolutions each at 600 Hz while undergoing vehicle motion. a) Local reflectivity map, b) Local gradient magnitude of reflectivity map computed by \eqref{fusion}.}
	\label{fig:localmap}
\end{figure}

The application of our framework described in here is related to the localization of a mobile robot/vehicle in a prior-map of the ground. The idea here is to apply our framework for the task of performing localization corrections via matches between the prior-map of the ground previously reconstructed offline and a map of the locally perceived laser scans of the ground. Here, %To avoid usage of the reflectivity calibration to compensate for variations in the laser sources of the LIDARs which can be computationally intensive and/or expensive if using standard calibration targets 
we propose edge alignments as our matching mechanism via corresponding isotropic gradient magnitudes of the prior-map and local map of ground reflectivities. A similar matching criterion was envisioned in the work of \cite{Castorena.Kamilov.Boufounos2016} for LIDAR-to-vision registration. One of the advantages of this procedure is that a post-factory reflectivity calibration process to reduce the variations in response across the multiple lasers observing the ground is not required to compute any of the global or local maps. Moreover, the gradient magnitude of the local reflectivity map can be computed very efficiently on the fly by means of \eqref{fusion} with uniform weights and no denoising of the gradient components.  An example which illustrates a local reflectivity map along with its corresponding gradient magnitude map of reflectivity is shown in Figure \ref{fig:localmap}. Given this, the localization problem is posed as an optimization over $\tbf$ (i.e., the 3 DOF parameters: longitudinal (x), lateral (y) and head rotation (h) vehicle pose). Here, we propose to maximize the normalized mutual information \cite{Studholme1999} between gradient magnitudes of patches $A = | \g\Xbf_{\Omega_{\tbf}} |$ of the prior-map $\Xbf$ and the local map computed on the fly $B = | \g \Xbf_l |$. In other words, 
\begin{equation}
\widehat{\tbf} = \arg \max\limits_{ \tbf \in \Tcal } 
\left \{  
\text{NMI}\{ A, B \}
\right \},
\end{equation}
where $\Tcal$ is a small pose neighborhood set and $\text{NMI}$ is the normalized mutual information given as $\text{NMI}\{ A, B \} = ( H(A) + H(B) )/H(A,B)$ where $H(A)$ and $H(B)$ is the entropy of random variable $A$ and $B$, respectively and $H(A,B)$ is the joint entropy between $A$ and $B$.
Using the mutual information of gradients effectively decouples the dependencies of neighboring pixels in the measure \cite{Jumarie96}.

Included below are the results that include a performance comparison of our proposed method and the method in \cite{Levinson10} to vehicle localization. For our evaluation, we used the position estimate corrections obtained after application of offline graphSLAM (e.g., iSAM \cite{Kaess.Ranganathan.Dellaert2008}) as ground truth, similar to \cite{Levinson10}. Table \ref{tab:table1} summarizes the localization performance comparison reporting the longitudinal, lateral and head angle RMSE obtained when driving over roads with different features (i.e., lane markings). Here, our method matches edges computed via the gradient-map perspective fusion of \eqref{fusion} with uniform weights and no denoising to make localization computationally tractable in real time. Note that even in the case of using \eqref{fusion} with uniform weights and no denoising we achieve slightly better performance than \cite{Levinson10} which depends on a reflectivity calibration stage to reduce inter-beam reflectivity response variations.
%
% Spatial feature comparison
% 
\begin{table*}
	\caption{\label{tab:table1} RMSE Localization performance comparison}
	\centering
	\scalebox{0.87}{
	\begin{tabular}{lcccccc}
		\hline
		\\
		\multirow{2}{*}{Road Scene Category} &\multicolumn{3}{c}{Levinson and Thrun \cite{Levinson.Thrun2010}} & \multicolumn{3}{c}{Ours: Uniform weights no denoising} \\
		& Longitudinal & Lateral & Head & Longitudinal & Lateral & Head \\
		& (cm) & (cm) & (rads)& (cm) & (cm) & (rads)\\
		\hline
		\\
		Straight road 1, urban unmarked (UU) & 8.8 & 2.2 & 2.8e-3 & 2.8 & 1.7 & 3.4e-3 \\
		Straight road 2, UU & 10.0 & 0.9 & 1.5e-3 & 3.9 & 0.6 & 1.4e-3 \\
		Straight road 3, UU & 4.8 & 1.0 & 5.1e-4 & 2.7 & 1.1 & 4.7e-4 \\			
		Curvy road, UU & 7.6  &  2.6 & 4e-3 & 4.9 & 1.5 & 3e-3 \\
		Wide road driving in circles, UU & 3.4 & 2.3 & 3.5e-3 & 2.7 & 2.1 & 3.2e-3 \\
		Curvy road 1 urban multiple marked lanes (UMM) & 5.8 & 2.3 & 2.2e-3 & 4.7 & 1.9 & 1.9e-3 \\
		Curvy road 2 UMM & 4.9 & 4.0 & 2.9e-3 & 1.9 & 2.2 & 2.8e-3\\
		Straight road 1 urban marked (UM) & 12.6 & 1.8 & 2.5e-3 & 8.4 & 1.6 & 2.2e-3 \\
		Straight road 2 UM & 5.1 & 1.7 & 2.9e-3 & 3.1 & 1.0 & 2.4e-3 \\
		Straight road 3 UM & 7.2 & 1.1 & 9.7e-4 & 5.1 & 1.9 & 7.3e-4 \\
		Straight road 4 UM & 4.3 & 1.2 & 3.7e-4 & 1.5 & 1.1 & 3.5e-4 \\
		\hline
	\end{tabular}
	}
\end{table*}
%

\begin{figure}
	\centering
	\subfloat[Longitudinal error]{\includegraphics[width=0.5\linewidth]{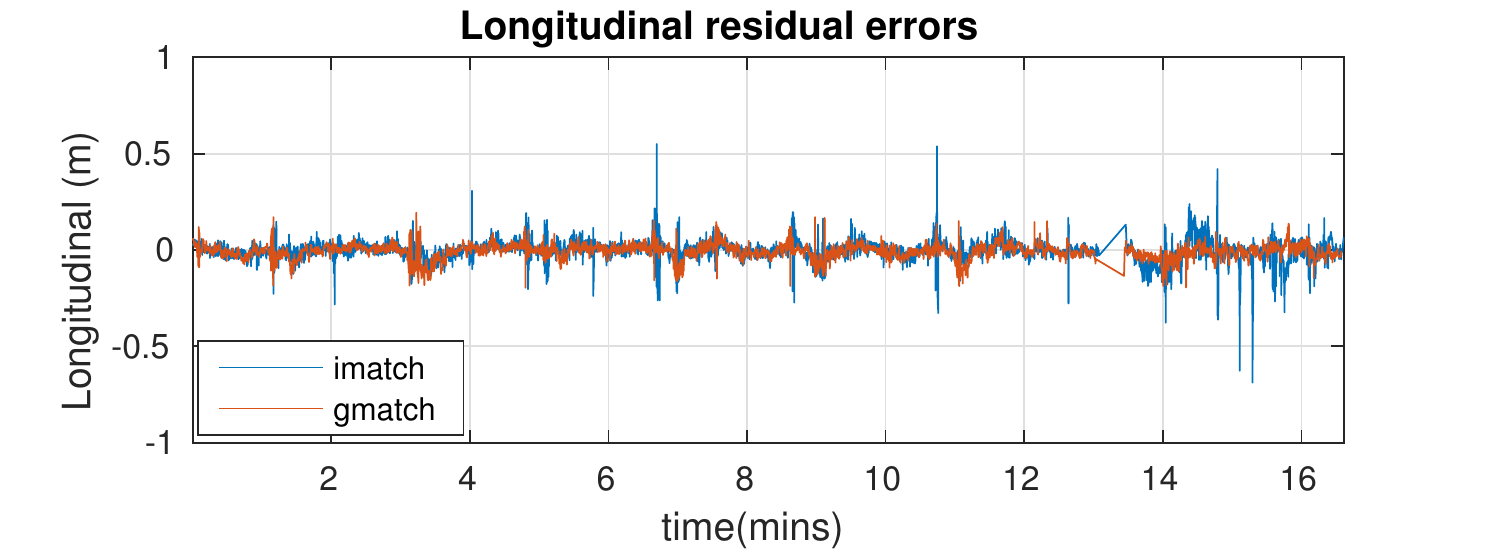}}
	\subfloat[Lateral error]{\includegraphics[width=0.5\linewidth]{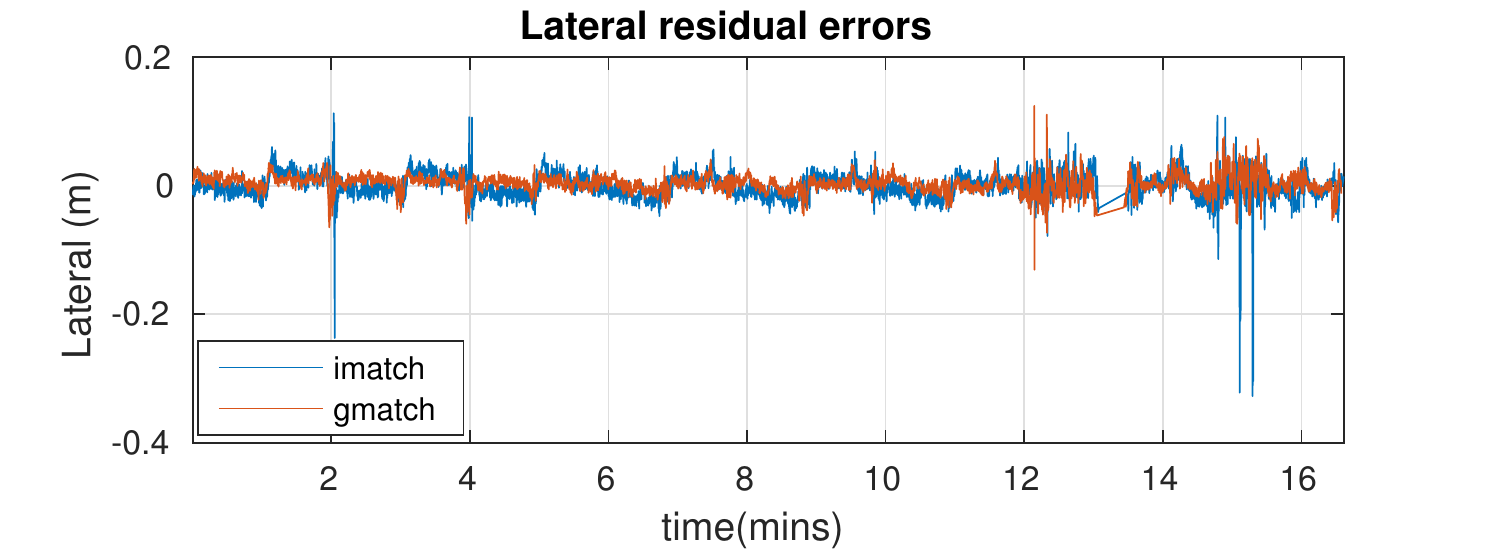}}
	
	\subfloat[Head angle error]{\includegraphics[width=0.5\linewidth]{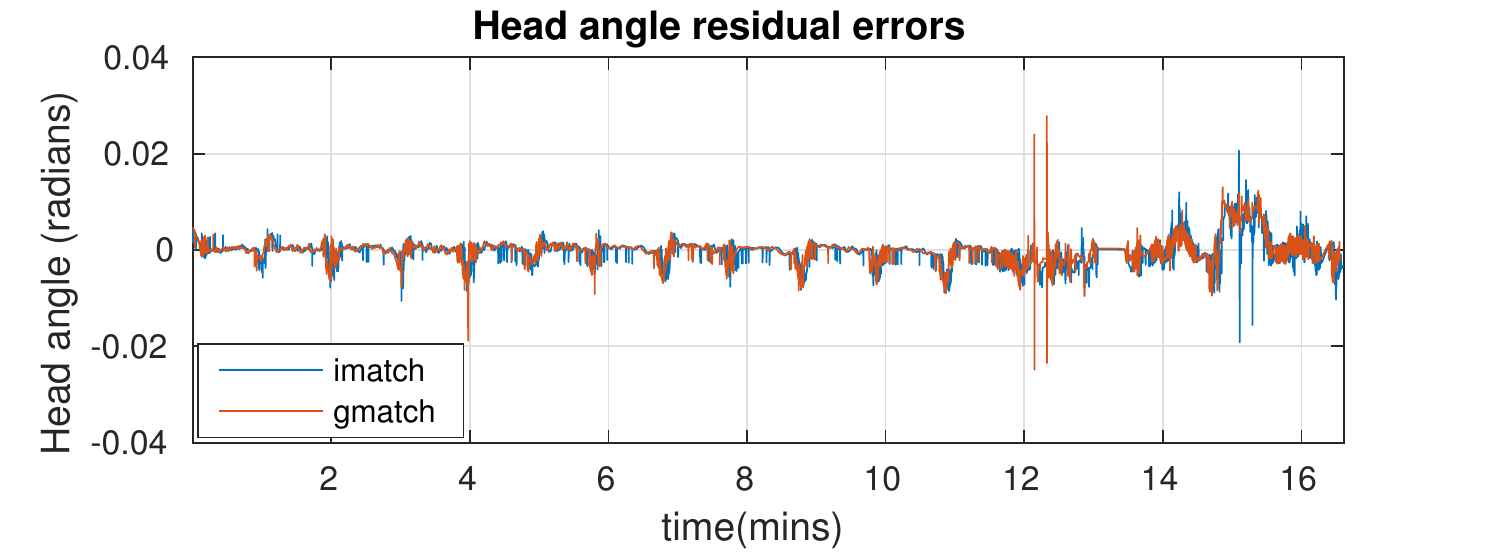}}

	\caption{Localization residual error comparisons. (a) Longitudinal errors. (b) Lateral errors. c) Head angle errors.}
	\label{fig:localization}
\end{figure}

% 17 min experimentation
In addition, we also include Figure \ref{fig:localization} which illustrates the performance comparison of about 8500 registration matches performed over a lapse of $~$17 mins. We found that an RMSE of 3.4 cm, 1.2 cm and 2.5$e-3$ radians for the longitudinal, lateral and head angle, respectively with our proposed method. These reported values are better than those achieved by our implementation of the method of \cite{Levinson10} that resulted in RMSE values of 5.5 cm, 2.1 cm and 2.5$e-3$ in the longitudinal, lateral and head angle directions, respectively. Note that our proposed method achieves this performance without the requirement of performing the post-factory calibration of each laser beams' reflectivity which can currently take up to 4 hrs to a skilled artisan per vehicle. Such cost is prohibitive when production of these scales up to the masses.

\subsection{Road marking extraction}
\label{extraction}

\begin{figure}
	\centering
	\setlength\tabcolsep{1.5pt}
	\begin{tabular}{cccc}
		& UM Patch 1 & UM Patch 2 & UMM Patch 3 \\
		
		\rotatebox[origin=l]{90}{L\&T \cite{Levinson.Thrun2010}}
		& \includegraphics[width=0.18\linewidth]{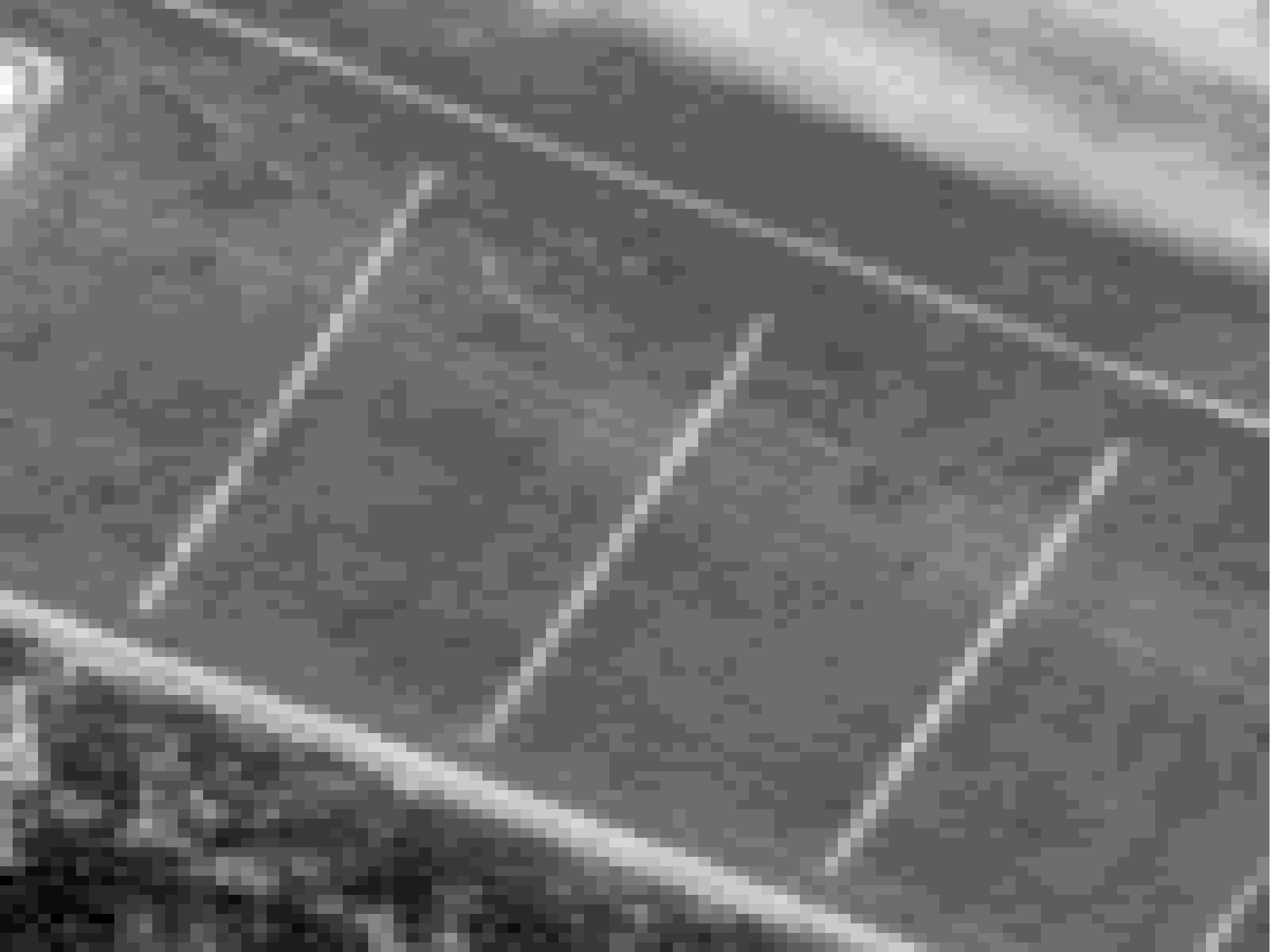}
		& \includegraphics[width=0.182\linewidth]{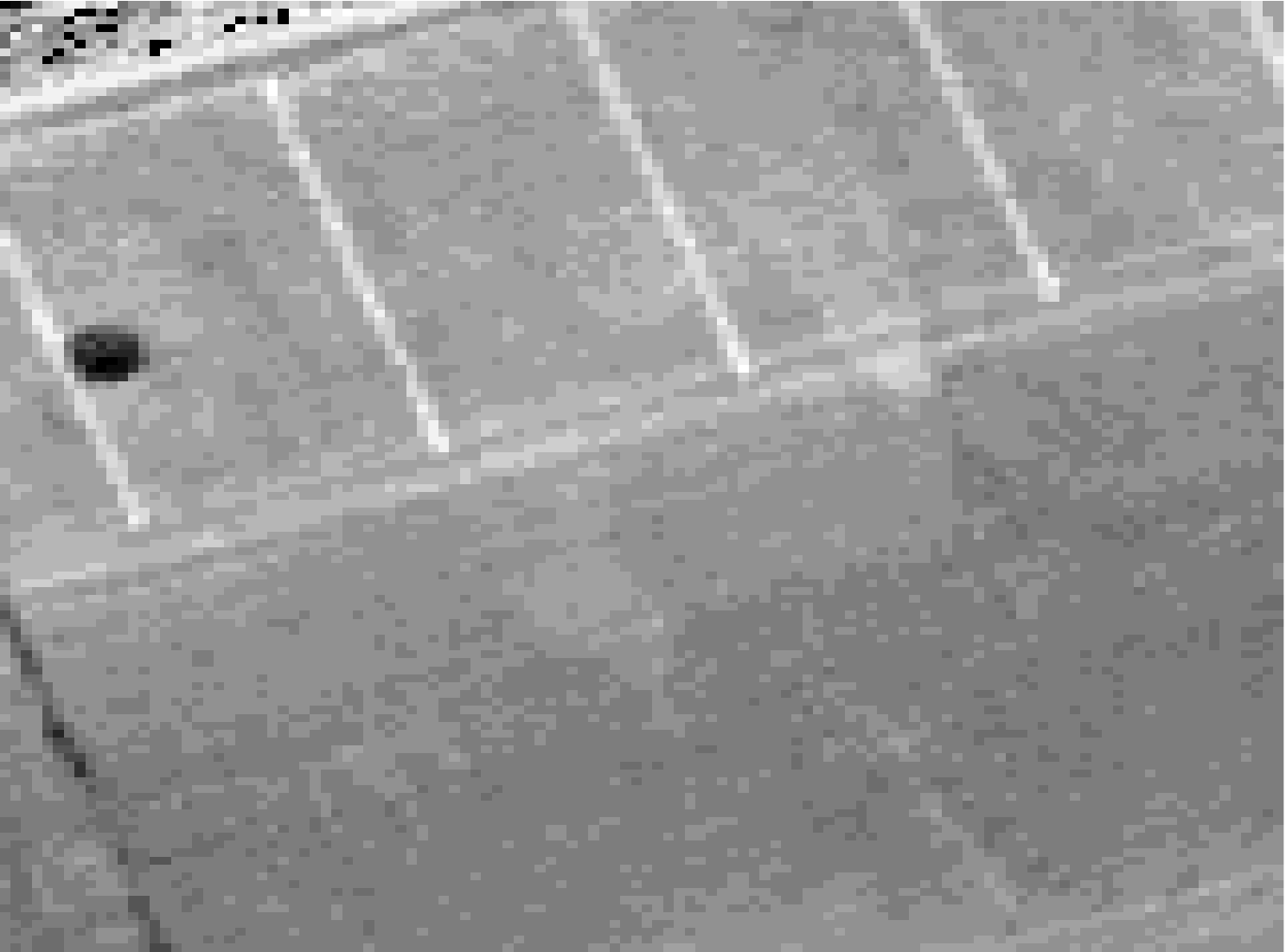}
		& \includegraphics[width=0.18\linewidth]{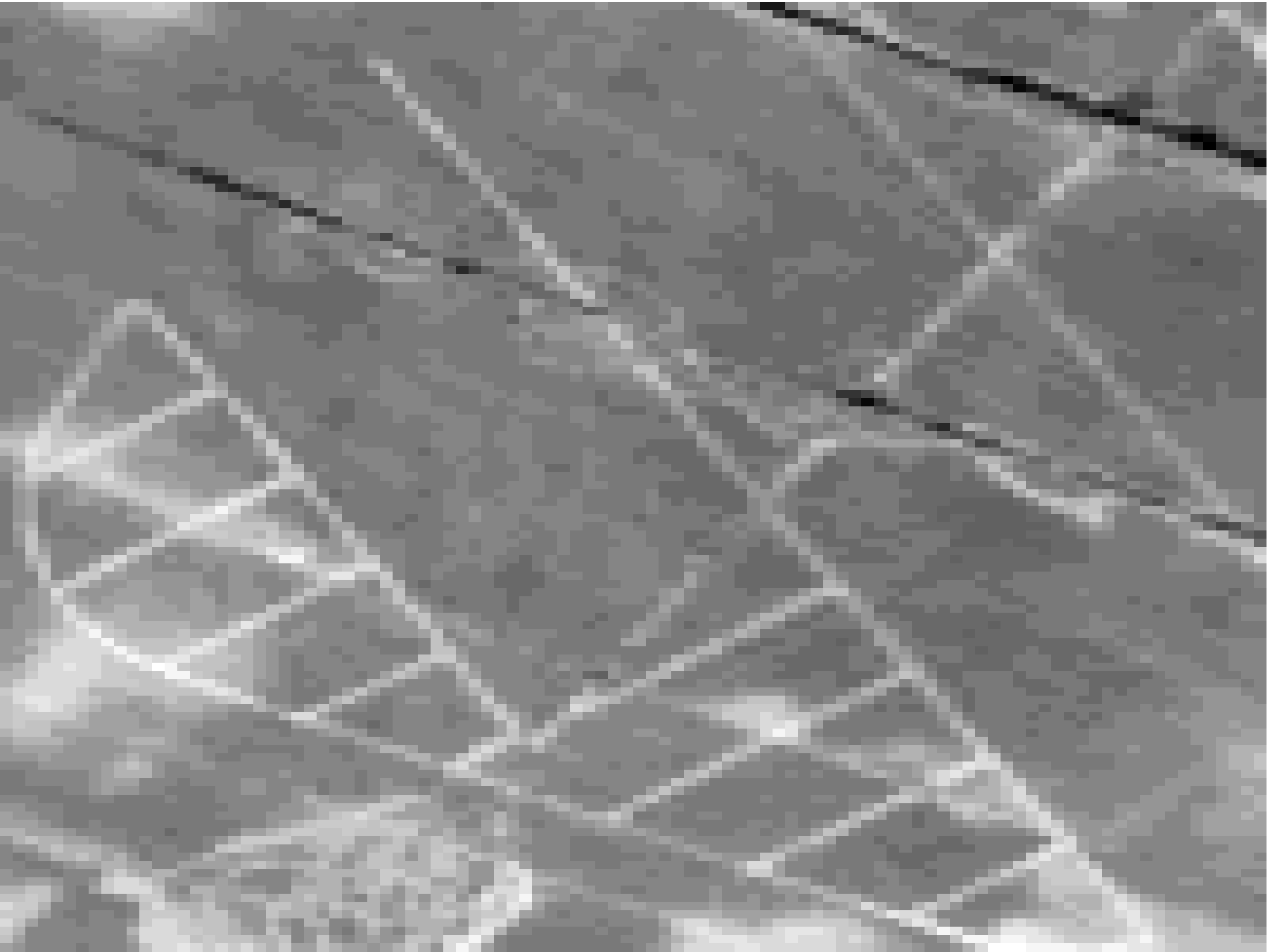} \\
		
		\rotatebox[origin=l]{90}{Ours}
		& \includegraphics[width=0.18\linewidth]{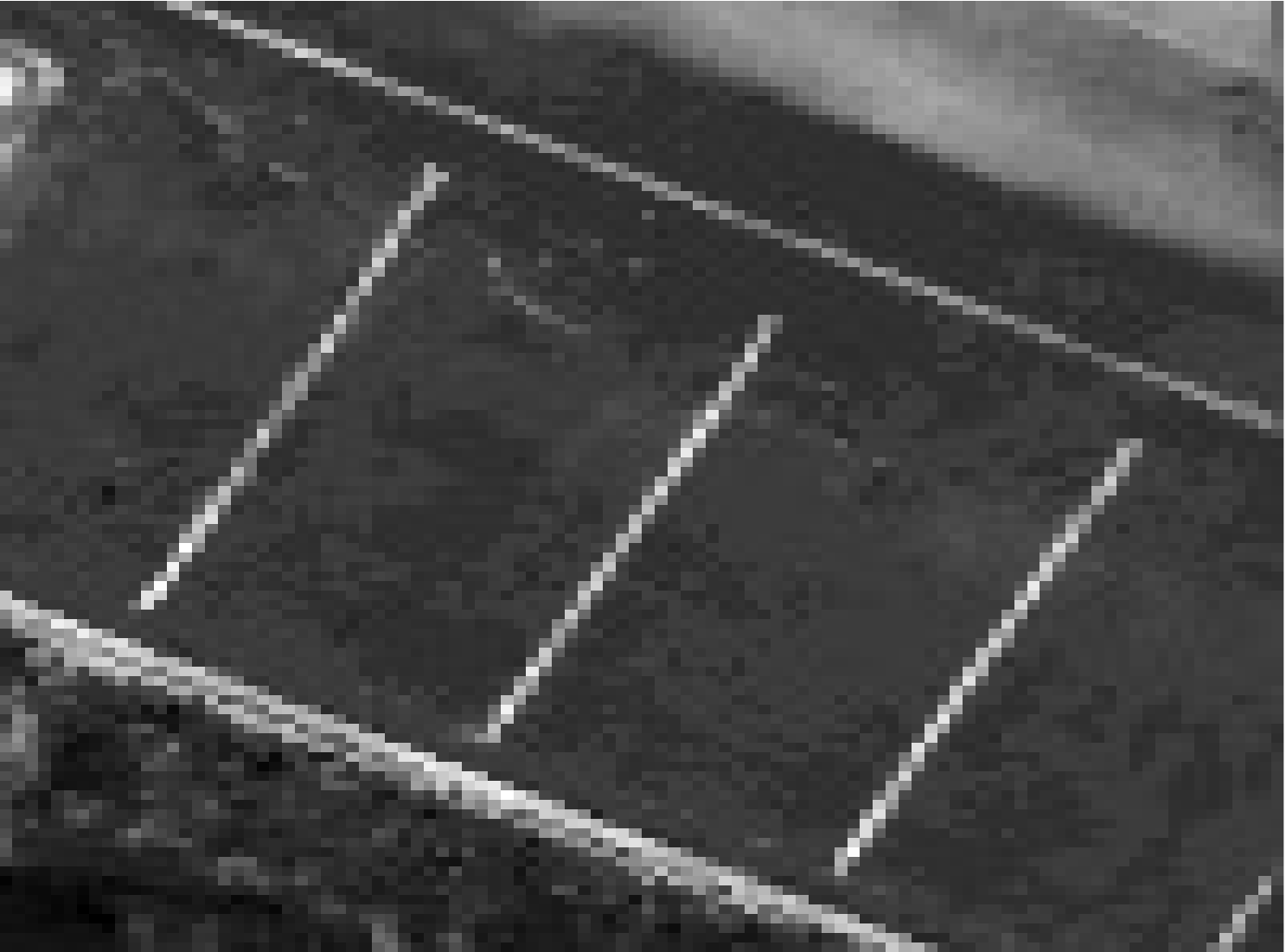}
		& \includegraphics[width=0.18\linewidth]{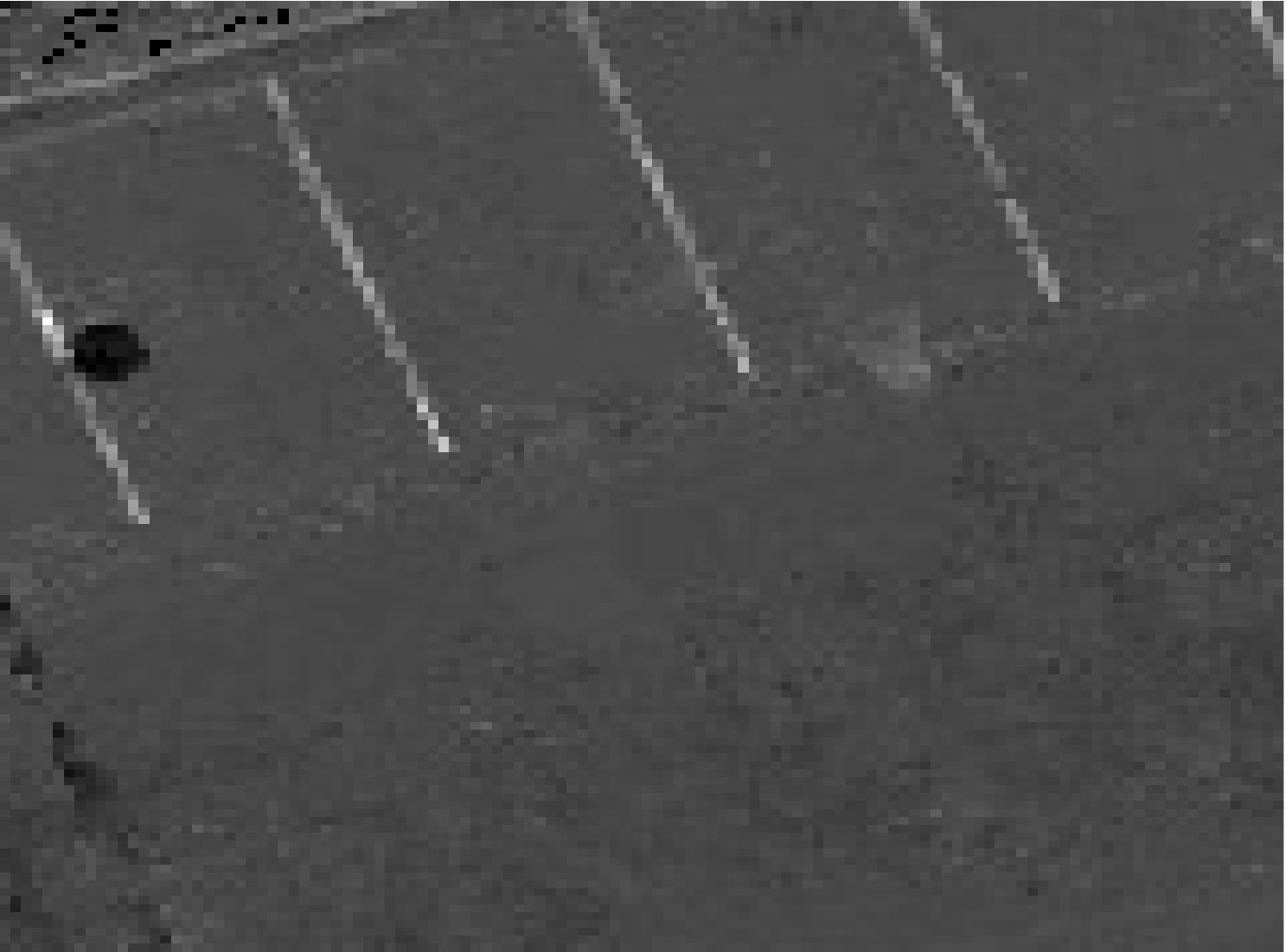}
		& \includegraphics[width=0.178\linewidth]{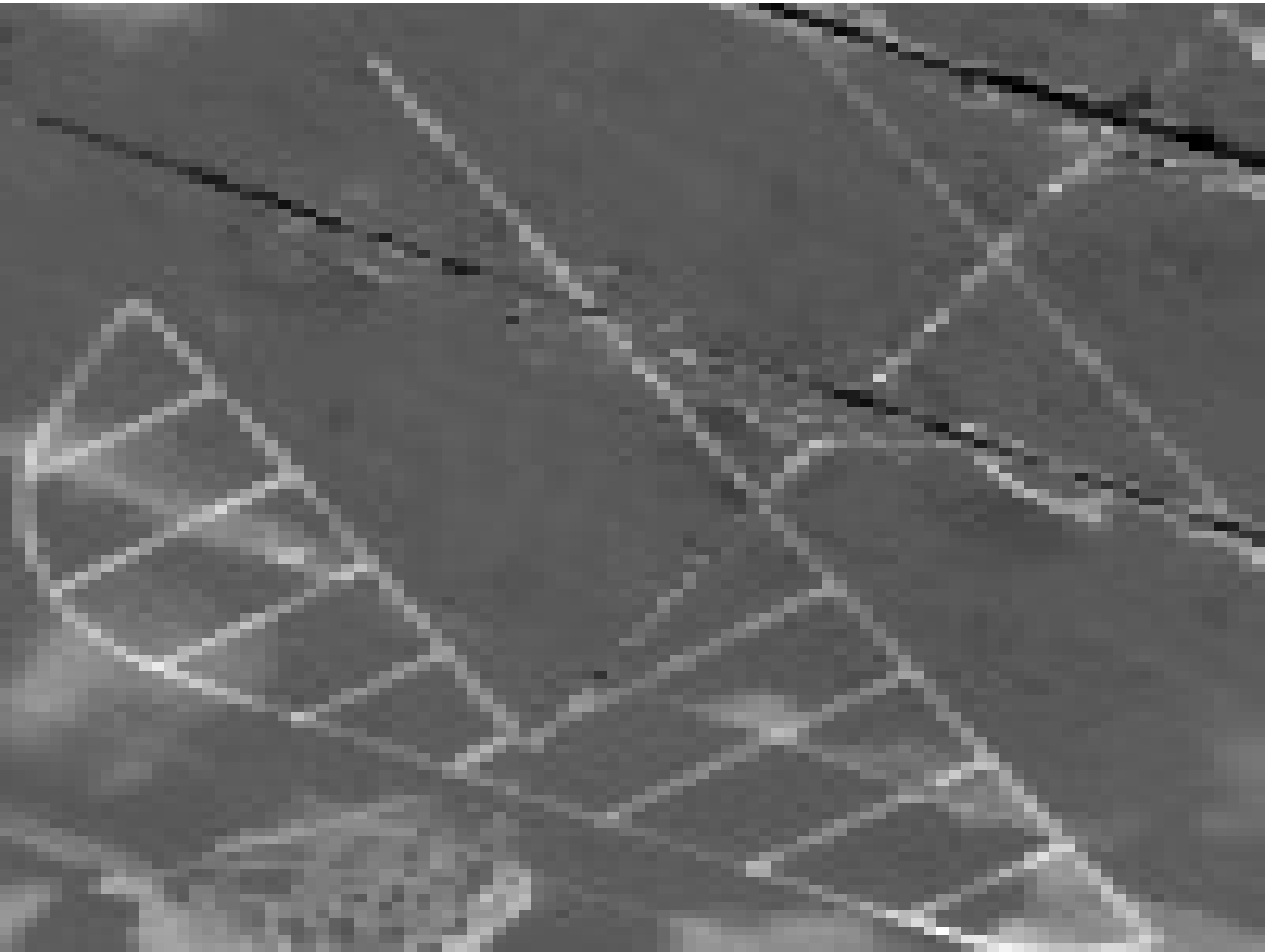} \\
		
		\rotatebox[origin=l]{90}{\shortstack{ L\&T \\Thresh} }
		& \includegraphics[width=0.18\linewidth]{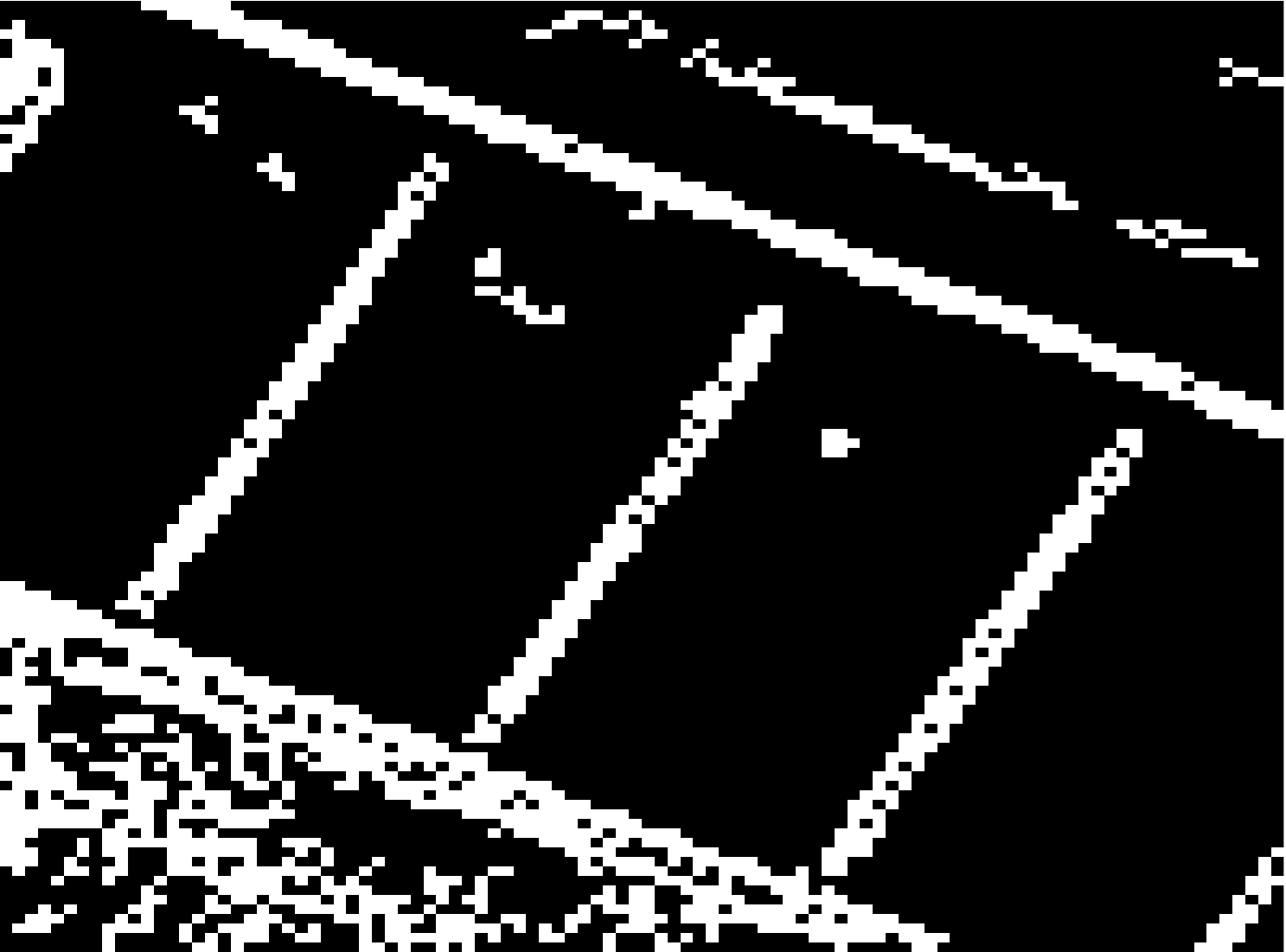}
		& \includegraphics[width=0.18\linewidth]{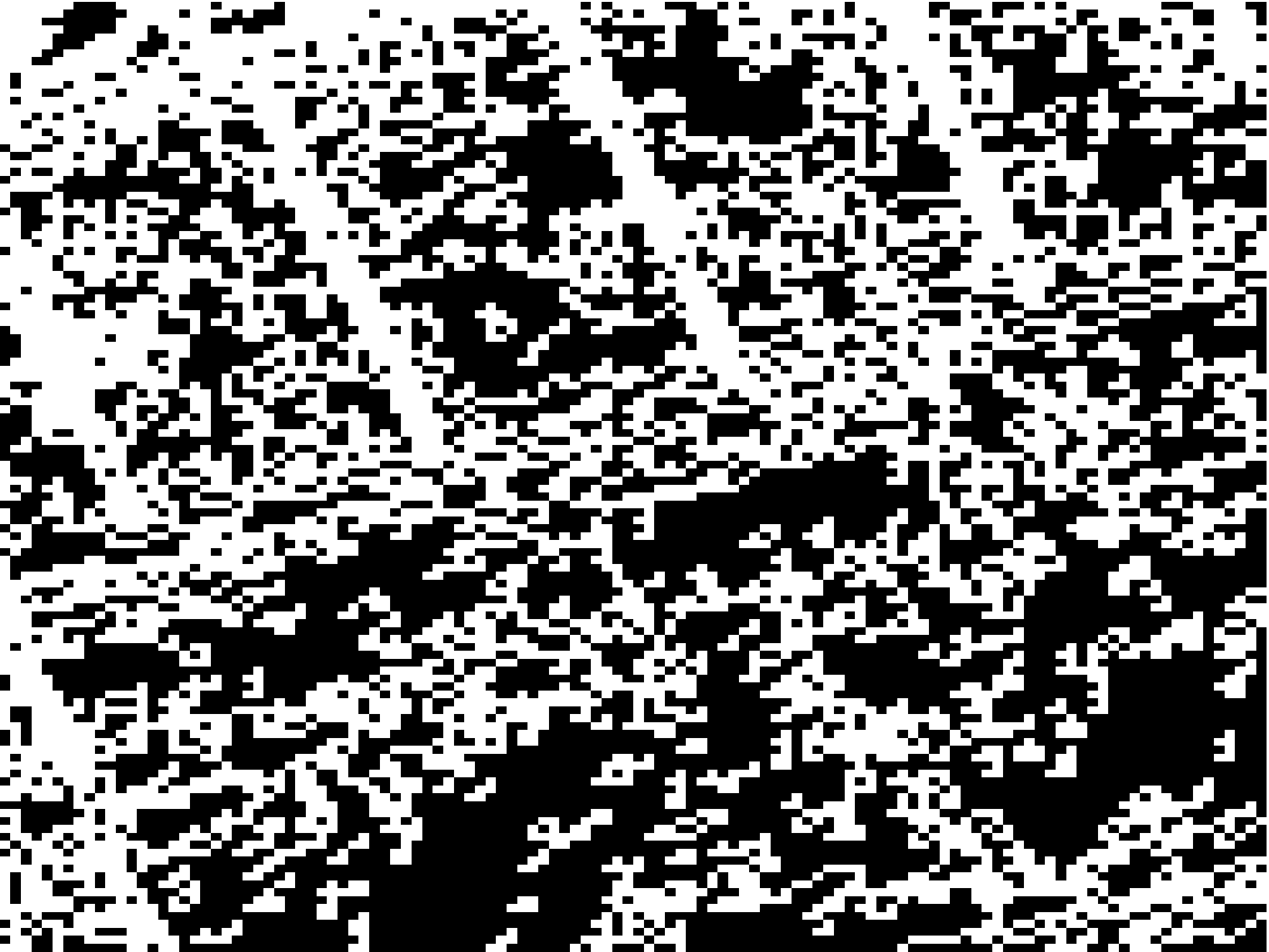}
		& \includegraphics[width=0.18\linewidth]{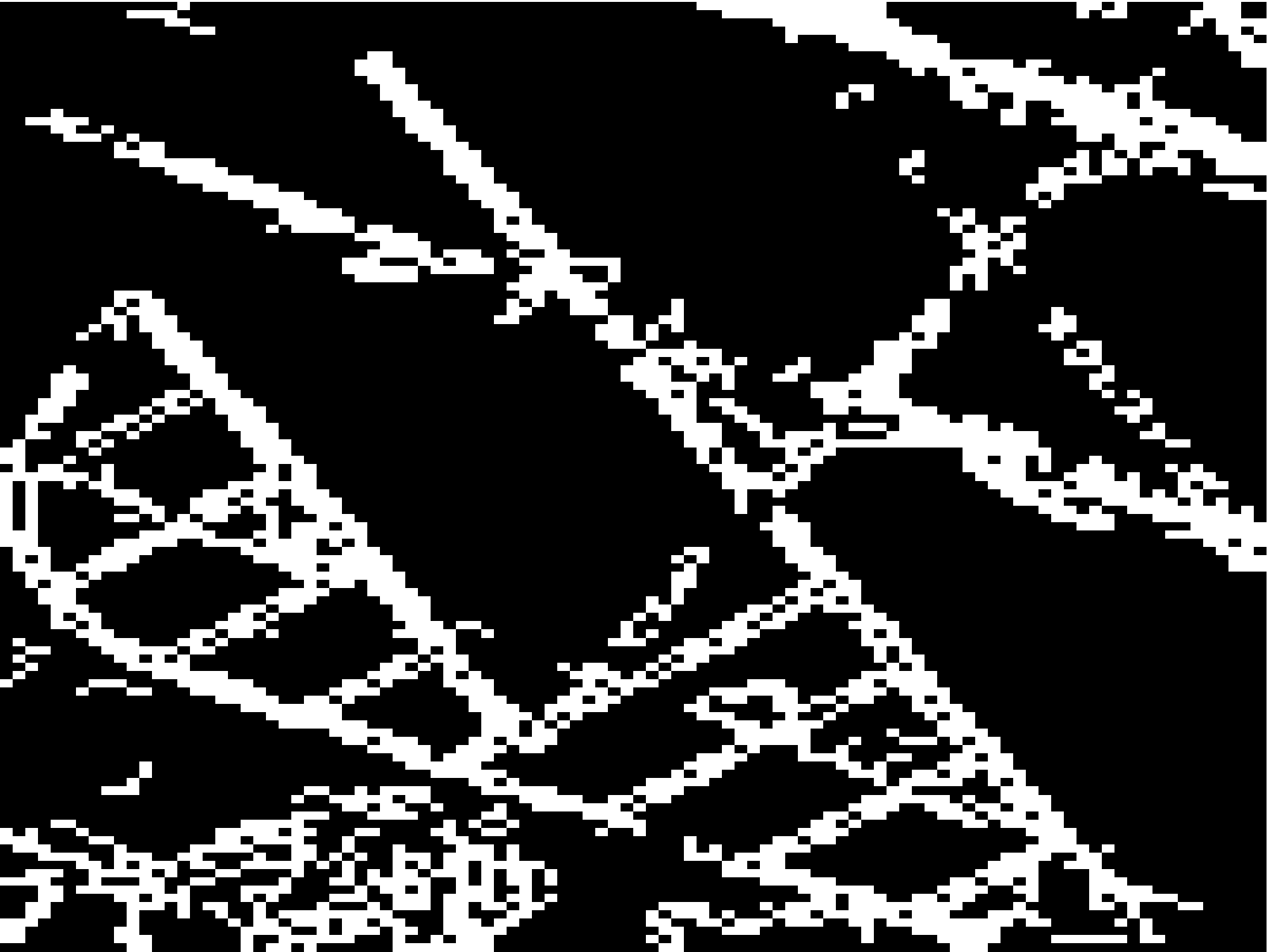} \\
		
		\rotatebox[origin=l]{90}{\shortstack{ Ours \\ Thresh} }	
		& \includegraphics[width=0.18\linewidth]{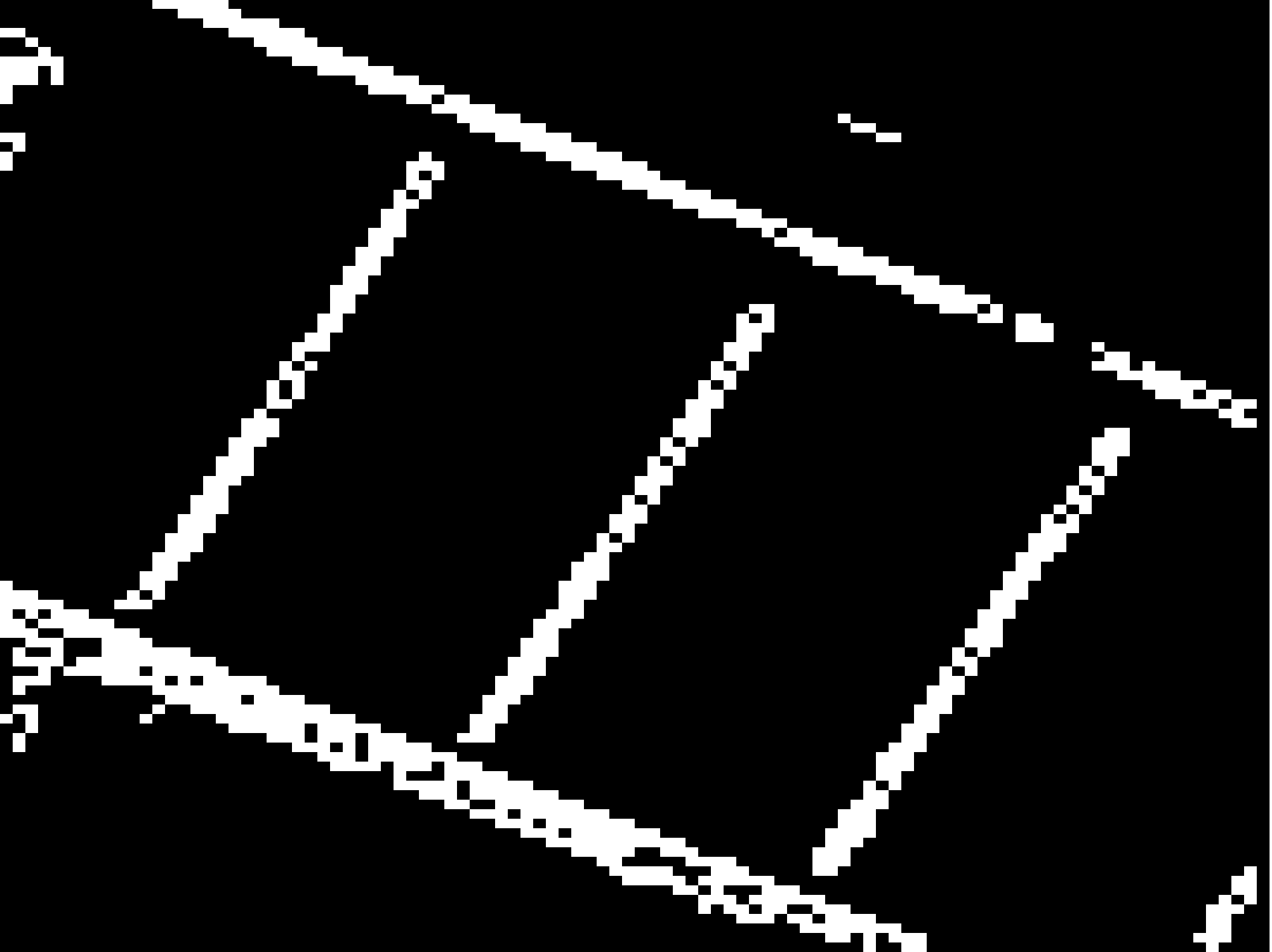}
		& \includegraphics[width=0.18\linewidth]{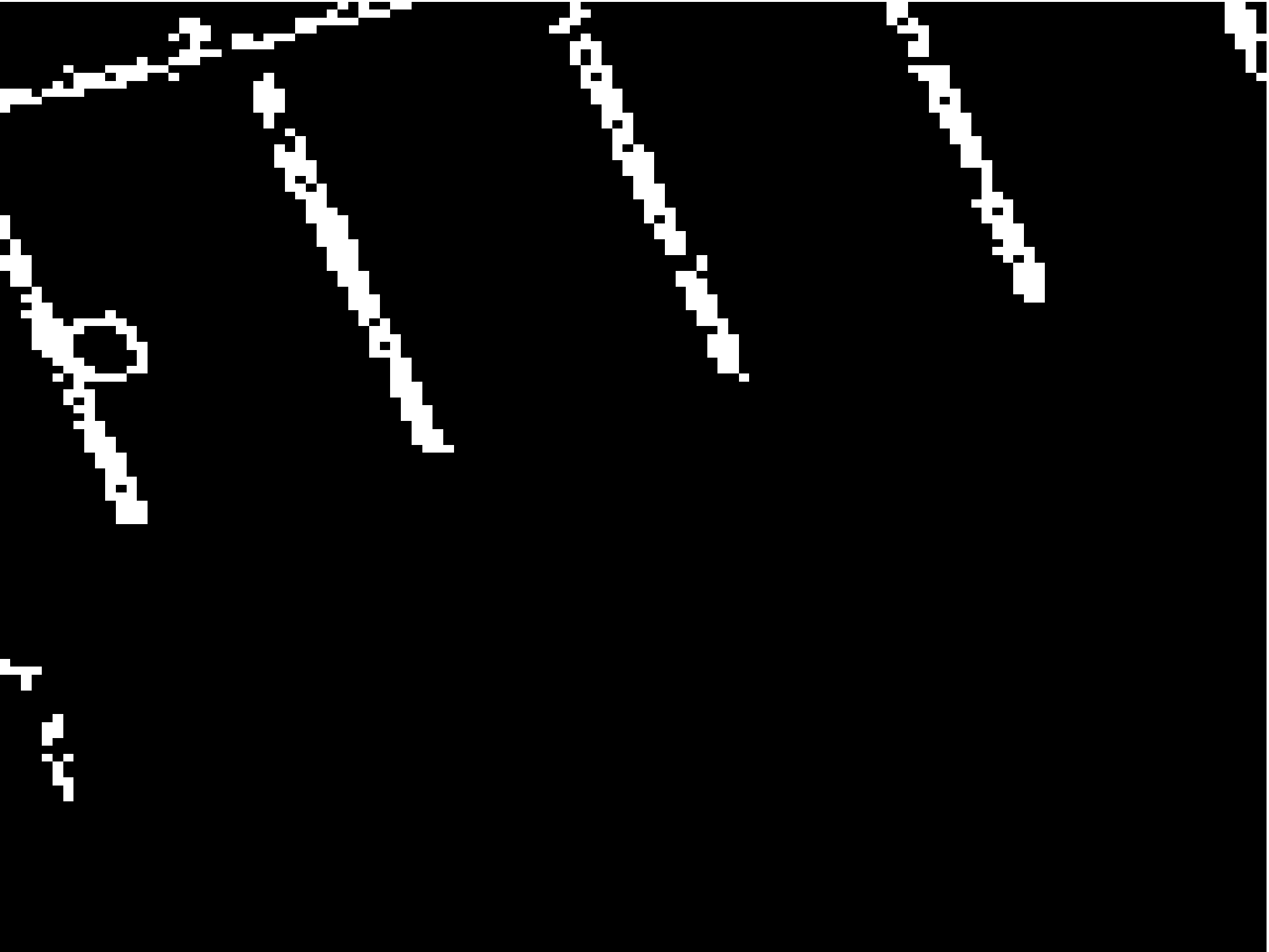}
		& \includegraphics[width=0.18\linewidth]{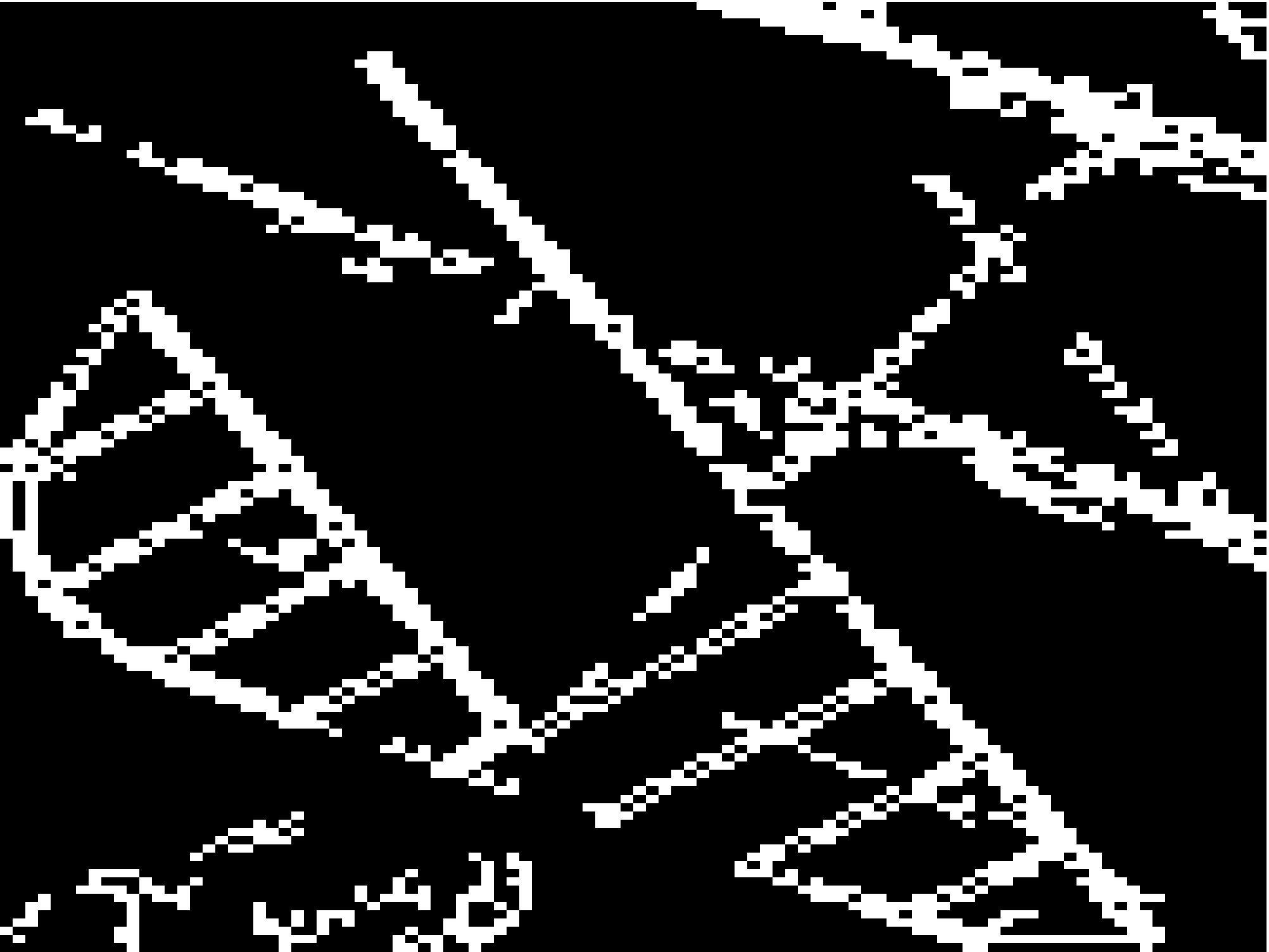} \\
		
		\rotatebox[origin=l]{90}{\shortstack{ Ground \\ Truth} }	
		& \includegraphics[width=0.18\linewidth]{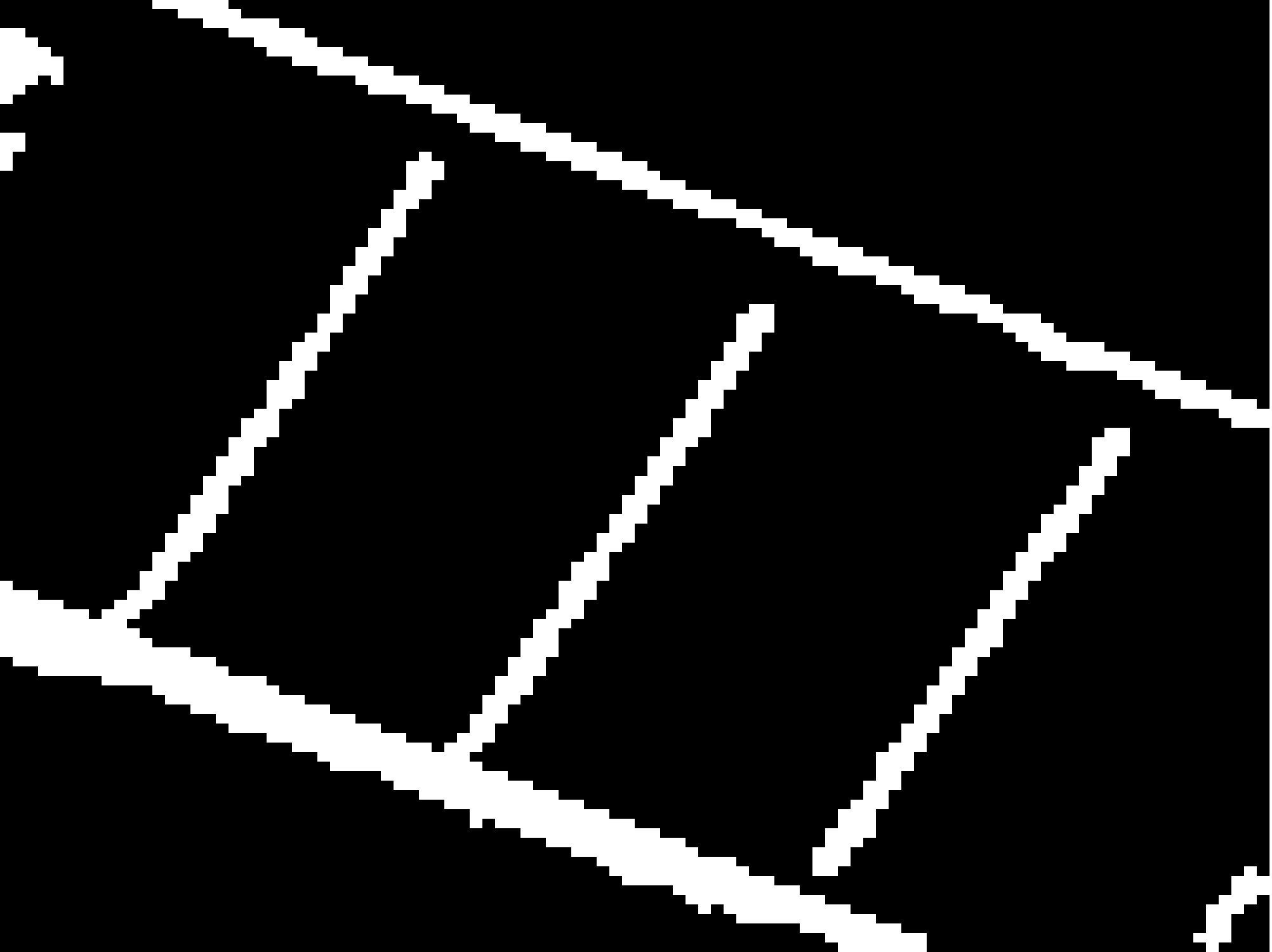}
		& \includegraphics[width=0.18\linewidth]{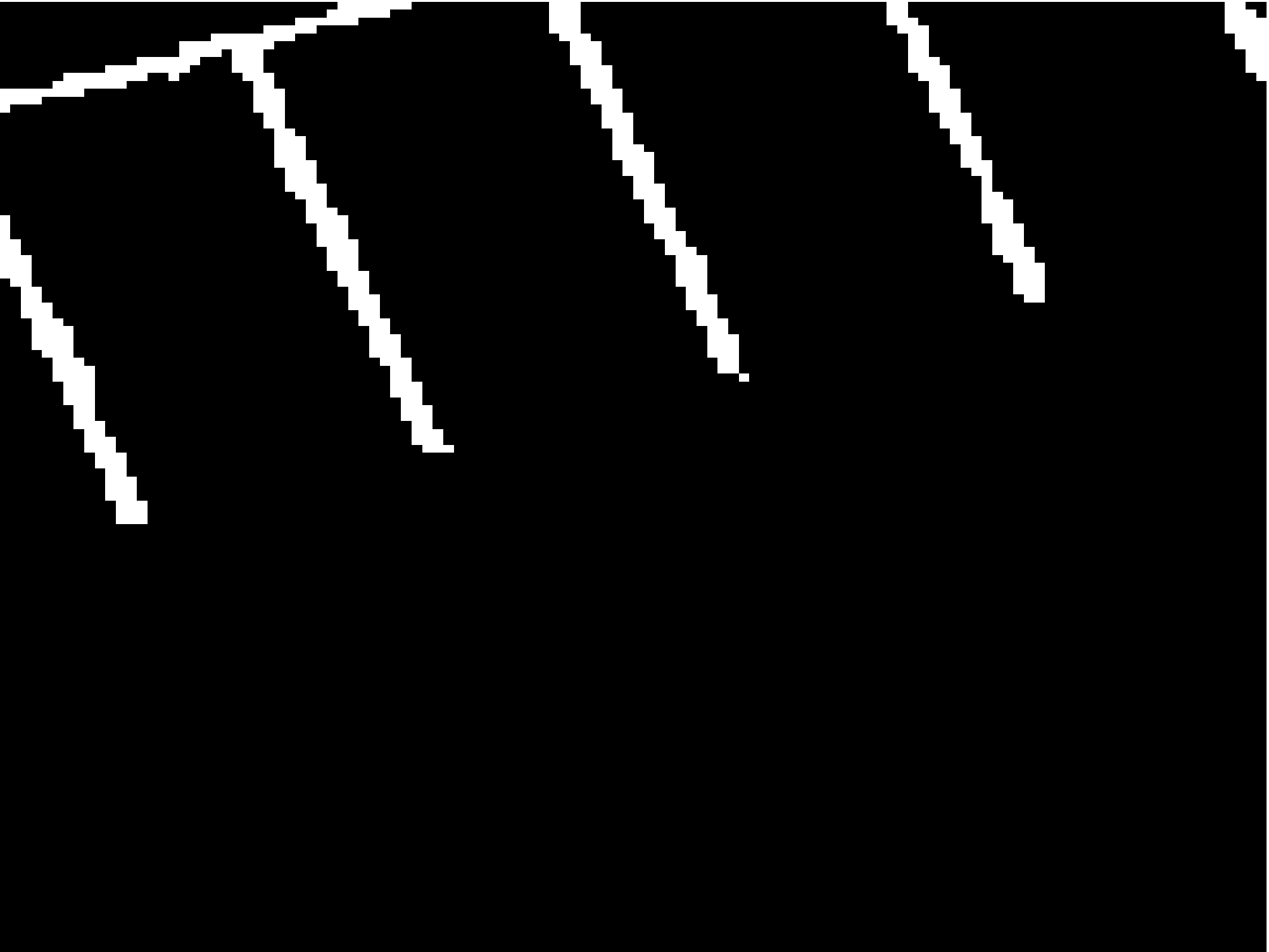}
		& \includegraphics[width=0.18\linewidth]{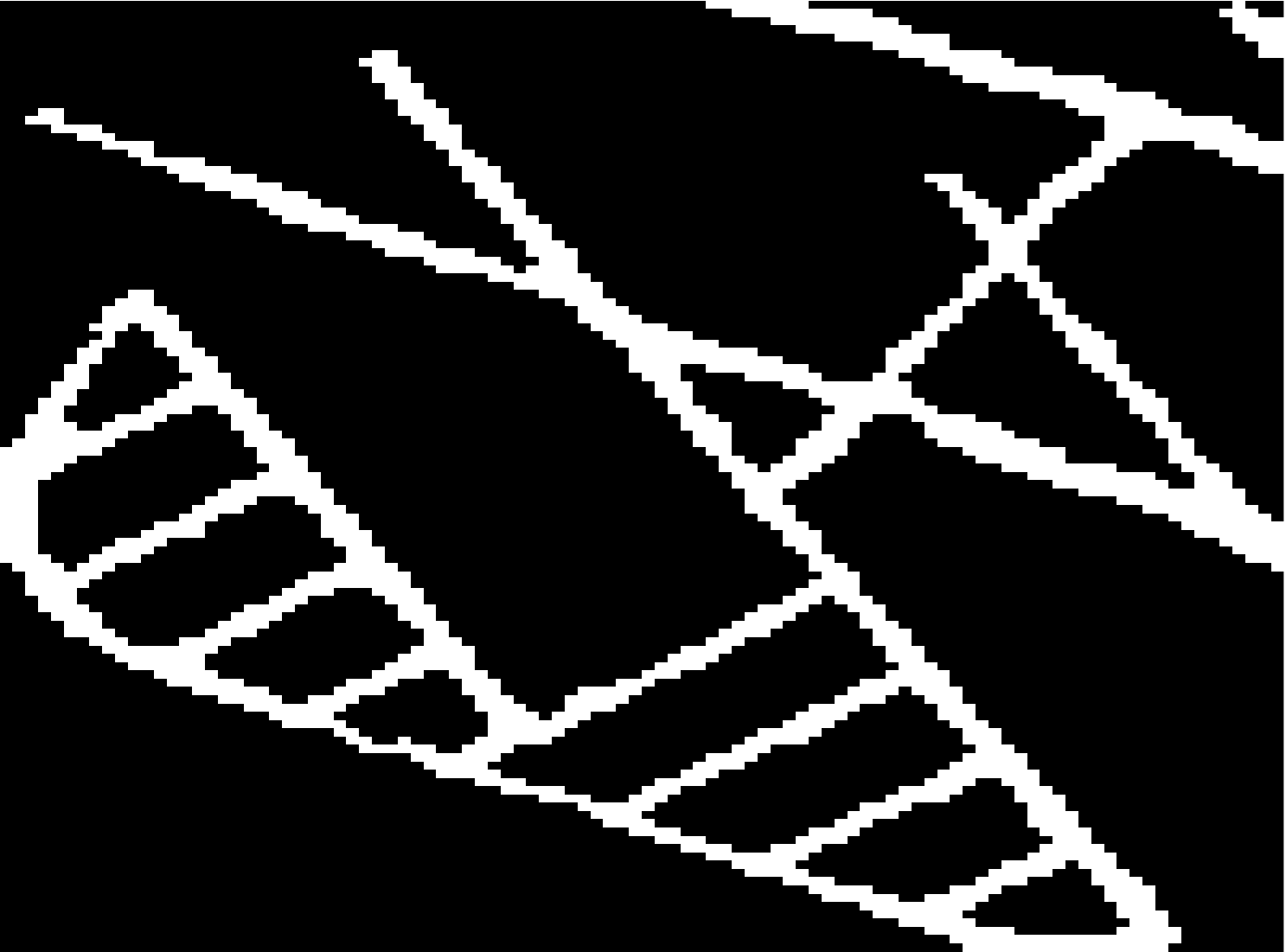} \\
	\end{tabular}
	\caption{Road marking segmentation examples.}
	\label{fig:extraction}
\end{figure}

In this sub-section we demonstrate the possibilities of our edge based fusion and map reconstruction formulation to achieve substantial improvements in road mark extraction applications from LIDAR based maps \cite{Kammel2008, Pu.Rutzinger.Vosselman.Elberink.2011, Hata2014, Guan14}. For this, we include here experimentation that compares the performance of a road mark extraction approach in patches reconstructed by the calibration method of \cite{Levinson.Thrun2010} and ours. To perform the assessment we base our segmentation approach and analysis on the work of \cite{Guan14}. Figure \ref{fig:extraction} illustrates the results obtained for 3 patch examples each of size $100 \times 100$ cells which corresponds to a $ 10 \times 10$ meter area. The last row in Figure \ref{fig:extraction} illustrates the reference ground truth patch obtained through the method described also in \cite{Guan14}. In addition, we also include Table \ref{tab:table2} which presents the metric evaluation results of these examples. Here, the F-score represents an overall score between a completeness measure that describes how complete the extracted markings are and the correctness measure which describes the ratio of the validity of the extracted marks. 

Summarizing, we find that there is a significant improvement on road mark extraction from maps obtained using our formulation. Patches reconstructed with the calibration based method present both lower road marking contrast from global smoothing and higher background reflectivity variations originating from the non-uniform laser responses, vehicle motion and scanning patterns. These two issues make the task of road mark segmentation a bit more challenging in comparison with one that would use the maps reconstructed using our formulation. The later approach which is characterized with maps of better quality, enhanced contrast, background uniformity and reduced artifact formation facilitates and improves upon road mark segmentation.    

\begin{table}
	\caption{\label{tab:table2} Road marking evaluation}
	\centering
	\scalebox{0.85}{
	\begin{tabular}{lcccc}
		\hline
		\\
		& Approach & Completeness & Correctness & F-score \\
		\hline
		\\
		\multirow{2}{*}{UM Patch 1} & L\&T \cite{Levinson.Thrun2010} & 0.8828 & 0.5596 & 0.6850 \\
		& Ours & 0.8744 & 0.9812 & \textbf{0.9247} \\
		\hline		
		\\
		\multirow{2}{*}{UM Patch 2} & L\&T \cite{Levinson.Thrun2010} & 0.9116 & 0.0800 & 0.1476\\
		& Ours & 0.8632 & 0.9675 & \textbf{0.8867} \\
		\hline
		\\
		\multirow{2}{*}{UMM Patch 3} & L\&T \cite{Levinson.Thrun2010} & 0.8051 & 0.6349 & 0.7100\\
		& Ours & 0.8046 & 0.8457 & \textbf{0.8246} \\		
		\hline
	\end{tabular}
	}
\end{table}
%

 		% Experiments
\section{Conclusion}
\label{sec:Conclusion}

In this investigation we proposed a novel computational mapping framework specifically designed to preserve edge sharpness in maps generated with data from robots/vehicles equipped with laser scanners. %Our new framework generates map-perspective gradients and applies the operators of selection, denoising and fusion followed by a reconstruction based on Poisson's formulation. %The selection and denoising operators are applied using iterative algorithms that minimize an $\ell_1$ sparse regularized least squares formulation. Fusion is then carried out simply as a sparsely weighted average of the map-perspective gradients. Finally, a Poisson formulation is used to reconstruct the map of ground reflectivity aimed at minimizing an $\ell_2$ term promoting consistency with the fused gradients of map-perspectives and a term that ensures equality constraints  with reference measurement data. 
Our experimentation demonstrates that our edge guided fusion and reconstruction formulation achieves substantial improvements over calibration based ones. In particular, we demonstrate improved map reconstruction quality, elimination of artifact formation from vehicle motion and laser scanning patterns, denoising capabilities and also, improvements in applications including localization and road mark segmentation. 
The main reason of this being that our method optimizes the estimate of the map of ground reflectivity instead of the individual laser responses.
Moreover, our approach removes the requirement of any post-factory reflectivity calibration. This represents a significant advantage if one considers the cost implied in calibration against standard reference targets and/or the time requirements of these processes, which are currently unfeasible for robots/vehicles produced in mass.

 		% Conclusions
\appendix
\section{Appendix}
\label{Sec:Appendix}

 Assuming a column-wise vectorization of a 2D signal of size $ N_y \times N_x $, the first order forward difference discrete gradient is defined point-wise as
\begin{equation} \label{Gradient}
[\gXbf ]_n = 
\begin{pmatrix} 
[\gxXbf]_n \\ 
[\gyXbf]_n 
\end{pmatrix} = 
\begin{pmatrix}
[\Xbf]_{n + N_y} - [\Xbf]_n \\
[\Xbf]_{n+1} - [\Xbf]_n
\end{pmatrix}.	
\end{equation} 
were $ \g_x $ and $ \g_y $ represent the horizontal and vertical components, respectively. 
In a similar way, the discrete Laplacian operator $ \L : \mathbb{R}^N \rightarrow \mathbb{R}^N$ is defined point-wise as
\begin{equation} \label{Laplacian}
[\L \Xbf ]_n = -4 [\Xbf]_n + [\Xbf]_{n+1} + [\Xbf]_{n-1} + [\Xbf]_{n+N_y} + [\Xbf]_{n-N_y}	
\end{equation}

\subsection{Soft-tresholding algorithm}

\begin{algorithm}[t]
	\caption{ Anisotropic FISTA for denoising gradients.} \label{gradientFieldDenoising}
	\begin{algorithmic}[1]
		\BState \textbf{input:} Gradient $ \gYbf $, step $\gamma$, and thresholding const. $\tau$.
		
		\BState \textbf{for each component field $ k \in \{x,y\}$}
		\State \quad \textbf{set: }  $t \leftarrow 1$, $s^0 \leftarrow \g_k \Ybf $ 
		\BState \quad \textbf{repeat} %$T = T_{init}$ \mbox{through}  $T = T_{final}$
		\State \quad \quad $\pmb{s}^t \leftarrow \eta_{\tau}(\pmb{s}^{t-1} - \gamma \nabla \Dcal_3( \pmb{s}^{t-1}) ) $
		\State \quad \quad $ q_t \leftarrow \frac{1}{2} \left ( 1 + \sqrt{1+4 q^2_{t-1}} \right )$ 
		\State \quad \quad $\g_k \Xbf^t \leftarrow \pmb{s}^t + ( (q_{t-1} -1)/q^t) ( \pmb{s}^t - \pmb{s}^{t-1}) $.
		\State \quad \quad $ t \leftarrow t + 1$
		\BState \quad \textbf{until: } stopping criterion
		\State \textbf{return: } Denoised gradient of map-perspective $ \gXbf^t $
	\end{algorithmic}
\end{algorithm}
To denoise a map-perspective, we propose to apply the soft-thresholding \cite{Donoho95} to its gradient-field. This method solves the sparse promoting least squares optimization  
\begin{equation} \label{denoising}
	\g_k \Xbf = \arg \min\limits_{ \g_k \Xbf \in \Xcal_{\g} } 
	\left \{  
	\Dcal_3( \g_k \Xbf ) + \lambda \Rcal( \g_k \Xbf )
	\right \},
\end{equation}
for each horizontal and vertical $ k = \{x,y\}$ direction, independently. Here,
$\lambda > 0$ controls the amount of regularization (i.e., sparsity).
The first term in \eqref{denoising} measures the gradient fidelity defined in the least squares sense as
\begin{equation} \label{Eq:l2Fidelity}
	\Dcal_3( \g_k \Xbf ) =  \frac{1}{2} 
	\left \| \g_k \Ybf - \g_k \Xbf \right \|_{\ell_2}^2 
\end{equation}
while the second term is the non-smooth $\ell_1$ sparse promoting regularizer defined by
\begin{equation} \label{Eq:l1Regularizer}
	\Rcal( \g_k \Xbf ) = \| \g_k \Xbf \|_{\ell_1}.
\end{equation}
The optimization in \eqref{denoising} can be iteratively solved using the accelerated gradient descent in \cite{Nesterov1983} along with the non-convex proximity projection method of \cite{Beck.Teboulle2009b}. The complete Algorithm in \ref{gradientFieldDenoising} has a rate of convergence of $O(1/k^2)$.    
 		% Appendix

%%%%%%%%%%%%%%%%%%%%%%%%%%%%%%%%%%%%%%%%%%%%%
%% Bibliography
%%%%%%%%%%%%%%%%%%%%%%%%%%%%%%%%%%%%%%%%%%%%%

{\small
\bibliographystyle{files/IEEEbib}
\bibliography{main}
}

\end{document}